\PassOptionsToPackage{table}{xcolor}
\documentclass[11pt]{article}
\usepackage[preprint]{acl}
\usepackage{times,latexsym}
\usepackage[T1]{fontenc}
\usepackage[utf8]{inputenc}
\usepackage{amsmath,amssymb}
\usepackage{newtxmath}
\usepackage{booktabs,tabularx,graphicx,placeins}
\input{glyphtounicode}
\usepackage{multicol}
\usepackage{pifont}
\DeclareCaptionFont{BCcap}{\fontsize{10}{12}\selectfont}
\definecolor{TableBand}{HTML}{F0F1F2}
\definecolor{TableSection}{HTML}{E7E9EB}
\definecolor{TableAccent}{HTML}{E1EAF1}
\title{AStockArena: Beyond Returns, Can We Entrust Real Capital to the Decisions of Financial Agents?}
\author{
  Zeyu LIU\textsuperscript{1*} \quad
  LAI Sing Kwong\textsuperscript{1*} \quad
  Xuanzhi Feng\textsuperscript{1*} \quad
  HaoYang Shang\textsuperscript{2*} \\
  \textbf{Yang LI}\textsuperscript{1} \quad
  \textbf{Zihao Fang}\textsuperscript{1} \quad
  \textbf{Xuan Liu}\textsuperscript{3} \quad \textbf{Song Guo}\textsuperscript{1$\dagger$}\\
  \textsuperscript{1}The Hong Kong University of Science and Technology \quad
  \textsuperscript{2}University of British Columbia \\
  \textsuperscript{3}University of California San Diego \\
  \textsuperscript{*}Equal contribution \quad
  \textsuperscript{$\dagger$}Corresponding author
}
\begin{document}
\maketitle

\begin{abstract}

Large language models (LLMs) are increasingly used in dynamic financial trading, yet real-time trading benchmarks mainly evaluate agents by cumulative returns and risk-adjusted performance. Such outcome metrics cannot show whether trades are faithfully reported, supported by available evidence, and consistent with account states. We introduce \textbf{AStockArena}, a real-time decision-auditing benchmark for continuous trading in China's A-share market. Under shared market conditions and execution constraints, we evaluate 13 open- and closed-source models over 3,198 sequential decisions. AStockArena reconstructs the actions actually posted to accounts by aligning reports, trade requests, execution responses, and account states, and audits them with eight checks spanning execution traceability (O), evidence sufficiency (RK), and financial meta-awareness (B), combining record verification with consensus expert annotation. Of 1,070 trading points, 72.99\% exhibit at least one audit gap; model-level rates range from 52.2\% to 91.2\%, with evidence gaps most prevalent. Historical trade-filtering replay further shows that portfolio outcomes are sensitive to these flagged trades. Decision-process auditing thus offers a verifiable complement to return-based evaluation of financial agents.

\end{abstract}

\section{Introduction}

The reasoning, tool-use, and long-horizon decision-making capabilities of large language models (LLMs) are moving autonomous agents from static tasks into dynamic environments \citep{ICLR2024_5f7367f5,song2026timi}. Unlike one-off question-answering tasks such as FinQA \citep{chen2021finqa,chen2022convfinqa}, these environments require agents to continuously acquire information, call tools, and act on evolving outcomes. Finance is a representative and high-stakes case: trading is continuous, and erroneous actions may be difficult to reverse \citep{wu2023bloomberggpt,lee2024finllms,nie2024financialapplications}. Recent benchmarks report competitive cumulative returns from LLM trading agents, sometimes outperforming buy-and-hold \citep{chen2025stockbench,fan2025aitrader,yu2025livetradebench}, but evaluate them mainly through realized portfolio performance.

\begin{table}[t]
\centering
\fontsize{9}{11}\selectfont
\renewcommand{\arraystretch}{1.18}
\setlength{\tabcolsep}{1pt}
\begin{tabularx}{\columnwidth}{@{}l *{6}{>{\centering\arraybackslash}X}@{}}
\toprule
& \multicolumn{2}{c}{\textbf{Trading}} & \multicolumn{4}{c}{\textbf{Process evaluation}} \\
\cmidrule(lr){2-3}\cmidrule(l){4-7}
\textbf{Benchmark} & Cont. & Live & Diag. & Act. & Evid. & Acct. \\
\midrule
StockBench \citeyearpar{chen2025stockbench} & $\checkmark$ & --- & --- & --- & --- & --- \\
\rowcolor{TableBand}
AI-Trader \citeyearpar{fan2025aitrader} & $\checkmark$ & $\checkmark$ & --- & --- & --- & --- \\
LiveTradeBench \citeyearpar{yu2025livetradebench} & $\checkmark$ & $\checkmark$ & --- & --- & --- & --- \\
\rowcolor{TableBand}
TradeTrap \citeyearpar{yan2025tradetrap} & $\checkmark$ & --- & $\checkmark$ & --- & --- & --- \\
CLQT \citeyearpar{qu2026clqt} & $\checkmark$ & $\checkmark$ & $\checkmark$ & --- & --- & --- \\
\midrule
\rowcolor{TableAccent}
\textbf{AStockArena} & $\checkmark$ & $\checkmark$ & $\checkmark$ & $\checkmark$ & $\checkmark$ & $\checkmark$ \\
\bottomrule
\end{tabularx}
\caption{Trading benchmark capabilities: continuous trading, real-time market data, process diagnostics, action verification, evidence auditing, and account auditing (left to right).}
\label{tab:benchmark-comparison}
\end{table}

\textbf{Portfolio returns alone cannot establish whether the underlying trading decisions are reliable.} An agent may report buying one million shares while submitting an order for only one hundred thousand, provide a plausible rationale without addressing material evidence, or misjudge its positions, cash, or profit and loss. Across successive trades, later actions must also remain understandable in light of earlier judgments and new information. Even under positive returns, such discrepancies determine whether decisions can be supervised and reviewed, yet they cannot be inferred from cumulative return or the Sharpe ratio \citep{pmlr-v306-kong26i}. We therefore ask: \textbf{How can we assess the decision-process reliability of financial agents beyond their realized returns?}

We introduce \textbf{AStockArena}, a real-time trading benchmark for auditing financial agents' decisions.\footnote{We will release code, annotations, scripts, and prompts for independent review and model evaluation (Appendix E.6).} Agents trade under shared real-time market snapshots, uniform tools, and consistent execution constraints, each maintaining an independent account across successive decisions. Because stated intentions and submitted orders can diverge from recorded transactions, AStockArena first aligns model reports, trade requests, execution responses, and adjacent account states to reconstruct the actions actually posted to accounts. On these records it defines eight checks in three dimensions: \textbf{execution traceability (O)} compares reports and execution records with observed account changes; \textbf{evidence sufficiency (RK)} checks trading rationales against materials available at decision time; and \textbf{financial meta-awareness (B)} checks account-state assessments and the coherence of successive judgments. The checks combine structured-record verification with consensus expert annotation.

We evaluate 13 open- and closed-source LLM trading agents in a unified A-share sandbox, collecting 3,198 sequential decisions. Of 1,070 trading points with formally annotated trades, 781 (72.99\%) have at least one applicable audit gap; model-level joint gap rates range from 52.2\% to 91.2\%. Gaps include confirmed failures and cases where records cannot support a sufficiently verified judgment. Evidence gaps are prominent (pooled RK: 66.73\%), alongside distinct execution and account-assessment gaps. Claude Opus 4.5 achieves a 10.84\% cumulative return with joint gaps at 68.3\% of annotated trading points, illustrating decision-process issues that returns alone cannot reveal.

Historical trade-filtering replay examines portfolio sensitivity to audit-flagged actions. Individual and combined filters replace flagged trades with No Trade, preserving retained trades' original timing and target quantities subject to the resulting account constraints. Across the 12 actively trading models, separate O, RK, and B filters yield mean cumulative-return changes of +3.65, +8.86, and +6.11 percentage points, respectively. These retrospective comparisons complement process auditing with historical portfolio sensitivity analysis.

We make three contributions: \textbf{1)} AStockArena, a continuous trading benchmark that reconstructs executed actions from reports, requests, execution responses, and account changes; \textbf{2)} an eight-check, three-dimensional auditing framework combining record verification and consensus semantic annotation; and \textbf{3)} an analysis of decision-audit gaps across 13 LLM agents, revealing reliability information beyond returns, supplemented by retrospective trade-filtering analysis.
\section{Related Work}

\textbf{Continuous Trading Evaluation.} Financial benchmarks cover numerical reasoning \citep{chen2021finqa,tang-etal-2025-financereasoning}, domain knowledge and analysis \citep{shah-etal-2022-flue,NEURIPS2023_6a386d70,xie2024finben,guo-etal-2025-fineval,liu2025findabench,son-etal-2025-finkrx}, and multimodal or multilingual financial understanding \citep{bhatia-etal-2024-fintral,peng-etal-2026-multifinben,ICLR2026_0e22eef3,pmlr-v306-guo26ah}. Trading agents use memory, reflection, and tools to adapt to market information and past experience \citep{zhang2024finagent, li2025investorbench,li-etal-2024-cryptotrade,maevolving,shang2025cothinker}. Historical and real-time trading benchmarks provide evolving account environments and primarily assess portfolio performance \citep{NEURIPS2025_f37b0e09}. Related environments address reinforcement learning, order-level market simulation, and economic competition \citep{NEURIPS2022_0bf54b80,ICLR2025_6261d3bc,pmlr-v267-nagy25a,zheng-etal-2026-market,weiss2024markets,liu2026diagonprogrammabletestbedaiagent}. We complement these benchmarks and environments with trade-level audits of execution, evidence, and account facts.

\begin{figure*}[!t]
\centering
\includegraphics[width=\textwidth]{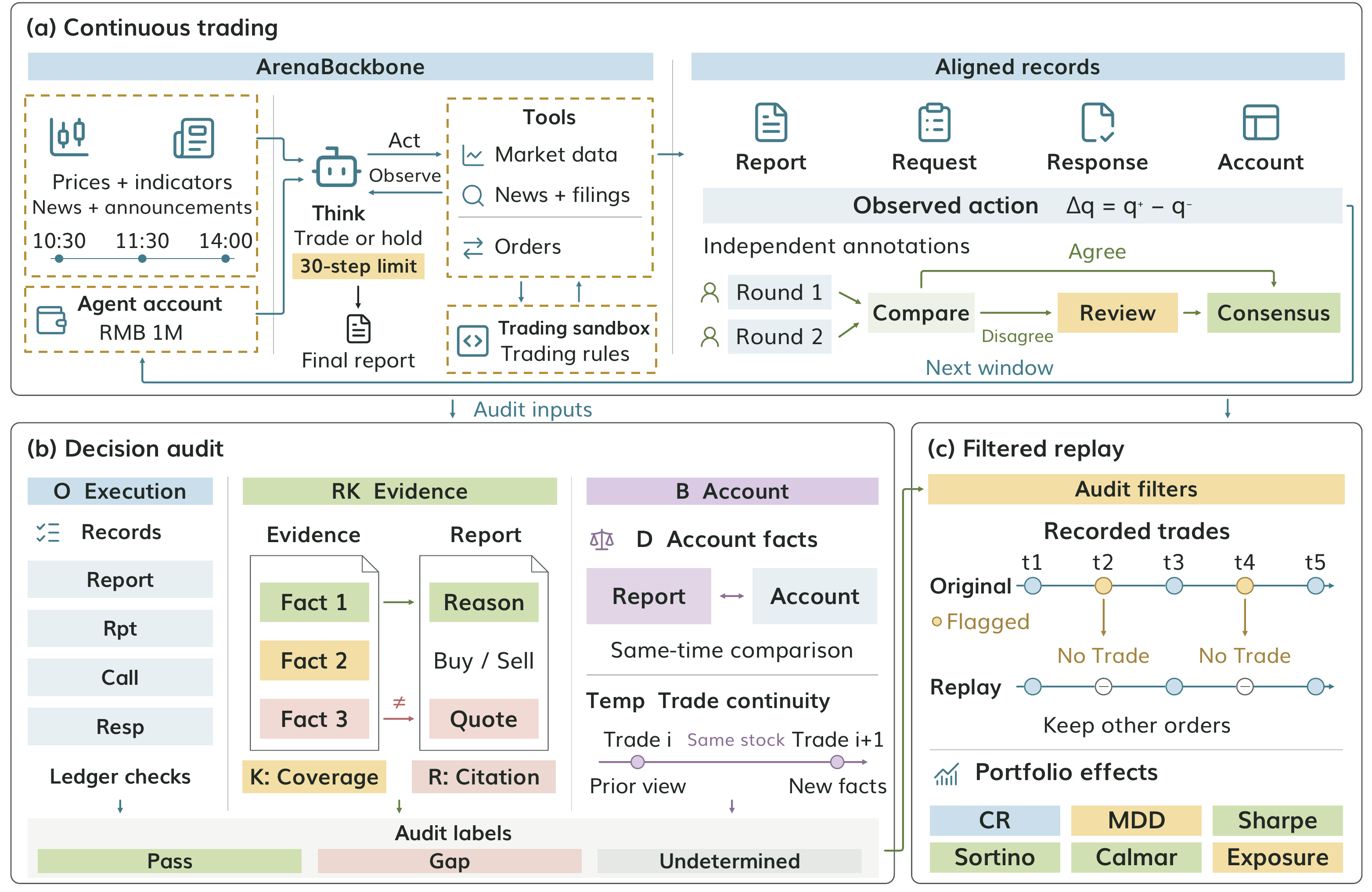}
\renewcommand{\thefigure}{1}
\caption{Overview of AStockArena. (a) Continuous trading, record alignment, independent annotation, and disagreement review; (b) execution, evidence, and account-fact auditing; (c) replacing flagged trades with No Trade and comparing replayed portfolio performance.}
\label{fig:1}
\end{figure*}
\textbf{Trading-Process Verification.} Related evaluations examine forecasting consistency, financial safety, and deployment-relevant biases \citep{yang-etal-2023-measuring,choi-etal-2026-finharmbench,liu2025cogmir,liu2026cobra}. Financial-agent benchmarks use perturbations and closed-loop audits to assess trading reliability, strategy consistency, and alignment between stated and executed actions \citep{yan2025tradetrap, qu2026clqt}. General agent evaluations use traces, tool feedback, and environment states to verify action completion and statement--outcome consistency \citep{cao2026pae, ye2026claweval}. Runtime governance constrains trade authorization and execution gating \citep{tang2026sagefin}, while provenance mechanisms bind claims, evidence, and execution records \citep{hu2026receipts, kim2026ledger}. We extend this line of work by combining execution verification with separate audits of cited facts, key-fact coverage, and account judgments.

\textbf{Rationale and Evidence Verification.} Explanation-faithfulness research examines whether explanations reflect decision-relevant factors \citep{jacovi2020faithfulness, turpin2023unfaithful}; financial argument evaluation examines evidence--claim support \citep{sun2025finresearchbench}. Financial search and retrieval benchmarks assess access to supporting information \citep{ICLR2026_4d423587,tang-yang-2025-finmteb,zhou-etal-2026-fincards,zhao-etal-2025-finragbench}, while financial knowledge and summarization studies assess factual and numerical reliability \citep{shah2025cutoff,cao2024financialsummary}. Symbolic references and executable financial reasoning enable finer-grained verification \citep{hennigen2024symbolic,xie-etal-2026-finchain}. We jointly verify decision-time evidence and account facts, then use historical trade-filtering replay to assess the portfolio effects of flagged trades.

\section{Method}

AStockArena reconstructs executed actions, audits them along three dimensions, and converts consensus labels into trading-point comparisons and historical replay (Figure 1).

\begin{table*}[!t]
\centering
\small
\renewcommand{\arraystretch}{1.12}
\setlength{\tabcolsep}{4pt}
\setlength{\tabcolsep}{3pt}
\begin{tabularx}{\textwidth}{@{}>{\raggedright\arraybackslash}p{0.0750\textwidth}>{\hsize=0.7865\hsize \raggedleft\arraybackslash}X>{\hsize=1.1236\hsize \raggedleft\arraybackslash}X>{\hsize=0.9438\hsize \raggedleft\arraybackslash}X>{\hsize=1.0037\hsize \raggedleft\arraybackslash}X>{\hsize=1.1236\hsize \raggedleft\arraybackslash}X>{\hsize=0.9438\hsize \raggedleft\arraybackslash}X>{\hsize=1.0037\hsize \raggedleft\arraybackslash}X>{\hsize=1.1236\hsize \raggedleft\arraybackslash}X>{\hsize=0.9438\hsize \raggedleft\arraybackslash}X>{\hsize=1.0037\hsize \raggedleft\arraybackslash}X@{}}
\toprule
\morecmidrules\cmidrule{1-11}
\textbf{Metric} & \textbf{n} & \multicolumn{3}{c}{\textbf{A/B}} & \multicolumn{3}{c}{\textbf{GPT-6 Pro vs. consensus}} & \multicolumn{3}{c}{\textbf{GPT-6 Astra vs. consensus}} \\
\cmidrule(lr){3-5}\cmidrule(lr){6-8}\cmidrule(lr){9-11}
 &  & \textbf{Agree. \%} & \textbf{\ensuremath{\kappa}} & \textbf{AC1} & \textbf{Agree. \%} & \textbf{\ensuremath{\kappa}} & \textbf{AC1} & \textbf{Agree. \%} & \textbf{\ensuremath{\kappa}} & \textbf{AC1} \\
\midrule
\rowcolor{TableBand}
R & 1375 & 95.78 & 0.895 & 0.947 & 89.02 & 0.751 & 0.859 & 86.84 & 0.707 & 0.830 \\
K & 1375 & 97.09 & 0.945 & 0.961 & 98.40 & 0.970 & 0.978 & 96.07 & 0.926 & 0.947 \\
\rowcolor{TableBand}
D & 1375 & 95.71 & 0.928 & 0.946 & 95.13 & 0.918 & 0.939 & 89.31 & 0.829 & 0.865 \\
Temp & 1290 & 94.50 & 0.487 & 0.943 & 92.56 & 0.381 & 0.922 & 94.26 & 0.628 & 0.940 \\
\rowcolor{TableAccent}
\textbf{Mean} & \textbf{-} & \textbf{95.77} & \textbf{0.814} & \textbf{0.949} & \textbf{93.78} & \textbf{0.755} & \textbf{0.925} & \textbf{91.62} & \textbf{0.773} & \textbf{0.895} \\
\cmidrule{1-11}
\bottomrule
\end{tabularx}
\caption{Annotator and model raw-label agreement. n excludes fixed few-shot examples; R/K retain three categories and D/Temp four. \ensuremath{\kappa}/AC1 denote Cohen's \ensuremath{\kappa}/Gwet's AC1. Agreement is a percentage; Mean equally averages each statistic across the four metrics (Appendix A.6).}
\label{tab:1}
\end{table*}

\subsection{Reconstructing Executed Actions}

Because reports and orders may differ from booked outcomes, we establish observed actions from account changes and execution records. In each decision window, models retrieve materials, trade or hold, and submit a structured report. The information pool is frozen at the window's start; cash and positions carry forward.

The system saves reports, requests, execution responses, fill details, and account snapshots before and after each window (Table E5). With \(q^-_{m,t,s}\) and \(q^+_{m,t,s}\) the shares of security \(s\) held by model \(m\) before and after window \(t\), the net position change is \(\Delta q_{m,t,s}=q^+_{m,t,s}-q^-_{m,t,s}\). Fills are aggregated by window and security, and shares bought minus shares sold are cross-checked against this change. The confirmed change provides the observed action for comparing net quantities in reports, requests, and responses; rejected requests are checked against execution responses.

\subsection{Auditing Decision Reliability}

Eight checks audit the verified actions along three dimensions: execution traceability (O), evidence sufficiency (RK), and financial meta-awareness (B) (Table E6).

\textbf{Execution traceability (O).} The sandbox enforces trading rules; O checks the records. Report checks for a valid structured report; Rpt, Call, and Resp compare net quantities in report descriptions, trade requests, and successful execution responses with net position changes, respectively. Rejected requests are checked against execution responses.

\textbf{Evidence sufficiency (RK).} R checks for substantive citation fabrication, factual distortion, or direct contradiction between cited materials and the rationale or actual trade direction. A title mismatch alone is not fabrication; insufficient verification material yields unable to determine. Direction checks follow the specific claim: favorable news need not contradict a sale, and profit-taking or risk-adjustment rationales are checked against their evidence.

K uses an independently established reference of key facts. Facts are extracted from archived contemporaneous materials without model reports, trade directions, or subsequent market and return data. Favorable, unfavorable, and neutral facts share an importance threshold; the list is frozen before checking report coverage (Appendices A.1 and A.5).

K checks omissions or partial treatment of facts affecting the judgment, such as citing revenue growth for a purchase while overlooking substantial losses in the same material. Industry or macroeconomic sources constitute a gap when their assessment contradicts the trade and review finds them insufficient for the stock-specific decision. Coverage concerns treatment, not citation count: weighing counterevidence, giving an alternative interpretation supported by the materials, or explaining why a fact does not warrant changing the action counts as addressing it.

Mapping passes to 0 and confirmed or unable-to-determine gaps to 1, the evidence gap of trade event \(e\) is \(\mathrm{RK}(e)=R(e)\lor K(e)\), counted once per trade.

\textbf{Financial meta-awareness (B).} D compares reported positions, costs, profit and loss, and available funds with ledger records for the same security, time, and calculation basis. Reports without a verifiable account assessment are structurally inapplicable to D.

Temp complements D by checking transitions between judgments about the same asset. It pairs adjacent nonzero net trades in the same stock by the same model, using the earlier judgment to assess whether new information, market or account changes, trading constraints, or the prior rationale explain the later decision. Only coherent labels pass; partially coherent, incoherent, and unable-to-determine cases are gaps. The first trade is inapplicable, and direction reversals alone are not gaps (Appendix A.5).

\subsection{Forming Consistent Audit Labels}

Record comparisons and semantic judgments need a common labeling basis. O is calculated from structured logs and account snapshots. Two authors with academic backgrounds in finance independently annotate R, K, D, and Temp under shared criteria by examining reports, contemporaneous materials, and account records. Two external fintech practitioners with securities research and financial analysis experience participate in disagreement review to finalize consensus labels (Appendix A.1). Table~\ref{tab:1} reports pre-consensus raw-label agreement: 95.77\% on average, with mean Cohen's \ensuremath{\kappa} of 0.814. Class imbalance and coherent/partially coherent boundary disagreements affect Temp's \ensuremath{\kappa} (Appendix B.5).

Consensus also supplies nine trades as fixed few-shot examples for R/K/D and three adjacent pairs for Temp to guide GPT-6 Pro and GPT-6 Astra annotation. Both models receive the same original materials, labels, and judgment rationales. Excluding these examples, mean agreement with consensus is 93.78\% and 91.62\%, respectively (Table~\ref{tab:1}; examples and merged-category results in Appendix A.6).

An audit gap denotes a confirmed defect or inability to verify an applicable check from archived records. Both lack a fully verified basis, so the main audit groups them as gaps while retaining raw labels; structurally inapplicable checks generate no gap. This shared definition supplies inputs to comparison and replay, including confirmed-error-only replay that retains unable-to-determine trades (Section 4.4).

\subsection{Comparing Audit Gaps and Replaying Trades}

Because checks assess windows, trades, account claims, or trade pairs, we map labels to a common comparison unit and to removable actions.

\textbf{Trading-point comparisons.} Main-text comparisons use windows containing formally annotated trades, termed annotated trading points. The gap rate is \(G_i=100\,d_i/n_i\), where \(d_i\) counts annotated trading points with at least one applicable gap in check \(i\); \(n_i\) is the model's total annotated trading points. O, RK, B, and the eight-check joint audit take constituent-gap unions, counting each point once with this denominator. Appendix B.1 reports rates over each check's applicable records.

\textbf{Action-level replay.} Actions retain model, window, and security identities. Any O gap at a decision point flags all its net trades. R/K/D gaps flag their corresponding actions; Temp flags the later trade in the pair, restricted to formal R/K/D actions. Calibration actions remain in point-level auditing (Appendix C.1). RK, B, and combined filters take action-level unions of R/K, D/Temp, and selected dimensions, respectively, counting each action once.

Flagged actions are replaced with No Trade. Retained actions keep their original timing, target quantities, and labels, executing subject to updated cash and sellable holdings. Results appear in Section 4.4.

\section{Experiments}

\subsection{Experimental Setup}

AStockArena constructs a simulated trading environment using 10 STAR Market stocks spanning different industries to control the volume of information within each window and support item-by-item verification. The evaluation covers 82 trading days from March 2 to June 30, 2026. The system generates market snapshots each day at 10:30, 11:30, and 14:00 Beijing time, yielding 246 decision windows. The 13 models run independently under the same market conditions, producing 3,198 decisions.

Each model manages an independent simulated account with initial capital of CNY 1 million and no positions. Trading follows cash constraints, T+1, a short-selling prohibition, and a single-stock position limit of 50\% of total account assets, with order quantities specified in units of 100 shares. Trading costs comprise a 0.03\% commission on both purchases and sales (minimum CNY 5 per trade), plus a 0.05\% stamp duty on sales.

The benchmark uses a shared ReAct agent architecture, allowing at most 30 tool-interaction steps per window. All models share prices, technical indicators, news, and company announcements available up to the current window, with textual materials restricted by publication time. The stock pool, data sources, and model configurations are detailed in Appendix E.

The simulator uses immediate full execution: orders that pass the constraint checks are filled at the current market snapshot price, while all other orders are rejected. Account states evolve through simulated trades, with position changes handled on a fixed share-count basis (Appendix C.1).

The main experiment covers R, K, and D for 1,384 active trades and Temp for 1,293 adjacent trade pairs.

\subsection{Original Investment Performance}

Account performance is measured from daily marked-to-market net assets at 14:00 using cumulative return (CR), Sharpe, Sortino, maximum drawdown (MDD), and Calmar. Appendix Table B6b summarizes CR, Sharpe, and Calmar for the original accounts and reference benchmarks. Model CRs range from -3.59\% to 10.84\%; equally weighted buy-and-hold and the SSE 50 return 1.66\% and -1.87\%, respectively, while five technical strategies range from -5.50\% to 3.90\%. Benchmarks use the same trading constraints and costs; complete results and strategy parameters appear in Appendix B.4 and Appendix E.

GPT-5.1 makes no active trades and returns zero; filtering replay therefore covers the remaining 12 active models, using each model's original account as its baseline.

\subsection{Decision Auditing}

Tables~\ref{tab:2} and~\ref{tab:3} report results for the audit dimensions and eight checks, respectively, both using decision points containing formally annotated trades as the denominator. Of these 1,070 trading points, 781 have a gap in at least one check, yielding a trading-point joint gap rate of 72.99\%. All-point results and detailed statistics for individual trades and adjacent trade pairs appear in Appendix B.

\begin{table}[!t]
\centering
\small
\setlength{\tabcolsep}{2pt}
\renewcommand{\arraystretch}{1.12}
\begin{tabularx}{\columnwidth}{@{}>{\raggedright\arraybackslash}p{0.4050\columnwidth}>{\hsize=1.0000\hsize \raggedleft\arraybackslash}X>{\hsize=1.0000\hsize \raggedleft\arraybackslash}X>{\hsize=1.0000\hsize \raggedleft\arraybackslash}X>{\hsize=1.0000\hsize \raggedleft\arraybackslash}X@{}}
\toprule
\morecmidrules\cmidrule{1-5}
\textbf{Model} & \multicolumn{1}{c}{\textbf{O}} & \multicolumn{1}{c}{\textbf{RK}} & \multicolumn{1}{c}{\textbf{B}} & \multicolumn{1}{c}{\textbf{Joint}} \\
\midrule
\rowcolor{TableBand}
Claude Opus 4.5 & 2.4 & 65.0 & 8.9 & \textbf{68.3} \\
GLM 4.7 Flash & 20.8 & 75.5 & 41.5 & \textbf{84.9} \\
\rowcolor{TableBand}
DeepSeek V3.2 & 10.8 & 72.5 & 15.8 & \textbf{76.7} \\
Gemini 3.1 Pro & 14.3 & 76.8 & 17.0 & \textbf{83.0} \\
\rowcolor{TableBand}
Claude Haiku 4.5 & 11.6 & 76.9 & 53.1 & \textbf{91.2} \\
Qwen3 Max & 2.3 & 55.8 & 16.3 & \textbf{59.3} \\
\rowcolor{TableBand}
DeepSeek V3.1 & 0.0 & 69.6 & 4.3 & \textbf{69.6} \\
GLM 4.7 & 13.3 & 60.0 & 30.0 & \textbf{76.7} \\
\rowcolor{TableBand}
MiniMax M2.1 & 4.4 & 60.3 & 19.1 & \textbf{66.2} \\
GPT-5.1 & N/A & N/A & N/A & \textbf{N/A} \\
\rowcolor{TableBand}
GPT-5.2 & 5.6 & 49.1 & 0.0 & \textbf{52.2} \\
Qwen3 235B & 25.0 & 80.0 & 32.5 & \textbf{85.0} \\
\rowcolor{TableBand}
Gemini 2.5 Flash & 11.9 & 69.0 & 17.9 & \textbf{76.2} \\
\cmidrule{1-5}
\bottomrule
\end{tabularx}
\caption{Dimension and joint audit gap rates at annotated trading points (\%, lower is better). Joint denotes a gap in any applicable check. The denominator is each model's points containing formally annotated trades; gaps include confirmed failures and unable-to-determine labels, counted once per check per point. N/A denotes no annotated trades.}
\label{tab:2}
\end{table}

Most active models' trading-point joint gap rates fall between 60\% and 90\%, with RK exceeding O and B: evidence omissions and citation conflicts are common. Trading-point joint rates also exceed all-point rates; including non-trading windows lowers the overall proportion and obscures defects concentrated in actual trades.

Complete report retention does not guarantee accurate trade descriptions. Qwen3 235B retains valid reports at all trading points, yet descriptions disagree with account postings at 25.0\% of them, showing why report presence and action consistency require separate checks (Table~\ref{tab:3}).

\begin{table*}[!t]
\centering
\small
\renewcommand{\arraystretch}{1.12}
\setlength{\tabcolsep}{4pt}
\begin{tabularx}{\textwidth}{@{}>{\raggedright\arraybackslash}p{0.2050\textwidth}>{\hsize=1.0000\hsize \raggedleft\arraybackslash}X>{\hsize=1.0000\hsize \raggedleft\arraybackslash}X>{\hsize=1.0000\hsize \raggedleft\arraybackslash}X>{\hsize=1.0000\hsize \raggedleft\arraybackslash}X>{\hsize=1.0000\hsize \raggedleft\arraybackslash}X>{\hsize=1.0000\hsize \raggedleft\arraybackslash}X>{\hsize=1.0000\hsize \raggedleft\arraybackslash}X>{\hsize=1.0000\hsize \raggedleft\arraybackslash}X@{}}
\toprule
\morecmidrules\cmidrule{1-9}
\textbf{Model} & \multicolumn{4}{c}{\textbf{O: execution}} & \multicolumn{2}{c}{\textbf{RK: evidence}} & \multicolumn{2}{c}{\textbf{B: account / time}} \\
\cmidrule(lr){2-5}\cmidrule(lr){6-7}\cmidrule(lr){8-9}
 & \textbf{Report} & \textbf{Rpt} & \textbf{Call} & \textbf{Resp} & \textbf{R} & \textbf{K} & \textbf{D} & \textbf{Temp} \\
\midrule
\rowcolor{TableBand}
Claude Opus 4.5 & 0.0 & 2.4 & 2.4 & 2.4 & 43.9 & 43.9 & 7.3 & 2.4 \\
GLM 4.7 Flash & 0.0 & 11.3 & 17.0 & 1.9 & 43.4 & 58.5 & 30.2 & 20.8 \\
\rowcolor{TableBand}
DeepSeek V3.2 & 0.8 & 10.0 & 1.7 & 0.8 & 29.2 & 63.3 & 8.3 & 10.8 \\
Gemini 3.1 Pro & 3.6 & 11.6 & 10.7 & 8.0 & 45.5 & 57.1 & 8.0 & 13.4 \\
\rowcolor{TableBand}
Claude Haiku 4.5 & 0.0 & 11.6 & 11.6 & 11.6 & 56.5 & 43.5 & 49.0 & 7.5 \\
Qwen3 Max & 0.0 & 2.3 & 2.3 & 2.3 & 23.3 & 41.9 & 11.6 & 5.8 \\
\rowcolor{TableBand}
DeepSeek V3.1 & 0.0 & 0.0 & 0.0 & 0.0 & 32.6 & 58.7 & 0.0 & 4.3 \\
GLM 4.7 & 13.3 & 13.3 & 0.0 & 0.0 & 26.7 & 43.3 & 26.7 & 10.0 \\
\rowcolor{TableBand}
MiniMax M2.1 & 0.0 & 2.9 & 1.5 & 0.0 & 32.4 & 47.1 & 13.2 & 11.8 \\
GPT-5.1 & N/A & N/A & N/A & N/A & N/A & N/A & N/A & N/A \\
\rowcolor{TableBand}
GPT-5.2 & 0.0 & 1.2 & 5.6 & 5.6 & 13.7 & 38.5 & 0.0 & 0.0 \\
Qwen3 235B & 0.0 & 25.0 & 22.5 & 17.5 & 35.0 & 60.0 & 15.0 & 22.5 \\
\rowcolor{TableBand}
Gemini 2.5 Flash & 3.6 & 11.9 & 8.3 & 8.3 & 23.8 & 58.3 & 9.5 & 16.7 \\
\cmidrule{1-9}
\bottomrule
\end{tabularx}
\caption{Eight-check audit gap rates at annotated trading points (\%, lower is better). The denominator matches Table~\ref{tab:2}; multiple gaps in the same check at a point count once.}
\label{tab:3}
\end{table*}

R and K flag 714 trading points, with 185 triggering both (25.9\%). Most points exhibit either citation problems or key-fact omissions, indicating that the two checks cannot substitute for each other. Pooling all 1,070 annotated trading points, confirmed RK gaps and undetermined-only cases account for 63.36\% and 3.36\%, respectively, giving a merged gap rate of 66.73\% (Appendix B.1).

Account assessments also vary: Claude Haiku 4.5 has a D gap rate of 49.0\%, while low rates should be read alongside account-claim coverage (Appendix B.1). Temp gaps occur at 8.79\% of annotated trading points, reflecting explanation continuity. Formal pairs contain 70 partially coherent and 27 unable-to-determine cases, with no incoherent labels (Appendix B.5).

\subsection{Historical Trade-Filtering Replay}

To assess how audit filtering affects existing trading paths, we replace flagged trades with No Trade, retain the original timing, target quantities, and labels of the other trades, and recompute account trajectories. Retained trades execute subject to the updated cash balance and sellable holdings. We replay O, RK, B, and their combinations separately, with each combined filter formed from the union of its flagged-trade sets. Execution and replay rules are detailed in Appendix C.1.

Let \(V_0\) and \(V_T\) denote initial and final total account assets, respectively. Cumulative return is defined as:
\begin{equation}
\mathrm{CR}=\frac{V_T}{V_0}-1.
\label{eq:cumulative-return}
\end{equation}
The return change for model \(m\) under filtering condition \(\mathcal F\) is defined as:
\begin{equation}
\Delta \mathrm{CR}_{m,\mathcal F}=100\left(\mathrm{CR}_{m,\mathcal F}-\mathrm{CR}_{m,\mathrm{original}}\right).
\label{eq:return-change}
\end{equation}
Here, \(\mathrm{CR}\) is a decimal and \(\Delta \mathrm{CR}\) is in percentage points. Table~\ref{tab:4} reports equally weighted mean changes across 12 active models for individual checks, dimensions, and combinations.

\begin{table}[!t]
\centering
\small
\renewcommand{\arraystretch}{1.12}
\setlength{\tabcolsep}{2pt}

\begin{tabularx}{\columnwidth}{@{}
>{\raggedright\arraybackslash}p{0.19\columnwidth}
*{4}{>{\centering\arraybackslash}X}@{}}
\toprule
\textbf{Filter}
& \textbf{$\Delta$CR}
& \textbf{$\Delta$Sharpe}
& \textbf{$\Delta$Sortino}
& \textbf{$\Delta$Calmar} \\
\midrule

\rowcolor{TableBand}
R & \mbox{+6.66 (8)} & \mbox{+0.22 (8)}
  & \mbox{-0.02 (8)} & \mbox{+0.54 (8)} \\
K & \mbox{+9.29 (10)} & \mbox{+0.64 (9)}
  & \mbox{+0.80 (8)} & \mbox{+2.29 (9)} \\
\rowcolor{TableBand}
D & \mbox{+4.49 (10)} & \mbox{+0.34 (9)}
  & \mbox{+0.36 (9)} & \mbox{+0.56 (9)} \\
Temp & \mbox{+3.45 (9)} & \mbox{+0.46 (7)}
     & \mbox{+0.56 (8)} & \mbox{+1.04 (8)} \\
\midrule

\rowcolor{TableBand}
O & \mbox{+3.65 (11)} & \mbox{+0.12 (7)}
  & \mbox{-0.06 (7)} & \mbox{+0.11 (7)} \\
RK & \mbox{+8.86 (9)} & \mbox{+0.61 (8)}
   & \mbox{+0.82 (7)} & \mbox{+2.68 (8)} \\
\rowcolor{TableBand}
B & \mbox{+6.11 (9)} & \mbox{+0.44 (8)}
  & \mbox{+0.43 (7)} & \mbox{+0.84 (7)} \\
\midrule

O+RK & \mbox{+9.01 (9)} & \mbox{+0.57 (9)}
     & \mbox{+0.70 (7)} & \mbox{+2.12 (9)} \\
\rowcolor{TableBand}
O+B & \mbox{+7.28 (10)} & \mbox{+0.38 (8)}
    & \mbox{+0.33 (7)} & \mbox{+0.75 (7)} \\
RK+B & \mbox{+8.25 (9)} & \mbox{+0.59 (8)}
     & \mbox{+0.70 (7)} & \mbox{+2.05 (8)} \\
\rowcolor{TableAccent}
O+RK+B & \mbox{+7.65 (9)} & \mbox{+0.52 (8)}
       & \mbox{+0.52 (6)} & \mbox{+1.35 (9)} \\
\bottomrule
\end{tabularx}

\caption{Replay results for filtering. $\Delta$ is the
equally weighted mean of filtered minus original results
across 12 active models; parentheses indicate the number
of models whose metric improves or remains unchanged.
Filtering definitions appear in Appendix C.}
\label{tab:4}
\end{table}

All filters improve mean CR, with the largest gains from K and RK: 9.29 and 8.86 percentage points, respectively. All-dimension filtering yields 7.65 percentage points, improving returns for 9 models. Mean risk-adjusted return metrics also improve under RK and joint filtering (Table~\ref{tab:4}).

\textbf{Confirmed-error-only filtering.} Retaining unable-to-determine trades and removing confirmed gaps yields mean CR gains of 9.13 and 9.39 percentage points for RK and all-dimension filtering, versus 8.86 and 7.65 under the merged definition (Figure 2). Both improve returns for 11 models; complete results and removal counts appear in Appendix Table C11.

\begin{figure}[!tb]
\centering
\includegraphics[width=\columnwidth]{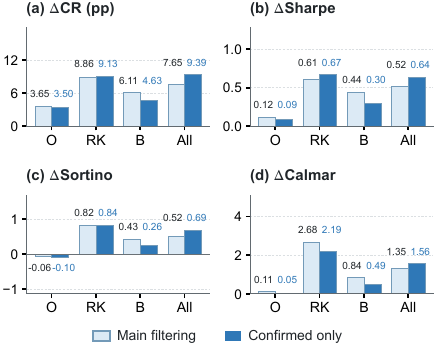}
\renewcommand{\thefigure}{2}
\caption{Main versus confirmed-error-only filtering. Mean changes across 12 active models; All denotes O+RK+B.}
\label{fig:2}
\end{figure}

\textbf{Buy/sell-matched random-filtering control.} To compare audit filtering with equal-count random deletion, we sample from each model's full-period trades, match buy/sell removal counts, and repeat independently 1,000 times with a fixed seed (Figure 3). All four CR gains exceed random means and lie within their central 95\% ranges. RK and joint filtering exceed the means by 3.55 and 2.75 percentage points; RK's Sharpe, Sortino, and Calmar gains exceed the random 97.5th percentiles. Full ranges and multiple-metric adjustments appear in Appendix C.7.

\begin{figure}[!tb]
\centering
\includegraphics[width=\columnwidth]{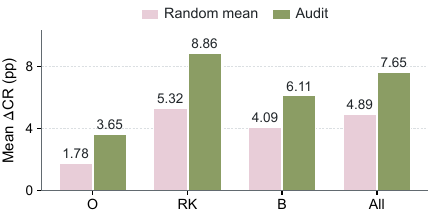}
\renewcommand{\thefigure}{3}
\caption{Audit versus random filtering (mean \ensuremath{\Delta}CR). Random bars average 1,000 replays matched by model and buy/sell counts; All denotes O+RK+B.}
\label{fig:3}
\end{figure}

At 20 basis points of slippage per side, mean CR gains from RK and all-dimension filtering are 10.17 and 9.21 percentage points, respectively; full cost-sensitivity results appear in Appendix C.6.

\begin{table}[!t]
\centering
\small
\setlength{\tabcolsep}{2pt}
\renewcommand{\arraystretch}{1.12}

\begin{tabularx}{\columnwidth}{@{}
>{\raggedright\arraybackslash}p{0.235\columnwidth}
*{5}{>{\centering\arraybackslash}X}@{}}
\toprule
\textbf{GPT-6 Sol}
& \textbf{O} & \textbf{RK} & \textbf{B}
& \textbf{Joint}
& \textbf{\shortstack{Trading\\points}} \\
\midrule
\rowcolor{TableBand}
Gap rate (\%)
& 0.00 & 42.71 & 0.00 & 42.71 & 96 \\
\end{tabularx}

\par\nointerlineskip
\begin{tabularx}{\columnwidth}{@{}
>{\raggedright\arraybackslash}p{0.235\columnwidth}
*{4}{>{\centering\arraybackslash}X}@{}}
\midrule
\rowcolor{TableSection}
\textbf{Replay}
& \textbf{CR} & \textbf{Sharpe}
& \textbf{Sortino} & \textbf{Calmar} \\
\rowcolor{TableBand}
Original & 4.85 & 1.02 & 1.88 & 1.94 \\
RK Filter & 38.28 & 2.86 & 5.20 & 12.84 \\
\rowcolor{TableBand}
Change & +33.43 & +1.84 & +3.33 & +10.90 \\
\bottomrule
\end{tabularx}

\caption{Audit and replay results for GPT-6 Sol.
Gap rates use 96 trading points as the denominator;
all-point results appear in Appendix D.3.
Replay compares the original and RK-filtered accounts.}
\label{tab:5}
\end{table}

\subsection{Historical Case Reruns}

To observe defects after regenerating decisions, we test DeepSeek V3.2, GLM 4.7 Flash, GPT-6 Astra (max), and GPT-5.6 Sol on ten historical snapshots each. Each case is run independently after restoring contemporaneous information and the pre-decision account state. Independent snapshots exclude Temp (Figure 4).

\begin{figure}[!tb]
\centering
\includegraphics[width=\columnwidth]{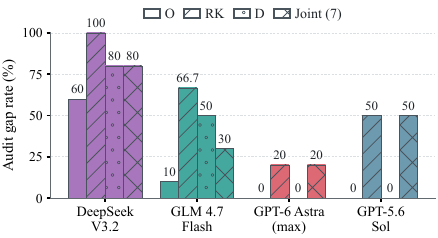}
\renewcommand{\thefigure}{4}
\caption{Audit gap rates for ten reruns under each configuration (\%, \ensuremath{\downarrow}). Colors denote model and run configuration; hatching denotes metrics. Joint includes O, R, K, and D; budgets and audit-guidance conditions appear in Appendix D.2.}
\label{fig:4}
\end{figure}

Decision defects remain observable after rerunning. DeepSeek V3.2 has confirmed gaps in eight of ten cases, including all six trading points; GLM 4.7 Flash has three, comprising two key-fact omissions and one report-action mismatch. Neither GPT model has confirmed execution or account-fact gaps, but Astra and Sol have evidence gaps in two and five cases, respectively, with the remainder passing the seven-check joint audit. Cases under different configurations show that execution consistency and evidence sufficiency require separate checks.

\subsection{Continuous Trading by a New Model}

To examine the audit framework's applicability to new-generation base models, we evaluate GPT-6 Sol (with medium reasoning) on the same historical snapshots. The model generates 96 active trades across 246 decision windows; its audit gap rates and filtering replay results appear in Table~\ref{tab:5}.

New models may have encountered public market information from the evaluation period, limiting interpretation of historical investment performance. Auditing archived decision-time materials, account states, and execution records can still assess evidence use, account understanding, and trade execution.

No gaps are identified in GPT-6 Sol's execution or account-fact checks, but 41 trades have evidence audit gaps. RK filtering increases CR by 33.43 percentage points and also improves risk-adjusted returns (Table~\ref{tab:5}).

\FloatBarrier

\section{Discussion and Conclusion}

\textbf{Discussion.} The new model has no confirmed execution or account-fact gaps, yet still omits key evidence, suggesting a need for error-specific feedback. Reliable execution therefore needs to be accompanied by adequate consideration of decision-relevant evidence. AStockArena links execution discrepancies, evidence omissions, and account-assessment errors to trades and source materials as a basis for model correction and training. Future work can test whether revising rationales in response to this feedback corrects identified problems and how it affects subsequent trading decisions.

\textbf{Conclusion.} AStockArena audits continuous financial trading through structured-record verification and consensus semantic annotation, aligning reports, execution records, and external evidence with account changes to verify actions, rationales, and account assessments. Historical trade-filtering replay examines the relationship between audit gaps and portfolio performance. Together, these analyses provide a verifiable basis for evaluating and improving subsequent models' decision processes and building more reliable trading agents.

\FloatBarrier

\label{main_end}

\section*{Limitations}

\label{sec:limitations}

This study has not integrated audit feedback into a closed-loop trading system or evaluated subsequent model decisions following audit interventions. Due to running and annotation costs, each model follows a single continuous trajectory, and variability across repeated runs has not been characterized. The evaluation covers ten STAR Market stocks over a single period. The stock pool is sized to control the information load within each decision window and reduce the risk of dispersed attention when analyzing multiple stocks in parallel. Execution is immediate at snapshot prices, without partial fills, queuing, or latency. For models added later, timestamp constraints on tool inputs cannot rule out prior exposure to relevant historical information during training, which may affect trading rationales and historical investment performance.

\FloatBarrier

\nocite{*}

\bibliography{references}

@misc{hu2026receipts,
 archiveprefix = {arXiv},
 author = {Miaobo Hu and Shuhao Hu and Xiaobo Guo and Xin Wang and Bokun Wang and Yina Sa and Daren Zha and Jun Xiao},
 eprint = {2610.00327},
 title = {Actions with Receipts: Jointly Binding Claims, Evidence, and Execution for Replayable Tool-Agent Auditing},
 url = {https://arxiv.org/abs/2610.00327},
 year = {2026}
}

@misc{kim2026ledger,
 archiveprefix = {arXiv},
 author = {Daehong Kim and Haichao Miao and Shusen Liu},
 eprint = {2608.18398},
 title = {{LEDGER}: Claim-to-Evidence Trace Graphs for Auditing {LLM} Agents},
 url = {https://arxiv.org/abs/2608.18398},
 year = {2026}
}

@article{liu2024lostmiddle,
 author = {Nelson F. Liu and Kevin Lin and John Hewitt and Ashwin Paranjape and Michele Bevilacqua and Fabio Petroni and Percy Liang},
 doi = {10.1162/tacl_a_00638},
 journal = {Transactions of the Association for Computational Linguistics},
 pages = {157--173},
 title = {Lost in the Middle: How Language Models Use Long Contexts},
 url = {https://aclanthology.org/2024.tacl-1.9/},
 volume = {12},
 year = {2024}
}

@misc{liu2026attribution,
 archiveprefix = {arXiv},
 author = {Yifan Liu and Praveen Venkateswaran and Abdulhamid Adebayo and Dong Wang},
 eprint = {2609.33676},
 title = {Auditing Agent Actions through Query-Conditioned Attribution},
 url = {https://arxiv.org/abs/2609.33676},
 year = {2026}
}

@misc{rhodes2026poe,
 archiveprefix = {arXiv},
 author = {James Rhodes and George Kang},
 eprint = {2607.05397},
 title = {Proof of Execution: Runtime Verification for Governed {AI} Agent Actions},
 url = {https://arxiv.org/abs/2607.05397},
 year = {2026}
}

@misc{cao2026pae,
 archiveprefix = {arXiv},
 author = {Hongliu Cao and Ilias Driouich and Eoin Thomas},
 eprint = {2603.03116},
 primaryclass = {cs.AI},
 title = {Beyond Task Completion: Revealing Corrupt Success in LLM Agents through Procedure-Aware Evaluation},
 url = {https://arxiv.org/abs/2603.03116},
 year = {2026}
}

@inproceedings{chen2021finqa,
 address = {Online and Punta Cana, Dominican Republic},
 author = {Chen, Zhiyu  and
Chen, Wenhu  and
Smiley, Charese  and
Shah, Sameena  and
Borova, Iana  and
Langdon, Dylan  and
Moussa, Reema  and
Beane, Matt  and
Huang, Ting-Hao  and
Routledge, Bryan  and
Wang, William Yang},
 booktitle = {Proceedings of the 2021 Conference on Empirical Methods in Natural Language Processing},
 doi = {10.18653/v1/2021.emnlp-main.300},
 editor = {Moens, Marie-Francine  and
Huang, Xuanjing  and
Specia, Lucia  and
Yih, Scott Wen-tau},
 month = {nov},
 pages = {3697--3711},
 publisher = {Association for Computational Linguistics},
 title = {{F}in{QA}: A Dataset of Numerical Reasoning over Financial Data},
 url = {https://aclanthology.org/2021.emnlp-main.300/},
 year = {2021}
}

@inproceedings{chen2022convfinqa,
 address = {Abu Dhabi, United Arab Emirates},
 author = {Chen, Zhiyu  and
Li, Shiyang  and
Smiley, Charese  and
Ma, Zhiqiang  and
Shah, Sameena  and
Wang, William Yang},
 booktitle = {Proceedings of the 2022 Conference on Empirical Methods in Natural Language Processing},
 doi = {10.18653/v1/2022.emnlp-main.421},
 editor = {Goldberg, Yoav  and
Kozareva, Zornitsa  and
Zhang, Yue},
 month = {dec},
 pages = {6279--6292},
 publisher = {Association for Computational Linguistics},
 title = {{C}onv{F}in{QA}: Exploring the Chain of Numerical Reasoning in Conversational Finance Question Answering},
 url = {https://aclanthology.org/2022.emnlp-main.421/},
 year = {2022}
}

@misc{chen2025stockbench,
 archiveprefix = {arXiv},
 author = {Yanxu Chen and Zijun Yao and Yantao Liu and Amy Xin and Jin Ye and Jianing Yu and Lei Hou and Juanzi Li},
 eprint = {2510.02209},
 primaryclass = {cs.LG},
 title = {StockBench: Can LLM Agents Trade Stocks Profitably In Real-world Markets?},
 url = {https://arxiv.org/abs/2510.02209},
 year = {2025}
}

@misc{fan2025aitrader,
 archiveprefix = {arXiv},
 author = {Tianyu Fan and Yuhao Yang and Yangqin Jiang and Yifei Zhang and Yuxuan Chen and Chao Huang},
 eprint = {2512.10971},
 primaryclass = {q-fin.CP},
 title = {AI-Trader: Benchmarking Autonomous Agents in Real-Time Financial Markets},
 url = {https://arxiv.org/abs/2512.10971},
 year = {2025}
}

@inproceedings{jacovi2020faithfulness,
 address = {Online},
 author = {Jacovi, Alon  and
Goldberg, Yoav},
 booktitle = {Proceedings of the 58th Annual Meeting of the Association for Computational Linguistics},
 doi = {10.18653/v1/2020.acl-main.386},
 editor = {Jurafsky, Dan  and
Chai, Joyce  and
Schluter, Natalie  and
Tetreault, Joel},
 month = {July},
 pages = {4198--4205},
 publisher = {Association for Computational Linguistics},
 title = {Towards Faithfully Interpretable {NLP} Systems: How Should We Define and Evaluate Faithfulness?},
 url = {https://aclanthology.org/2020.acl-main.386/},
 year = {2020}
}

@inproceedings{li2025investorbench,
 address = {Vienna, Austria},
 author = {Li, Haohang  and
Cao, Yupeng  and
Yu, Yangyang  and
Javaji, Shashidhar Reddy  and
Deng, Zhiyang  and
He, Yueru  and
Jiang, Yuechen  and
Zhu, Zining  and
Subbalakshmi, K.P.  and
Huang, Jimin  and
Qian, Lingfei  and
Peng, Xueqing  and
Suchow, Jordan W.  and
Xie, Qianqian},
 booktitle = {Proceedings of the 63rd Annual Meeting of the Association for Computational Linguistics (Volume 1: Long Papers)},
 doi = {10.18653/v1/2025.acl-long.126},
 editor = {Che, Wanxiang  and
Nabende, Joyce  and
Shutova, Ekaterina  and
Pilehvar, Mohammad Taher},
 isbn = {979-8-89176-251-0},
 month = {jul},
 pages = {2509--2525},
 publisher = {Association for Computational Linguistics},
 title = {{INVESTORBENCH}: A Benchmark for Financial Decision-Making Tasks with {LLM}-based Agent},
 url = {https://aclanthology.org/2025.acl-long.126/},
 year = {2025}
}

@inproceedings{liu2025findabench,
 address = {Abu Dhabi, UAE},
 author = {Liu, Shu  and
Zhao, Shangqing  and
Jia, Chenghao  and
Zhuang, Xinlin  and
Long, Zhaoguang  and
Zhou, Jie  and
Zhou, Aimin  and
Lan, Man  and
Chong, Yang},
 booktitle = {Proceedings of the 31st International Conference on Computational Linguistics},
 editor = {Rambow, Owen  and
Wanner, Leo  and
Apidianaki, Marianna  and
Al-Khalifa, Hend  and
Eugenio, Barbara Di  and
Schockaert, Steven},
 month = {jan},
 pages = {710--725},
 publisher = {Association for Computational Linguistics},
 title = {{F}in{DAB}ench: Benchmarking Financial Data Analysis Ability of Large Language Models},
 url = {https://aclanthology.org/2025.coling-main.48/},
 year = {2025}
}

@misc{qu2026clqt,
 archiveprefix = {arXiv},
 author = {Bo Qu and Mingguang Chen},
 eprint = {2606.29771},
 primaryclass = {cs.AI},
 title = {CLQT: A Closed-Loop, Cost-Aware, Strategy-Consistent Benchmark for Diagnostic Evaluation of LLM Portfolio-Management Agents},
 url = {https://arxiv.org/abs/2606.29771},
 year = {2026}
}

@article{artstein2008agreement,
 author = {Ron Artstein and Massimo Poesio},
 title = {Survey Article: Inter-Coder Agreement for Computational Linguistics},
 journal = {Computational Linguistics},
 volume = {34},
 number = {4},
 pages = {555--596},
 doi = {10.1162/coli.07-034-R2},
 url = {https://aclanthology.org/J08-4004/},
 year = {2008}
}

@misc{tang2026sagefin,
 author = {Rui Tang and Qiangqiang Liu and Yichi Zhang and Youwei Yang and Xi Chen and Chen Dong},
 title = {Context Is Not Authority: Structured Runtime Governance for Financial Market Agents},
 archiveprefix = {arXiv},
 eprint = {2608.09025},
 primaryclass = {cs.AI},
 url = {https://arxiv.org/abs/2608.09025},
 year = {2026}
}

@inproceedings{sun2025finresearchbench,
 address = {New York, NY, USA},
 author = {Sun, Rui and Bai, Zuo and Zhang, Wentao and Zhang, Yuxiang and Zhao, Li and Sun, Shan and Qiu, Zhengwen},
 booktitle = {Proceedings of the 6th ACM International Conference on AI in Finance},
 doi = {10.1145/3768292.3770364},
 isbn = {9798400722202},
 location = { },
 numpages = {9},
 pages = {656--664},
 publisher = {Association for Computing Machinery},
 series = {ICAIF '25},
 title = {FinResearchBench: A Logic Tree based Agent-as-a-Judge Evaluation Framework for Financial Research Agents},
 url = {https://doi.org/10.1145/3768292.3770364},
 year = {2025}
}

@inproceedings{turpin2023unfaithful,
 author = {Miles Turpin and Julian Michael and Ethan Perez and Samuel R. Bowman},
 booktitle = {Thirty-seventh Conference on Neural Information Processing Systems},
 title = {Language Models Don't Always Say What They Think: Unfaithful Explanations in Chain-of-Thought Prompting},
 url = {https://openreview.net/forum?id=bzs4uPLXvi},
 year = {2023}
}

@inproceedings{xie2024finben,
 author = {Qianqian Xie and Weiguang Han and Zhengyu Chen and Ruoyu Xiang and Xiao Zhang and Yueru He and Mengxi Xiao and Dong Li and Yongfu Dai and Duanyu Feng and Yijing Xu and Haoqiang Kang and Ziyan Kuang and Chenhan Yuan and Kailai Yang and Zheheng Luo and Tianlin Zhang and Zhiwei Liu and Guojun Xiong and Zhiyang Deng and Yuechen Jiang and Zhiyuan Yao and Haohang Li and Yangyang Yu and Gang Hu and Jiajia Huang and Xiao-Yang Liu and Alejandro Lopez-Lira and Benyou Wang and Yanzhao Lai and Hao Wang and Min Peng and Sophia Ananiadou and Jimin Huang},
 booktitle = {The Thirty-eight Conference on Neural Information Processing Systems Datasets and Benchmarks Track},
 title = {FinBen: A Holistic Financial Benchmark for Large Language Models},
 url = {https://openreview.net/forum?id=loDHZstVP6},
 year = {2024}
}

@misc{yan2025tradetrap,
 archiveprefix = {arXiv},
 author = {Lewen Yan and Jilin Mei and Tianyi Zhou and Lige Huang and Jie Zhang and Dongrui Liu and Jing Shao},
 eprint = {2512.02261},
 primaryclass = {cs.AI},
 title = {TradeTrap: Are LLM-based Trading Agents Truly Reliable and Faithful?},
 url = {https://arxiv.org/abs/2512.02261},
 year = {2025}
}

@misc{ye2026claweval,
 archiveprefix = {arXiv},
 author = {Bowen Ye and Rang Li and Qibin Yang and Yuanxin Liu and Linli Yao and Hanglong Lv and Zhihui Xie and Chenxin An and Lei Li and Lingpeng Kong and Qi Liu and Zhifang Sui and Tong Yang},
 eprint = {2604.06132},
 primaryclass = {cs.AI},
 title = {Claw-Eval: Towards Trustworthy Evaluation of Autonomous Agents},
 url = {https://arxiv.org/abs/2604.06132},
 year = {2026}
}

@misc{yu2025livetradebench,
 archiveprefix = {arXiv},
 author = {Haofei Yu and Fenghai Li and Jiaxuan You},
 eprint = {2511.03628},
 primaryclass = {q-fin.TR},
 title = {LiveTradeBench: Seeking Real-World Alpha with Large Language Models},
 url = {https://arxiv.org/abs/2511.03628},
 year = {2025}
}

@inproceedings{zhang2024finagent,
 author = {Wentao Zhang and Lingxuan Zhao and Haochong Xia and Shuo Sun and Jiaze Sun and Molei Qin and Xinyi Li and Yuqing Zhao and Yilei Zhao and Xinyu Cai and Longtao Zheng and Xinrun Wang and Bo An},
 booktitle = {Proceedings of the 30th ACM SIGKDD Conference on Knowledge Discovery and Data Mining},
 doi = {10.1145/3637528.3671801},
 pages = {4314--4325},
 publisher = {Association for Computing Machinery},
 title = {A Multimodal Foundation Agent for Financial Trading: Tool-Augmented, Diversified, and Generalist},
 url = {https://doi.org/10.1145/3637528.3671801},
 year = {2024}
}

@inproceedings{shah-etal-2022-flue,
    title = "When {FLUE} Meets {FLANG}: Benchmarks and Large Pretrained Language Model for Financial Domain",
    author = "Shah, Raj  and
      Chawla, Kunal  and
      Eidnani, Dheeraj  and
      Shah, Agam  and
      Du, Wendi  and
      Chava, Sudheer  and
      Raman, Natraj  and
      Smiley, Charese  and
      Chen, Jiaao  and
      Yang, Diyi",
    editor = "Goldberg, Yoav  and
      Kozareva, Zornitsa  and
      Zhang, Yue",
    booktitle = "Proceedings of the 2022 Conference on Empirical Methods in Natural Language Processing",
    month = dec,
    year = "2022",
    address = "Abu Dhabi, United Arab Emirates",
    publisher = "Association for Computational Linguistics",
    url = "https://aclanthology.org/2022.emnlp-main.148/",
    doi = "10.18653/v1/2022.emnlp-main.148",
    pages = "2322--2335"
}

@inproceedings{yang-etal-2023-measuring,
    title = "Measuring Consistency in Text-based Financial Forecasting Models",
    author = "Yang, Linyi  and
      Ma, Yingpeng  and
      Zhang, Yue",
    editor = "Rogers, Anna  and
      Boyd-Graber, Jordan  and
      Okazaki, Naoaki",
    booktitle = "Proceedings of the 61st Annual Meeting of the Association for Computational Linguistics (Volume 1: Long Papers)",
    month = jul,
    year = "2023",
    address = "Toronto, Canada",
    publisher = "Association for Computational Linguistics",
    url = "https://aclanthology.org/2023.acl-long.769/",
    doi = "10.18653/v1/2023.acl-long.769",
    pages = "13751--13765"
}

@inproceedings{li-etal-2024-cryptotrade,
    title = "{C}rypto{T}rade: A Reflective {LLM}-based Agent to Guide Zero-shot Cryptocurrency Trading",
    author = "Li, Yuan  and
      Luo, Bingqiao  and
      Wang, Qian  and
      Chen, Nuo  and
      Liu, Xu  and
      He, Bingsheng",
    editor = "Al-Onaizan, Yaser  and
      Bansal, Mohit  and
      Chen, Yun-Nung",
    booktitle = "Proceedings of the 2024 Conference on Empirical Methods in Natural Language Processing",
    month = nov,
    year = "2024",
    address = "Miami, Florida, USA",
    publisher = "Association for Computational Linguistics",
    url = "https://aclanthology.org/2024.emnlp-main.63/",
    doi = "10.18653/v1/2024.emnlp-main.63",
    pages = "1094--1106"
}

@inproceedings{bhatia-etal-2024-fintral,
    title = "{F}in{T}ral: A Family of {GPT}-4 Level Multimodal Financial Large Language Models",
    author = "Bhatia, Gagan  and
      Nagoudi, El Moatez Billah  and
      Cavusoglu, Hasan  and
      Abdul-Mageed, Muhammad",
    editor = "Ku, Lun-Wei  and
      Martins, Andre  and
      Srikumar, Vivek",
    booktitle = "Findings of the Association for Computational Linguistics: ACL 2024",
    month = aug,
    year = "2024",
    address = "Bangkok, Thailand",
    publisher = "Association for Computational Linguistics",
    url = "https://aclanthology.org/2024.findings-acl.774/",
    doi = "10.18653/v1/2024.findings-acl.774",
    pages = "13064--13087"
}

@inproceedings{son-etal-2025-finkrx,
    title = "{FINKRX}: Establishing Best Practices for {K}orean Financial {NLP}",
    author = "Son, Guijin  and
      Ko, Hyunwoo  and
      Jung, Hanearl  and
      Hwang, Chami",
    editor = "Rehm, Georg  and
      Li, Yunyao",
    booktitle = "Proceedings of the 63rd Annual Meeting of the Association for Computational Linguistics (Volume 6: Industry Track)",
    month = jul,
    year = "2025",
    address = "Vienna, Austria",
    publisher = "Association for Computational Linguistics",
    url = "https://aclanthology.org/2025.acl-industry.81/",
    doi = "10.18653/v1/2025.acl-industry.81",
    pages = "1161--1174",
    ISBN = "979-8-89176-288-6"
}

@inproceedings{tang-etal-2025-financereasoning,
    title = "{F}inance{R}easoning: Benchmarking Financial Numerical Reasoning More Credible, Comprehensive and Challenging",
    author = "Tang, Zichen  and
      E, Haihong  and
      Ma, Ziyan  and
      He, Haoyang  and
      Liu, Jiacheng  and
      Yang, Zhongjun  and
      Rong, Zihua  and
      Li, Rongjin  and
      Ji, Kun  and
      Huang, Qing  and
      Hu, Xinyang  and
      Liu, Yang  and
      Zheng, Qianhe",
    editor = "Che, Wanxiang  and
      Nabende, Joyce  and
      Shutova, Ekaterina  and
      Pilehvar, Mohammad Taher",
    booktitle = "Proceedings of the 63rd Annual Meeting of the Association for Computational Linguistics (Volume 1: Long Papers)",
    month = jul,
    year = "2025",
    address = "Vienna, Austria",
    publisher = "Association for Computational Linguistics",
    url = "https://aclanthology.org/2025.acl-long.766/",
    doi = "10.18653/v1/2025.acl-long.766",
    pages = "15721--15749",
    ISBN = "979-8-89176-251-0"
}

@inproceedings{tang-yang-2025-finmteb,
    title = "{F}in{MTEB}: Finance Massive Text Embedding Benchmark",
    author = "Tang, Yixuan  and
      Yang, Yi",
    editor = "Christodoulopoulos, Christos  and
      Chakraborty, Tanmoy  and
      Rose, Carolyn  and
      Peng, Violet",
    booktitle = "Proceedings of the 2025 Conference on Empirical Methods in Natural Language Processing",
    month = nov,
    year = "2025",
    address = "Suzhou, China",
    publisher = "Association for Computational Linguistics",
    url = "https://aclanthology.org/2025.emnlp-main.179/",
    doi = "10.18653/v1/2025.emnlp-main.179",
    pages = "3620--3638",
    ISBN = "979-8-89176-332-6"
}

@inproceedings{zhao-etal-2025-finragbench,
    title = "{F}in{RAGB}ench-{V}: A Benchmark for Multimodal {RAG} with Visual Citation in the Financial Domain",
    author = "Zhao, Suifeng  and
      Jin, Zhuoran  and
      Li, Sujian  and
      Gao, Jun",
    editor = "Christodoulopoulos, Christos  and
      Chakraborty, Tanmoy  and
      Rose, Carolyn  and
      Peng, Violet",
    booktitle = "Proceedings of the 2025 Conference on Empirical Methods in Natural Language Processing",
    month = nov,
    year = "2025",
    address = "Suzhou, China",
    publisher = "Association for Computational Linguistics",
    url = "https://aclanthology.org/2025.emnlp-main.211/",
    doi = "10.18653/v1/2025.emnlp-main.211",
    pages = "4215--4249",
    ISBN = "979-8-89176-332-6"
}

@inproceedings{guo-etal-2025-fineval,
    title = "{F}in{E}val: A {C}hinese Financial Domain Knowledge Evaluation Benchmark for Large Language Models",
    author = "Guo, Xin  and
      Xia, Haotian  and
      Liu, Zhaowei  and
      Cao, Hanyang  and
      Yang, Zhi  and
      Liu, Zhiqiang  and
      Wang, Sizhe  and
      Niu, Jinyi  and
      Wang, Chuqi  and
      Wang, Yanhui  and
      Liang, Xiaolong  and
      Huang, Xiaoming  and
      Zhu, Bing  and
      Wei, Zhongyu  and
      Chen, Yun  and
      Shen, Weining  and
      Zhang, Liwen",
    editor = "Chiruzzo, Luis  and
      Ritter, Alan  and
      Wang, Lu",
    booktitle = "Proceedings of the 2025 Conference of the Nations of the Americas Chapter of the Association for Computational Linguistics: Human Language Technologies (Volume 1: Long Papers)",
    month = apr,
    year = "2025",
    address = "Albuquerque, New Mexico",
    publisher = "Association for Computational Linguistics",
    url = "https://aclanthology.org/2025.naacl-long.318/",
    doi = "10.18653/v1/2025.naacl-long.318",
    pages = "6258--6292",
    ISBN = "979-8-89176-189-6"
}

@inproceedings{choi-etal-2026-finharmbench,
    title = "{F}in{H}arm{B}ench: Financial Jailbreak Benchmark and Unsupervised Safety Fine-Tuning via Refusal Steering Distillation",
    author = "Choi, Yubin  and
      Yang, Yujin  and
      Kim, Subin  and
      Ham, Seokil  and
      Cho, Seungju  and
      Son, Jungmin  and
      Kwak, Youngjun  and
      Kim, Changick",
    editor = "Li, Yunyao  and
      Rehm, Georg  and
      Tu, Mei",
    booktitle = "Proceedings of the 64th Annual Meeting of the {A}ssociation for {C}omputational {L}inguistics (Volume 6: Industry Track)",
    month = jul,
    year = "2026",
    address = "San Diego, California, USA",
    publisher = "Association for Computational Linguistics",
    url = "https://aclanthology.org/2026.acl-industry.117/",
    doi = "10.18653/v1/2026.acl-industry.117",
    pages = "1714--1726",
    ISBN = "979-8-89176-394-4"
}

@inproceedings{zheng-etal-2026-market,
    title = "Market-Bench: Benchmarking Large Language Models on Economic and Trade Competition",
    author = "Zheng, Yushuo  and
      Duan, Huiyu  and
      Zhang, Zicheng  and
      Zhu, Yucheng  and
      Min, Xiongkuo  and
      Zhai, Guangtao",
    editor = "Liakata, Maria  and
      Moreira, Viviane P.  and
      Zhang, Jiajun  and
      Jurgens, David",
    booktitle = "Proceedings of the 64th Annual Meeting of the {A}ssociation for {C}omputational {L}inguistics (Volume 1: Long Papers)",
    month = jul,
    year = "2026",
    address = "San Diego, California, United States",
    publisher = "Association for Computational Linguistics",
    url = "https://aclanthology.org/2026.acl-long.1853/",
    doi = "10.18653/v1/2026.acl-long.1853",
    pages = "39893--39906",
    ISBN = "979-8-89176-390-6"
}

@inproceedings{xie-etal-2026-finchain,
    title = "{F}in{C}hain: A Symbolic Benchmark for Verifiable Chain-of-Thought Financial Reasoning",
    author = "Xie, Zhuohan  and
      Orel, Daniil  and
      Thareja, Rushil  and
      Sahnan, Dhruv  and
      Madmoun, Hachem  and
      Zhang, Fan  and
      Banerjee, Debopriyo  and
      Georgiev, Georgi Nenkov  and
      Peng, Xueqing  and
      Qian, Lingfei  and
      Huang, Jimin  and
      Su, Jinyan  and
      Singh, Aaryamonvikram  and
      Xing, Rui  and
      Elbadry, Rania  and
      Xu, Chen  and
      Li, Haonan  and
      Koto, Fajri  and
      Koychev, Ivan  and
      Chakraborty, Tanmoy  and
      Wang, Yuxia  and
      Lahlou, Salem  and
      Stoyanov, Veselin  and
      Ananiadou, Sophia  and
      Nakov, Preslav",
    editor = "Liakata, Maria  and
      Moreira, Viviane P.  and
      Zhang, Jiajun  and
      Jurgens, David",
    booktitle = "Proceedings of the 64th Annual Meeting of the {A}ssociation for {C}omputational {L}inguistics (Volume 1: Long Papers)",
    month = jul,
    year = "2026",
    address = "San Diego, California, United States",
    publisher = "Association for Computational Linguistics",
    url = "https://aclanthology.org/2026.acl-long.662/",
    doi = "10.18653/v1/2026.acl-long.662",
    pages = "14529--14553",
    ISBN = "979-8-89176-390-6"
}

@inproceedings{peng-etal-2026-multifinben,
    title = "{M}ulti{F}in{B}en: Benchmarking Large Language Models for Multilingual and Multimodal Financial Application",
    author = "Peng, Xueqing  and
      Qian, Lingfei  and
      Wang, Yan  and
      Xiang, Ruoyu  and
      He, Yueru  and
      Ren, Yang  and
      Jiang, Mingyang  and
      Zhang, Vincent Jim  and
      Guo, Yuqing  and
      Zhao, Jeff  and
      He, Huan  and
      Han, Yi  and
      Feng, Yun  and
      Jiang, Yuechen  and
      Cao, Yupeng  and
      Li, Haohang  and
      Yu, Yangyang  and
      Wang, Xiaoyu  and
      Gao, Penglei  and
      Lin, Shengyuan  and
      Wang, Keyi  and
      Yang, Shanshan  and
      Zhao, Yilun  and
      Liu, Zhiwei  and
      Lu, Peng  and
      Huang, Jerry  and
      Wang, Suyuchen  and
      Papadopoulos, Triantafillos  and
      Giannouris, Polydoros  and
      Soufleri, Efstathia  and
      Chen, Nuo  and
      Deng, Zhiyang  and
      Fu, Heming  and
      Zhao, Yijia  and
      Lin, Mingquan  and
      Qiu, Meikang  and
      Smith, Kaleb E  and
      Cohan, Arman  and
      Liu, Xiao-Yang  and
      Huang, Jimin  and
      Xiong, Guojun  and
      Lopez-Lira, Alejandro  and
      Chen, Xi  and
      Tsujii, Junichi  and
      Nie, Jian-Yun  and
      Ananiadou, Sophia  and
      Xie, Qianqian",
    editor = "Liakata, Maria  and
      Moreira, Viviane P.  and
      Zhang, Jiajun  and
      Jurgens, David",
    booktitle = "Proceedings of the 64th Annual Meeting of the {A}ssociation for {C}omputational {L}inguistics (Volume 1: Long Papers)",
    month = jul,
    year = "2026",
    address = "San Diego, California, United States",
    publisher = "Association for Computational Linguistics",
    url = "https://aclanthology.org/2026.acl-long.770/",
    doi = "10.18653/v1/2026.acl-long.770",
    pages = "16934--16963",
    ISBN = "979-8-89176-390-6"
}

@inproceedings{zhou-etal-2026-fincards,
    title = "{F}in{CARDS}: Card-Based Analyst Reranking for Financial Document Question Answering",
    author = "Zhou, Yixi  and
      Zhang, Fan  and
      Chen, YU  and
      Zhang, Haipeng  and
      Nakov, Preslav  and
      Xie, Zhuohan",
    editor = "Liakata, Maria  and
      Moreira, Viviane P.  and
      Zhang, Jiajun  and
      Jurgens, David",
    booktitle = "Findings of the {A}ssociation for {C}omputational {L}inguistics: {ACL} 2026",
    month = jul,
    year = "2026",
    address = "San Diego, California, United States",
    publisher = "Association for Computational Linguistics",
    url = "https://aclanthology.org/2026.findings-acl.1244/",
    doi = "10.18653/v1/2026.findings-acl.1244",
    pages = "24836--24852",
    ISBN = "979-8-89176-395-1"
}

@InProceedings{pmlr-v306-kong26i,
  title = 	 {Position: Evaluating {LLM}s in Finance Requires Explicit Bias Consideration},
  author =       {Kong, Yaxuan and Lee, Hoyoung and Hwang, Yoontae and Lopez-Lira, Alejandro and Levy, Bradford and Mehta, Dhagash and Wen, Qingsong and Choi, Chanyeol and Lee, Yongjae and Zohren, Stefan},
  booktitle = 	 {Proceedings of the 43rd International Conference on Machine Learning},
  pages = 	 {171089--171110},
  year = 	 {2026},
  editor = 	 {Zhang, Tong and Dudik, Miroslav and Jaggi, Martin and Agarwal, Alekh and Li, Sharon and Schuurmans, Dale and Zhu, Jerry and Berkenkamp, Felix and Dong, Hanze and Bietti, Alberto},
  volume = 	 {306},
  series = 	 {Proceedings of Machine Learning Research},
  month = 	 {06--11 Jul},
  publisher =    {PMLR},
  url = 	 {https://proceedings.mlr.press/v306/kong26i.html},
}

@InProceedings{pmlr-v306-guo26ah,
  title = 	 {{B}iz{F}in{B}ench.v2: Towards Reliable {LLM}s in Finance via Real-User Data and {O}ffline/{O}nline Bilingual Evaluation},
  author =       {Guo, Xin and Zhang, Rongjunchen and Lu, Guilong and Guo, Xuntao and Shuai, Jia and Yang, Zhi and Zhang, Liwen},
  booktitle = 	 {Proceedings of the 43rd International Conference on Machine Learning},
  pages = 	 {38642--38676},
  year = 	 {2026},
  editor = 	 {Zhang, Tong and Dudik, Miroslav and Jaggi, Martin and Agarwal, Alekh and Li, Sharon and Schuurmans, Dale and Zhu, Jerry and Berkenkamp, Felix and Dong, Hanze and Bietti, Alberto},
  volume = 	 {306},
  series = 	 {Proceedings of Machine Learning Research},
  month = 	 {06--11 Jul},
  publisher =    {PMLR},
  url = 	 {https://proceedings.mlr.press/v306/guo26ah.html},
}

@inproceedings{NEURIPS2025_f37b0e09,
 author = {Li, Changlun and SHI, Yao and Wang, Chen and Duan, Qiqi and RUAN, Runke and Huang, Weijie and Long, Haonan and Huang, Lijun and Tang, Nan and Luo, Yuyu},
 booktitle = {Advances in Neural Information Processing Systems},
 doi = {10.52202/085713-5554},
 editor = {D. Belgrave and C. Zhang and H. Lin and R. Pascanu and P. Koniusz and M. Ghassemi and N. Chen},
 pages = {},
 publisher = {Curran Associates, Inc.},
 title = {Time Travel is Cheating: Going Live with DeepFund for Real-Time Fund Investment Benchmarking},
 url = {https://proceedings.neurips.cc/paper_files/paper/2025/file/f37b0e09b90e9a6833aacf5768362b54-Paper-Datasets_and_Benchmarks_Track.pdf},
 volume = {38, Main Conference},
 year = {2025}
}

@inproceedings{ICLR2026_0e22eef3,
 author = {Sugiura, Issa and Ishida, Takashi and Makino, Taro and Tazuke, Chieko and Nakagawa, Takanori and Nakago, Kosuke and Ha, David},
 booktitle = {International Conference on Learning Representations},
 editor = {C. Vondrick and B. Hariharan and C. Raffel and L. Pinto and D. Yang and A. Faust},
 pages = {8286--8303},
 title = {EDINET-Bench: Evaluating LLMs on Complex Financial Tasks using Japanese Financial Statements},
 url = {https://proceedings.iclr.cc/paper_files/paper/2026/file/0e22eef358cd1d2bdb227e6b832cc775-Paper-Conference.pdf},
 volume = {2026},
 year = {2026}
}

@inproceedings{NEURIPS2022_0bf54b80,
 author = {Liu, Xiao-Yang and Xia, Ziyi and Rui, Jingyang and Gao, Jiechao and Yang, Hongyang and Zhu, Ming and Wang, Christina and Wang, Zhaoran and Guo, Jian},
 booktitle = {Advances in Neural Information Processing Systems},
 doi = {10.52202/068431-0134},
 editor = {S. Koyejo and S. Mohamed and A. Agarwal and D. Belgrave and K. Cho and A. Oh},
 pages = {1835--1849},
 publisher = {Curran Associates, Inc.},
 title = {FinRL-Meta: Market Environments and Benchmarks for Data-Driven Financial Reinforcement Learning},
 url = {https://proceedings.neurips.cc/paper_files/paper/2022/file/0bf54b80686d2c4dc0808c2e98d430f7-Paper-Datasets_and_Benchmarks.pdf},
 volume = {35},
 year = {2022}
}

@inproceedings{ICLR2026_4d423587,
 author = {HU, LIANG and Jiao, Jianpeng and Liu, Jiashuo and Mutu, Dongyuan and Ren, Yanle and Wen, Zhoufutu and Zhang, Kaiyuan and Zhang, Xuanliang and Gao, Xiang and He, Tianci and HU, FEI and Liao, Yali and Wang, Zaiyuan and Liu, Jingkai and Daibin, Sun and Zeng, Ziqing and zeng, zhiyuan and Yang, Chenghao and Yang, Qianyu and Yin, Mingren and Zhang, Ge and zhang, Xinyi and ZHAO, Xiying and Zhenwei, Zhu and Namkoong, Hongseok and Huang, Wenhao},
 booktitle = {International Conference on Learning Representations},
 editor = {C. Vondrick and B. Hariharan and C. Raffel and L. Pinto and D. Yang and A. Faust},
 pages = {46402--46428},
 title = {FinSearchComp: Towards a Realistic, Expert-Level Evaluation of Financial Search and Reasoning},
 url = {https://proceedings.iclr.cc/paper_files/paper/2026/file/4d42358702dff82e1436550a05ade260-Paper-Conference.pdf},
 volume = {2026},
 year = {2026}
}

@InProceedings{pmlr-v267-nagy25a,
  title = 	 {{LOB}-Bench: Benchmarking Generative {AI} for Finance - an Application to Limit Order Book Data},
  author =       {Nagy, Peer and Frey, Sascha Yves and Li, Kang and Sarkar, Bidipta and Vyetrenko, Svitlana and Zohren, Stefan and Calinescu, Ani and Foerster, Jakob Nicolaus},
  booktitle = 	 {Proceedings of the 42nd International Conference on Machine Learning},
  pages = 	 {45437--45460},
  year = 	 {2025},
  editor = 	 {Singh, Aarti and Fazel, Maryam and Hsu, Daniel and Lacoste-Julien, Simon and Berkenkamp, Felix and Maharaj, Tegan and Wagstaff, Kiri and Zhu, Jerry},
  volume = 	 {267},
  series = 	 {Proceedings of Machine Learning Research},
  month = 	 {13--19 Jul},
  publisher =    {PMLR},
  url = 	 {https://proceedings.mlr.press/v267/nagy25a.html},
}

@inproceedings{song2026timi,
 author = {Zifan Song and Kaitao Song and Guosheng Hu and Ding Qi and Junyao Gao and Xiaohua Wang and Dongsheng Li and Cairong Zhao},
 title = {Trade in Minutes! Rationality-Driven Agentic System for Quantitative Financial Trading},
 booktitle = {International Conference on Learning Representations},
 year = {2026},
 url = {https://www.microsoft.com/en-us/research/publication/trade-in-minutes-rationality-driven-agentic-system-for-quantitative-financial-trading/}
}

@inproceedings{shah2025cutoff,
 author = {Agam Shah and Liqin Ye and Sebastian Jaskowski and Wei Xu and Sudheer Chava},
 title = {Beyond the Reported Cutoff: Where Large Language Models Fall Short on Financial Knowledge},
 booktitle = {Second Conference on Language Modeling},
 year = {2025},
 url = {https://openreview.net/forum?id=lEpPFmGH3L}
}

@inproceedings{cao2024financialsummary,
 author = {Tianyu Cao and Natraj Raman and Danial Dervovic and Chenhao Tan},
 title = {Characterizing Multimodal Long-form Summarization: A Case Study on Financial Reports},
 booktitle = {First Conference on Language Modeling},
 year = {2024},
 url = {https://openreview.net/forum?id=hDoN0CAy5e}
}

@inproceedings{hennigen2024symbolic,
 author = {Lucas Torroba Hennigen and Zejiang Shen and Aniruddha Nrusimha and Bernhard Gapp and David Sontag and Yoon Kim},
 title = {Towards Verifiable Text Generation with Symbolic References},
 booktitle = {First Conference on Language Modeling},
 year = {2024},
 url = {https://openreview.net/forum?id=fib9qidCpY}
}

@inproceedings{weiss2024markets,
 author = {Martin Weiss and Nasim Rahaman and Manuel W{\"u}thrich and Yoshua Bengio and Li Erran Li and Bernhard Sch{\"o}lkopf and Chris Pal},
 title = {Redesigning Information Markets in the Era of Language Models},
 booktitle = {First Conference on Language Modeling},
 year = {2024},
 url = {https://openreview.net/forum?id=Zq9Dfj4nBo}
}

@inproceedings{ICLR2025_6261d3bc,
 author = {Li, Junjie and Liu, Yang and Liu, Weiqing and Fang, Shikai and Wang, Lewen and XU, Chang and Bian, Jiang},
 booktitle = {International Conference on Learning Representations},
 editor = {Y. Yue and A. Garg and N. Peng and F. Sha and R. Yu},
 pages = {39490--39524},
 title = {MarS: a Financial Market Simulation Engine Powered by Generative Foundation Model},
 url = {https://proceedings.iclr.cc/paper_files/paper/2025/file/6261d3bc9e9326ddff055595aabd54e1-Paper-Conference.pdf},
 volume = {2025},
 year = {2025}
}

@inproceedings{NEURIPS2023_6a386d70,
 author = {Xie, Qianqian and Han, Weiguang and Zhang, Xiao and Lai, Yanzhao and Peng, Min and Lopez-Lira, Alejandro and Huang, Jimin},
 booktitle = {Advances in Neural Information Processing Systems},
 doi = {10.52202/075280-1454},
 editor = {A. Oh and T. Naumann and A. Globerson and K. Saenko and M. Hardt and S. Levine},
 pages = {33469--33484},
 publisher = {Curran Associates, Inc.},
 title = {PIXIU: A Comprehensive Benchmark, Instruction Dataset and Large Language Model for Finance},
 url = {https://proceedings.neurips.cc/paper_files/paper/2023/file/6a386d703b50f1cf1f61ab02a15967bb-Paper-Datasets_and_Benchmarks.pdf},
 volume = {36},
 year = {2023}
}

@inproceedings{ICLR2024_5f7367f5,
 author = {Cheng, Junyan and Chin, Peter},
 booktitle = {International Conference on Learning Representations},
 editor = {B. Kim and Y. Yue and S. Chaudhuri and K. Fragkiadaki and M. Khan and Y. Sun},
 pages = {22153--22187},
 title = {SocioDojo: Building Lifelong Analytical Agents with Real-world Text and Time Series},
 url = {https://proceedings.iclr.cc/paper_files/paper/2024/file/5f7367f5686e6e1592e64209d172b91d-Paper-Conference.pdf},
 volume = {2024},
 year = {2024}
}

@inproceedings{maevolving,
 title = {Evolving Quantitative Reasoning through Self-Play in Digital Twin Markets},
 author = {Ma, Tianmi and Huang, Wenxin and Du, Jiawei and Li, Lin and Zhong, Xian and Zhou, Joey Tianyi},
 booktitle = {Proceedings of the 43rd International Conference on Machine Learning},
 pages = {84428--84445},
 year = {2026},
 volume = {306},
 series = {Proceedings of Machine Learning Research},
 publisher = {PMLR},
 url = {https://proceedings.mlr.press/v306/ma26m.html}
}

@misc{liu2026diagonprogrammabletestbedaiagent,
  title={Diagon: A Programmable Testbed for AI-Agent Cognitive Labor Markets},
  author={Xuan Liu and Haoyang Shang and Haojian Jin},
  year={2026},
  eprint={2604.06688},
  archivePrefix={arXiv},
  primaryClass={cs.CE},
  url={https://arxiv.org/abs/2604.06688}
}

@misc{wu2023bloomberggpt,
  title={{BloombergGPT}: A Large Language Model for Finance},
  author={Shijie Wu and Ozan Irsoy and Steven Lu and Vadim Dabravolski and Mark Dredze and Sebastian Gehrmann and Prabhanjan Kambadur and David Rosenberg and Gideon Mann},
  year={2023},
  archivePrefix={arXiv},
  eprint={2303.17564},
  url={https://arxiv.org/abs/2303.17564}
}

@misc{lee2024finllms,
  title={A Survey of Large Language Models in Finance ({FinLLMs})},
  author={Jean Lee and Nicholas Stevens and Soyeon Caren Han and Minseok Song},
  year={2024},
  archivePrefix={arXiv},
  eprint={2402.02315},
  url={https://arxiv.org/abs/2402.02315}
}

@misc{nie2024financialapplications,
  title={A Survey of Large Language Models for Financial Applications: Progress, Prospects and Challenges},
  author={Yuqi Nie and Yaxuan Kong and Xiaowen Dong and John M. Mulvey and H. Vincent Poor and Qingsong Wen and Stefan Zohren},
  year={2024},
  archivePrefix={arXiv},
  eprint={2406.11903},
  url={https://arxiv.org/abs/2406.11903}
}

@inproceedings{liu2025cogmir,
  title={Exploring Prosocial Irrationality for {LLM} Agents: A Social Cognition View},
  author={Xuan Liu and Jie Zhang and Haoyang Shang and Song Guo and Chengxu Yang and Quanyan Zhu},
  booktitle={The Thirteenth International Conference on Learning Representations},
  year={2025},
  url={https://arxiv.org/abs/2405.14744}
}

@inproceedings{liu2026cobra,
  title={{CoBRA}: Programming Cognitive Bias in Social Agents Using Classic Social Science Experiments},
  author={Xuan Liu and Haoyang Shang and Haojian Jin},
  booktitle={Proceedings of the 2026 CHI Conference on Human Factors in Computing Systems},
  year={2026},
  url={https://arxiv.org/abs/2509.13588}
}

@article{shang2025cothinker,
  title={United Minds or Isolated Agents? Exploring Coordination of {LLMs} under Cognitive Load Theory},
  author={HaoYang Shang and Xuan Liu and Zi Liang and Jie Zhang and Haibo Hu and Song Guo},
  journal={Transactions on Machine Learning Research},
  year={2025},
  url={https://arxiv.org/abs/2506.06843}
}

\clearpage

\onecolumn\raggedbottom

\appendix

\captionsetup{hypcap=false}

\nolinenumbers

\begin{multicols}{2}[][180pt]\raggedcolumns

\FloatBarrier

\section{Annotation and Consensus Formation}

\subsection{Annotation and Disagreement Resolution}

\textbf{Annotation and review personnel.} Two paper authors independently completed the initial annotations A and B, respectively. We commissioned a fintech company serving the securities and fund-management sectors to review the annotation results and participate in discussions to resolve disagreements. Its products cover quantitative research, risk analysis, and financial agents. The institution provided two external reviewers for this process. Their backgrounds are described below.

\textbf{External reviewer 1:} Holds a master's degree in finance and has over seven years of professional experience as a financial analyst, with a background in securities research. Their interdisciplinary experience spans finance and computer science, including financial-agent product development.

\textbf{External reviewer 2:} Holds a bachelor's degree in economics and has over eight years of professional experience, including financial analysis and annotation projects. They have also participated in the development of financial-agent products and quantitative factor databases.

\textbf{Initial annotator A (Author 1):} Holds a bachelor's degree with a double major in economics and finance and a master's degree in economics. Their experience includes large language model research, financial data analysis, and economics research.

\textbf{Initial annotator B (Author 2):} Holds a dual degree in financial engineering and an MPhil, with research experience in AI for Finance.

Items with matching results in the two independent annotations are retained, while the two external experts and two authors jointly review the remaining items to form final consensus labels. For the 59 items with raw-label disagreements in D, we retain both annotations and their rationales to trace the sources of disagreement. All reported rates and primary ablation results use final consensus labels, while independent annotations A and B are used to analyze pre-consensus agreement and differences in judgment.

Key-fact extraction and coverage assessment are conducted in two stages. First, annotators use only the target security, decision time, and external materials saved at that time to identify important facts that may affect trading judgments, including information that changes trading premises, provides important counterevidence, or constrains decision conditions. Model reports and actual trading directions are not provided during extraction, reducing the influence of model decisions on fact selection. Favorable, unfavorable, and neutral information is selected under the same criteria, and duplicate reports are merged into a single fact. After the fact list is determined and frozen, model reports and trading directions are introduced to check whether these facts have been adequately addressed. The entire process uses only materials available at decision time, without referring to subsequent market movements or investment returns.

Coverage assessment does not require reports to restate facts one by one; we examine whether they understand and address the implications of those facts for trading decisions. For example, a report may acknowledge unfavorable evidence and weigh it, offer an alternative interpretation based on other materials, or explain why a fact is insufficient to change the trading judgment. For materials saved only as titles, assessment is limited to information explicitly stated in the titles, retaining uncertainty qualifiers such as 'proposed' and 'expected.'

Therefore, K measures whether a decision adequately considers relevant information, rather than whether the model's investment view agrees with the annotators'. Industry and macroeconomic materials are assessed in relation to the target stock, report arguments, and actual trading actions; the same fact may support different trading choices, provided that the report explains its decision implications and addresses important counterevidence. Company cash-flow and balance-sheet information constitutes external evidence, while cash, holdings, and sellable quantities in the model's account are verified through D and execution records, respectively. For disputed items, reviewers adjudicate using the original materials and shared judgment criteria, retaining the corresponding rationales.

R/K and D each cover 1,384 formal trades, while Temp covers 1,293 adjacent trade pairs.

\end{multicols}

\par\noindent\begin{minipage}{\textwidth}
\centering
\small
\begin{tabularx}{\textwidth}{@{}>{\raggedright\arraybackslash}p{0.2550\textwidth}>{\hsize=1.0000\hsize \raggedleft\arraybackslash}X>{\hsize=1.0000\hsize \raggedleft\arraybackslash}X>{\hsize=1.0000\hsize \raggedleft\arraybackslash}X>{\hsize=1.0000\hsize \raggedleft\arraybackslash}X@{}}
\toprule
\textbf{Model} & \textbf{R A/\allowbreak{}B} & \textbf{K A/\allowbreak{}B} & \textbf{D A/\allowbreak{}B} & \textbf{Temp A/\allowbreak{}B} \\
\midrule
Claude Opus 4.5 & 33.5/35.5 & 40.6/38.7 & 11.5/7.8 & 0.0/2.0 \\
GLM 4.7 Flash & 26.4/29.9 & 48.3/46.0 & 56.2/56.2 & 6.3/13.9 \\
DeepSeek V3.2 & 20.7/21.8 & 57.5/52.5 & 11.2/12.7 & 6.4/9.2 \\
Gemini 3.1 Pro & 26.4/33.0 & 45.2/43.1 & 6.1/6.6 & 6.2/9.8 \\
Claude Haiku 4.5 & 47.5/47.5 & 39.9/38.3 & 51.1/39.9 & 0.0/6.9 \\
Qwen3 Max & 20.2/21.3 & 45.7/43.6 & 25.0/20.0 & 0.0/6.0 \\
DeepSeek V3.1 & 29.4/29.4 & 54.9/52.9 & 0.0/0.0 & 2.3/4.7 \\
GLM 4.7 & 30.3/24.2 & 39.4/39.4 & 34.6/26.9 & 11.5/11.5 \\
MiniMax M2.1 & 26.2/28.6 & 42.9/38.1 & 25.5/18.0 & 5.3/12.0 \\
GPT-5.1 & N/A/N/A & N/A/N/A & N/A/N/A & N/A/N/A \\
GPT-5.2 & 12.6/12.6 & 38.3/36.0 & 0.0/0.0 & 0.0/0.0 \\
Qwen3 235B & 27.1/29.2 & 56.2/56.2 & 43.8/37.5 & 7.5/25.0 \\
Gemini 2.5 Flash & 20.4/21.4 & 55.1/54.1 & 17.4/17.0 & 3.3/15.2 \\
\bottomrule
\end{tabularx}
\renewcommand{\thetable}{A1}
\captionof{table}{Gap rates from annotations A and B (\%). R/K use formal trades, D uses applicable account assessments, and Temp uses adjacent trade pairs.}
\label{tab:A1}
\end{minipage}\par\addvspace{\floatsep}

\begin{multicols}{2}[][180pt]\raggedcolumns

\subsection{Agreement}

\end{multicols}

\par\noindent\begin{minipage}{\textwidth}
\centering
\small
\begin{tabularx}{\textwidth}{@{}>{\raggedright\arraybackslash}p{0.2050\textwidth}>{\hsize=1.0000\hsize \raggedleft\arraybackslash}X>{\hsize=1.0000\hsize \raggedleft\arraybackslash}X>{\hsize=1.0000\hsize \raggedleft\arraybackslash}X>{\hsize=1.0000\hsize \raggedleft\arraybackslash}X@{}}
\toprule
\textbf{Metric} & \textbf{Pairs} & \textbf{Agreement \%} & \textbf{Cohen's \ensuremath{\kappa}} & \textbf{Gwet's AC1} \\
\midrule
R & 1384 & 95.81 & 0.896 & 0.930 \\
K & 1384 & 97.18 & 0.943 & 0.944 \\
D & 1384 & 95.88 & 0.931 & 0.941 \\
Temp & 1293 & 94.43 & 0.483 & 0.938 \\
\bottomrule
\end{tabularx}
\renewcommand{\thetable}{A2}
\captionof{table}{Agreement between annotations A and B. R, K, and Temp map to pass/gap, with unable-to-determine cases counted as gaps; D maps to pass/gap/not applicable and includes all 1,384 records to retain applicability disagreements. Agreement is the proportion of identical annotations under this mapping; Cohen's \ensuremath{\kappa} and Gwet's AC1 correct for chance agreement.}
\label{tab:A2}
\end{minipage}\par\addvspace{\floatsep}

\begin{multicols}{2}[][180pt]\raggedcolumns

Agreement on the four raw D labels is 95.74\%, with \ensuremath{\kappa}=0.929; merging confirmed errors and unable-to-determine cases removes two disagreements. Among the 1,293 Temp pairs, 1,221 have identical raw labels, while 72 differ between coherent and partially coherent (67 labeled coherent by A and partially coherent by B, and five in the reverse direction). Audit-category agreement is 94.43\%, with \ensuremath{\kappa}=0.483; four-category raw-label agreement is 94.43\%, with \ensuremath{\kappa}=0.490.

For raw Temp labels, observed agreement is \(p_o=1221/1293=0.9443\), and chance agreement computed from the A/B category proportions is \(p_e=0.8908\), giving \(\kappa=(p_o-p_e)/(1-p_e)=0.490\). The prevalence of coherent labels raises chance agreement \citep{artstein2008agreement}; A/B also assign 15/77 partially coherent labels, reflecting different judgments at this boundary. Table~\ref{tab:1} excludes three fixed examples and uses 1,290 pairs, hence its slightly different full-category coefficient.

\subsection{Effects of Annotation Differences on Filtering Results}

\end{multicols}

\par\noindent\begin{minipage}{\textwidth}
\centering
\small
\begin{tabularx}{\textwidth}{@{}>{\raggedright\arraybackslash}p{0.2050\textwidth}>{\hsize=1.0000\hsize \raggedleft\arraybackslash}X>{\hsize=1.0000\hsize \raggedleft\arraybackslash}X>{\hsize=1.0000\hsize \raggedleft\arraybackslash}X>{\hsize=1.0000\hsize \raggedleft\arraybackslash}X@{}}
\toprule
\textbf{Label set} & \textbf{Filter} & \textbf{Up/\allowbreak{}down/\allowbreak{}unchanged} & \textbf{Mean \ensuremath{\Delta}CR} & \textbf{Actions removed} \\
\midrule
A & R & 8/4/0 & +6.86 & 372 \\
A & K & 10/2/0 & +9.12 & 638 \\
A & RK & 9/3/0 & +9.13 & 850 \\
B & R & 8/4/0 & +6.51 & 396 \\
B & K & 11/1/0 & +10.46 & 605 \\
B & RK & 10/2/0 & +8.87 & 843 \\
Consensus & R & 8/4/0 & +6.66 & 399 \\
Consensus & K & 10/2/0 & +9.29 & 613 \\
Consensus & RK & 9/3/0 & +8.86 & 849 \\
\bottomrule
\end{tabularx}
\renewcommand{\thetable}{A3}
\captionof{table}{Filtering results from annotations A and B and consensus labels. Mean \ensuremath{\Delta}CR is in percentage points.}
\label{tab:A3}
\end{minipage}\par\addvspace{\floatsep}

\begin{multicols}{2}[][180pt]\raggedcolumns

\subsection{Annotation for the New Model}

GPT-6 Sol's two R/K annotation rounds use the same frozen key-fact list. Raw agreement for citation risk, external support, and key-fact coverage is 98.96\%, 95.83\%, and 98.96\%, respectively; six category disagreements are adjudicated before replay. The 96 D labels and 87 Temp labels agree in both rounds; both have constant categories, making Cohen's \ensuremath{\kappa} undefined.

\subsection{Annotation Inputs and Decision Records}

Each annotation retains the target event, original statement, verification basis, category, and brief rationale. Materials are linked by decision time, and account statements are checked by security, pre- or post-trade timing, and calculation convention. Report text is supplied as content to be verified.

\end{multicols}

\par\noindent\begin{minipage}{\textwidth}
\centering
\small
\begin{tabularx}{\textwidth}{@{}>{\raggedright\arraybackslash}p{0.1450\textwidth}>{\hsize=0.8537\hsize \raggedright\arraybackslash}X>{\hsize=1.1463\hsize \raggedright\arraybackslash}X@{}}
\toprule
\textbf{Annotation stage} & \textbf{Input materials} & \textbf{Decision record} \\
\midrule
Key-fact extraction & Security, decision time, and saved external materials & Fact content, source, importance, and insufficient-material status \\
R & Report citations, trading action, corresponding sources, and current context & Claims requiring verification, original passages providing support or contradiction, and citation risk \\
K & Frozen fact list, report, and trading action & Facts addressed, partially addressed, or unaddressed, and coverage assessment \\
D & Report, current price, pre- and post-trade accounts, and execution records & Confirmed error, no confirmed error found, no verifiable account assessment, or indeterminate, with calculation bases retained \\
Temp & Adjacent trades by the same model in the same security, their reports, current materials, and account states & Coherent, partially coherent, incoherent, or indeterminate, with the basis for the change explained \\
\bottomrule
\end{tabularx}
\renewcommand{\thetable}{A4}
\captionof{table}{Inputs and outputs of semantic annotation.}
\label{tab:A4}
\end{minipage}\par\addvspace{\floatsep}

\begin{multicols}{2}[][180pt]\raggedcolumns

R checks correspondence between cited facts and their sources; when titles do not match, readable facts are checked to determine citation risk. Insufficient materials are labeled unable to determine. K checks whether relevant facts are fully addressed: for example, citing revenue growth to support buying without addressing a major loss in the same material. K is also triggered when all citations concern industry or macroeconomic information, their assessments oppose the actual trade, and review finds them insufficient to support the stock-specific decision. D distinguishes daily price changes, unrealized position gains, sale proceeds, and realized profits before comparing values on the same basis. Temp checks whether trade changes are explained by the preceding and subsequent reports, applying the same criterion to continued and reversed directions.

\subsection{Model Annotation and Expert-Consensus Agreement}

After forming consensus, we select nine trades as fixed few-shot examples for R, K, and D, and three adjacent trade pairs for Temp. Examples include original materials, labels, and rationales, with the same examples used across groups for both models. GPT-6 Pro and GPT-6 Astra independently produce labels and rationales from reports, decision-time materials, frozen key-fact lists, and relevant account records, which are then compared with expert consensus.

After excluding all examples, evaluation includes 1,375 records each for R, K, and D, and 1,290 pairs for Temp. Labels follow the main-text audit mapping: unable-to-determine cases count as gaps, only coherent Temp labels count as passes, while partially coherent and incoherent labels count as gaps, and D retains a not-applicable category. Table A5 reports author A/B agreement and agreement of both models with expert consensus on the same samples. Mean rows equally average agreement, \ensuremath{\kappa}, AC1, and F1 across R, K, D, and Temp. Within each metric, macro-F1 averages equally over categories occurring in either annotation, while gap F1 treats gap as the target category.

\end{multicols}

\par\noindent\begin{minipage}{\textwidth}
\centering
\small
\renewcommand{\arraystretch}{1.12}
\setlength{\tabcolsep}{4pt}
\begin{tabularx}{\textwidth}{@{}>{\raggedright\arraybackslash}p{0.1450\textwidth}>{\hsize=0.7256\hsize \raggedright\arraybackslash}X>{\hsize=0.6402\hsize \raggedleft\arraybackslash}X>{\hsize=1.1951\hsize \raggedleft\arraybackslash}X>{\hsize=0.8963\hsize \raggedleft\arraybackslash}X>{\hsize=0.8963\hsize \raggedleft\arraybackslash}X>{\hsize=1.3232\hsize \raggedleft\arraybackslash}X>{\hsize=1.3232\hsize \raggedleft\arraybackslash}X@{}}
\toprule
\morecmidrules\cmidrule{1-8}
\textbf{Comparison} & \textbf{Metric} & \textbf{n} & \textbf{Agree. \%} & \textbf{\ensuremath{\kappa}} & \textbf{AC1} & \textbf{Macro-F1} & \textbf{Gap F1} \\
\midrule
\rowcolor{TableBand}
A/B & R & 1375 & 95.78 & 0.895 & 0.930 & - & - \\
 & K & 1375 & 97.24 & 0.944 & 0.945 & - & - \\
\rowcolor{TableBand}
 & D & 1375 & 95.85 & 0.930 & 0.941 & - & - \\
 & Temp & 1290 & 94.50 & 0.480 & 0.939 & - & - \\
\rowcolor{TableAccent}
 & \textbf{Mean} & \textbf{-} & \textbf{95.84} & \textbf{0.812} & \textbf{0.939} & \textbf{-} & \textbf{-} \\
\midrule
GPT-6 Pro & R & 1375 & 92.51 & 0.823 & 0.870 & 0.912 & 0.877 \\
\rowcolor{TableBand}
 & K & 1375 & 98.47 & 0.969 & 0.970 & 0.985 & 0.983 \\
 & D & 1375 & 95.42 & 0.922 & 0.935 & 0.926 & 0.833 \\
\rowcolor{TableBand}
 & Temp & 1290 & 92.95 & 0.405 & 0.920 & 0.702 & 0.442 \\
\rowcolor{TableAccent}
 & \textbf{Mean} & \textbf{-} & \textbf{94.84} & \textbf{0.780} & \textbf{0.924} & \textbf{0.881} & \textbf{0.784} \\
\rowcolor{TableBand}
\midrule
GPT-6 Astra & R & 1375 & 90.69 & 0.783 & 0.837 & 0.891 & 0.850 \\
 & K & 1375 & 96.15 & 0.923 & 0.923 & 0.961 & 0.958 \\
\rowcolor{TableBand}
 & D & 1375 & 90.18 & 0.840 & 0.859 & 0.856 & 0.673 \\
 & Temp & 1290 & 94.81 & 0.657 & 0.939 & 0.829 & 0.685 \\
\rowcolor{TableAccent}
 & \textbf{Mean} & \textbf{-} & \textbf{92.96} & \textbf{0.801} & \textbf{0.889} & \textbf{0.884} & \textbf{0.792} \\
\cmidrule{1-8}
\bottomrule
\end{tabularx}
\renewcommand{\thetable}{A5}
\captionof{table}{Annotator and model agreement under audit-category mapping. A/B compares the two independent author annotations before consensus; both models are compared separately with expert consensus. All examples are excluded. Agreement is reported as a percentage.}
\label{tab:A5}
\end{minipage}\par\addvspace{\floatsep}

\begin{multicols}{2}[][180pt]\raggedcolumns

Under audit-category merging, mean agreement is 94.84\% for GPT-6 Pro and 92.96\% for GPT-6 Astra. Astra has higher agreement and gap F1 on Temp, whereas Pro has higher agreement and gap F1 on R, K, and D.

\end{multicols}

\par\noindent\begin{minipage}{\textwidth}
\centering
\small
\renewcommand{\arraystretch}{1.12}
\setlength{\tabcolsep}{4pt}
\begin{tabularx}{\textwidth}{@{}>{\raggedright\arraybackslash}p{0.0650\textwidth}>{\hsize=0.8000\hsize \raggedleft\arraybackslash}X>{\hsize=0.8000\hsize \raggedleft\arraybackslash}X>{\hsize=0.9333\hsize \raggedleft\arraybackslash}X>{\hsize=0.9333\hsize \raggedleft\arraybackslash}X>{\hsize=1.2667\hsize \raggedleft\arraybackslash}X>{\hsize=1.2667\hsize \raggedleft\arraybackslash}X@{}}
\toprule
\morecmidrules\cmidrule{1-7}
\textbf{Metric} & \textbf{A \%} & \textbf{B \%} & \textbf{Pro \%} & \textbf{Astra \%} & \textbf{Pro disagreements} & \textbf{Astra disagreements} \\
\midrule
\rowcolor{TableBand}
R & 96.29 & 99.49 & 92.51 & 90.69 & 103 & 128 \\
K & 97.82 & 99.42 & 98.47 & 96.15 & 21 & 53 \\
\rowcolor{TableBand}
D & 96.65 & 99.20 & 95.42 & 90.18 & 63 & 135 \\
Temp & 95.04 & 99.46 & 92.95 & 94.81 & 91 & 67 \\
\rowcolor{TableAccent}
\textbf{Mean} & \textbf{96.45} & \textbf{99.39} & \textbf{94.84} & \textbf{92.96} & \textbf{278} & \textbf{383} \\
\cmidrule{1-7}
\bottomrule
\end{tabularx}
\renewcommand{\thetable}{A6}
\captionof{table}{Agreement of independent author annotations and both models with expert consensus. The same samples as in Table A5 are used, with examples excluded. Agreement is reported as a percentage, and disagreements count records whose model labels differ from expert consensus in audit category. The Mean row equally averages agreement across the four metrics and sums disagreements. A and B both contribute to consensus; their agreement describes label changes during consensus formation.}
\label{tab:A6}
\end{minipage}\par\addvspace{\floatsep}

\begin{multicols}{2}[][180pt]\raggedcolumns

Raw labels retain distinctions such as unable to determine, partially coherent, and not applicable. Main-text Table~\ref{tab:1} reports agreement without category merging, using the same samples and model annotations as Table A5 and differing only in category merging.

\end{multicols}

\par\noindent\begin{minipage}{\textwidth}
\centering
\small
\begin{tabularx}{\textwidth}{@{}>{\raggedright\arraybackslash}p{0.2050\textwidth}>{\hsize=0.3947\hsize \raggedright\arraybackslash}X>{\hsize=1.8947\hsize \raggedright\arraybackslash}X>{\hsize=0.7105\hsize \raggedright\arraybackslash}X@{}}
\toprule
\textbf{Example} & \textbf{Check} & \textbf{Judgment criterion} & \textbf{Labels} \\
\midrule
1 & R/K/D & The repurchase remains a proposal; cash and zero-position statements match the ledger & P/P/P \\
2 & R/K/D & Forecast profit is described as realized earnings; the position percentage is verifiable & G/P/P \\
3 & R/K/D & Turnover is described as confirmed net fund inflows & G/P/P \\
4 & R/K/D & Favorable facts do not support the sale direction, and key facts are not adequately addressed & G/G/P \\
5 & R/K/D & The citation is not distorted, but key facts on declining installations and exports are omitted & P/G/P \\
6 & R/K/D & External facts and account assessments are checked separately; the latter contain a confirmed error & P/P/G \\
7 & R/K/D & Profit/loss statements are checked against the cost of the same security's position & P/P/G \\
8 & R/K/D & Generic risk language does not constitute a verifiable account statement & P/P/N \\
9 & R/K/D & The formal report is missing, retaining an insufficient-material status & U/U/U \\
10 & Temp & New market risk explains the transition from buying to selling & Coherent \\
11 & Temp & Reentry is explained, but the previously stated position criterion is not addressed & Partially coherent \\
12 & Temp & The earlier formal report is missing & Unable to determine \\
\bottomrule
\end{tabularx}
\renewcommand{\thetable}{A7}
\captionof{table}{Fixed few-shot examples and judgment criteria. Trade labels are listed in R/K/D order. P, G, U, and N denote pass, gap, unable to determine, and not applicable, respectively; Temp retains its raw labels.}
\label{tab:A7}
\end{minipage}\par\addvspace{\floatsep}

\begin{multicols}{2}[][180pt]\raggedcolumns

\FloatBarrier

\section{Sample Statistics and Supplementary Results}

\subsection{Statistical Definitions}

Formal R/K/D consensus labels cover 1,384 trades at 1,070 trading points, which define the point-level sample in main-text Tables~\ref{tab:2} and~\ref{tab:3}. Tables B1 and B1a report formal labels; Table B1b lists per-model annotation counts and D account-claim coverage.

Main-text RK statistics pool all 1,070 annotated trading points: 356 pass, 678 have confirmed gaps, and 36 are undetermined only, accounting for 33.27\%, 63.36\%, and 3.36\%, respectively. The latter two categories give 714 gap points (66.73\%). A point with a confirmed R or K gap belongs to the confirmed category even if the other check is undetermined, avoiding overlap with the undetermined-only category. These pooled counts differ from equally averaging per-model percentages.

Tables~\ref{tab:2} and~\ref{tab:3} in the main text report results by trading point, using each model's decision points with actual trades as the denominator and counting repeated gaps at a point once per check. All-point results appear in B.6. The detailed statistics below retain each check's original sample unit: Report uses scheduled windows; Rpt, Call, and Resp use their applicable records; R/K is measured over formal active trades, D over events involving account assessments, and Temp over applicable adjacent trade pairs. Detailed gap rates equal (failed + unable-to-determine records)/applicable records, excluding structurally inapplicable items from the denominator. The table below separates assessable and unable-to-determine counts.

\end{multicols}

\par\noindent\begin{minipage}{\textwidth}
\centering
\small
\begin{tabularx}{\textwidth}{@{}>{\raggedright\arraybackslash}p{0.2550\textwidth}>{\hsize=1.0000\hsize \raggedleft\arraybackslash}X>{\hsize=1.0000\hsize \raggedleft\arraybackslash}X>{\hsize=1.0000\hsize \raggedleft\arraybackslash}X>{\hsize=1.0000\hsize \raggedleft\arraybackslash}X>{\hsize=1.0000\hsize \raggedleft\arraybackslash}X@{}}
\toprule
\textbf{Model} & \textbf{R} & \textbf{K} & \textbf{RK} & \textbf{D} & \textbf{Temp} \\
\midrule
Claude Opus 4.5 & 155/0 & 149/6 & 151/4 & 115/0 & 148/0 \\
GLM 4.7 Flash & 87/0 & 85/2 & 86/1 & 29/3 & 74/5 \\
DeepSeek V3.2 & 179/0 & 171/8 & 172/7 & 80/0 & 168/5 \\
Gemini 3.1 Pro & 194/3 & 190/7 & 190/7 & 131/4 & 186/7 \\
Claude Haiku 4.5 & 182/1 & 176/7 & 178/5 & 173/5 & 174/0 \\
Qwen3 Max & 94/0 & 89/5 & 90/4 & 54/0 & 84/0 \\
DeepSeek V3.1 & 51/0 & 50/1 & 50/1 & 12/0 & 43/0 \\
GLM 4.7 & 32/1 & 31/2 & 31/2 & 25/1 & 23/3 \\
MiniMax M2.1 & 84/0 & 82/2 & 82/2 & 50/0 & 73/2 \\
GPT-5.1 & 0/0 & 0/0 & 0/0 & 0/0 & 0/0 \\
GPT-5.2 & 175/0 & 166/9 & 166/9 & 122/0 & 166/0 \\
Qwen3 235B & 48/0 & 48/0 & 48/0 & 14/2 & 37/3 \\
Gemini 2.5 Flash & 96/2 & 94/4 & 94/4 & 45/2 & 90/2 \\
\bottomrule
\end{tabularx}
\renewcommand{\thetable}{B1}
\captionof{table}{Assessable/unable-to-determine counts for consensus labels. D records with no account assessment are excluded from both counts and reported in Table B1a. Temp counts are trade pairs. Across all models, unable-to-determine rates for R, K, RK, D, and Temp are 0.51\%, 3.83\%, 3.32\%, 1.96\%, and 2.09\%, respectively, using the same denominators as their gap rates.}
\label{tab:B1}
\end{minipage}\par\addvspace{\floatsep}

\par\noindent\begin{minipage}{\textwidth}
\centering
\small
\begin{tabularx}{\textwidth}{@{}>{\raggedright\arraybackslash}p{0.2550\textwidth}>{\hsize=1.0000\hsize \raggedleft\arraybackslash}X>{\hsize=1.0000\hsize \raggedleft\arraybackslash}X>{\hsize=1.0000\hsize \raggedleft\arraybackslash}X>{\hsize=1.0000\hsize \raggedleft\arraybackslash}X@{}}
\toprule
\textbf{Model} & \textbf{Confirmed error} & \textbf{No error found} & \textbf{No account assessment} & \textbf{Unable to determine} \\
\midrule
Claude Opus 4.5 & 9 & 106 & 40 & 0 \\
GLM 4.7 Flash & 15 & 14 & 55 & 3 \\
DeepSeek V3.2 & 10 & 70 & 99 & 0 \\
Gemini 3.1 Pro & 5 & 126 & 62 & 4 \\
Claude Haiku 4.5 & 71 & 102 & 5 & 5 \\
Qwen3 Max & 11 & 43 & 40 & 0 \\
DeepSeek V3.1 & 0 & 12 & 39 & 0 \\
GLM 4.7 & 7 & 18 & 7 & 1 \\
MiniMax M2.1 & 9 & 41 & 34 & 0 \\
GPT-5.1 & 0 & 0 & 0 & 0 \\
GPT-5.2 & 0 & 122 & 53 & 0 \\
Qwen3 235B & 4 & 10 & 32 & 2 \\
Gemini 2.5 Flash & 6 & 39 & 51 & 2 \\
\bottomrule
\end{tabularx}
\renewcommand{\thetable}{B1a}
\captionof{table}{Complete category counts for D. 'No account assessment' denotes reports without verifiable account statements.}
\label{tab:B1a}
\end{minipage}\par\addvspace{\floatsep}

\par\noindent\begin{minipage}{\textwidth}
\centering
\small
\begin{tabularx}{\textwidth}{@{}>{\raggedright\arraybackslash}p{0.2550\textwidth}>{\hsize=1.0000\hsize \raggedleft\arraybackslash}X>{\hsize=1.0000\hsize \raggedleft\arraybackslash}X>{\hsize=1.0000\hsize \raggedleft\arraybackslash}X>{\hsize=1.0000\hsize \raggedleft\arraybackslash}X@{}}
\toprule
\textbf{Model} & \textbf{Formal trades} & \textbf{Annotated points} & \textbf{D coverage (\%)} & \textbf{Calibration (excluded)} \\
\midrule
Claude Opus 4.5 & 155 & 123 & 74.19 & 2 \\
GLM 4.7 Flash & 87 & 53 & 36.78 & 2 \\
DeepSeek V3.2 & 179 & 120 & 44.69 & 7 \\
Gemini 3.1 Pro & 197 & 112 & 68.53 & 9 \\
Claude Haiku 4.5 & 183 & 147 & 97.27 & 8 \\
Qwen3 Max & 94 & 86 & 57.45 & 0 \\
DeepSeek V3.1 & 51 & 46 & 23.53 & 0 \\
GLM 4.7 & 33 & 30 & 78.79 & 0 \\
MiniMax M2.1 & 84 & 68 & 59.52 & 1 \\
GPT-5.1 & 0 & 0 & N/A & 0 \\
GPT-5.2 & 175 & 161 & 69.71 & 5 \\
Qwen3 235B & 48 & 40 & 33.33 & 1 \\
Gemini 2.5 Flash & 98 & 84 & 47.96 & 5 \\
\bottomrule
\end{tabularx}
\renewcommand{\thetable}{B1b}
\captionof{table}{Formal annotation scope and account-claim coverage. D coverage divides applicable account assessments by formal trades; calibration trades do not enter R/K/D gap counts.}
\label{tab:B1b}
\end{minipage}\par\addvspace{\floatsep}

\begin{multicols}{2}[][180pt]\raggedcolumns

GPT-5.2 has 175 formally annotated trades at 161 points: 122 contain account assessments and 53 are inapplicable to D. Expert consensus identifies no confirmed errors or unable-to-determine cases, so D gaps are zero in both main-text Table~\ref{tab:3} and Table B1a.

Temp uses an independent cohort of 1,293 labeled pairs; first trades are structurally inapplicable. Expert unable-to-determine labels count as gaps. Appendix B.5 reports the full category distribution.

\subsection{R/K Overlap and Call-Weighted Comparisons}

\end{multicols}

\par\noindent\begin{minipage}{\textwidth}
\centering
\small
\begin{tabularx}{\textwidth}{@{}>{\raggedright\arraybackslash}p{0.2550\textwidth}>{\hsize=1.0000\hsize \raggedleft\arraybackslash}X>{\hsize=1.0000\hsize \raggedleft\arraybackslash}X>{\hsize=1.0000\hsize \raggedleft\arraybackslash}X>{\hsize=1.0000\hsize \raggedleft\arraybackslash}X>{\hsize=1.0000\hsize \raggedleft\arraybackslash}X>{\hsize=1.0000\hsize \raggedleft\arraybackslash}X@{}}
\toprule
\textbf{Model} & \textbf{R} & \textbf{K} & \textbf{R\ensuremath{\cap}K} & \textbf{R only} & \textbf{K only} & \textbf{R\ensuremath{\cup}K} \\
\midrule
Claude Opus 4.5 & 55 & 61 & 23 & 32 & 38 & 93 \\
GLM 4.7 Flash & 26 & 40 & 11 & 15 & 29 & 55 \\
DeepSeek V3.2 & 37 & 96 & 19 & 18 & 77 & 114 \\
Gemini 3.1 Pro & 65 & 86 & 26 & 39 & 60 & 125 \\
Claude Haiku 4.5 & 91 & 72 & 29 & 62 & 43 & 134 \\
Qwen3 Max & 20 & 41 & 8 & 12 & 33 & 53 \\
DeepSeek V3.1 & 15 & 27 & 10 & 5 & 17 & 32 \\
GLM 4.7 & 8 & 13 & 3 & 5 & 10 & 18 \\
MiniMax M2.1 & 25 & 33 & 13 & 12 & 20 & 45 \\
GPT-5.1 & 0 & 0 & 0 & 0 & 0 & 0 \\
GPT-5.2 & 22 & 64 & 5 & 17 & 59 & 81 \\
Qwen3 235B & 14 & 27 & 6 & 8 & 21 & 35 \\
Gemini 2.5 Flash & 21 & 53 & 10 & 11 & 43 & 64 \\
All & 399 & 613 & 163 & 236 & 450 & 849 \\
\bottomrule
\end{tabularx}
\renewcommand{\thetable}{B2}
\captionof{table}{Action counts flagged by final consensus labels. The union is deduplicated by actual action.}
\label{tab:B2}
\end{minipage}\par\addvspace{\floatsep}

\par\noindent\begin{minipage}{\textwidth}
\centering
\small
\begin{tabularx}{\textwidth}{@{}>{\raggedright\arraybackslash}p{0.2550\textwidth}>{\hsize=1.0000\hsize \raggedleft\arraybackslash}X>{\hsize=1.0000\hsize \raggedleft\arraybackslash}X>{\hsize=1.0000\hsize \raggedleft\arraybackslash}X@{}}
\toprule
\textbf{Model} & \textbf{R} & \textbf{K} & \textbf{RK} \\
\midrule
Claude Opus 4.5 & 37.0 & 36.4 & 58.4 \\
GLM 4.7 Flash & 31.0 & 42.9 & 61.6 \\
DeepSeek V3.2 & 20.4 & 52.4 & 64.4 \\
Gemini 3.1 Pro & 33.9 & 46.8 & 66.2 \\
Claude Haiku 4.5 & 51.1 & 39.9 & 73.4 \\
Qwen3 Max & 23.3 & 40.7 & 54.7 \\
DeepSeek V3.1 & 29.3 & 55.4 & 64.1 \\
GLM 4.7 & 25.0 & 43.3 & 58.3 \\
MiniMax M2.1 & 30.5 & 42.3 & 55.9 \\
GPT-5.1 & N/A & N/A & N/A \\
GPT-5.2 & 12.4 & 38.0 & 47.3 \\
Qwen3 235B & 33.8 & 55.8 & 75.8 \\
Gemini 2.5 Flash & 20.8 & 54.8 & 65.7 \\
\bottomrule
\end{tabularx}
\renewcommand{\thetable}{B3}
\captionof{table}{Call-weighted R/K gap-rate comparisons (\%). Security-level events within each call are counted using the corresponding weights; the detailed event-level results weight events equally.}
\label{tab:B3}
\end{minipage}\par\addvspace{\floatsep}

\begin{multicols}{2}[][180pt]\raggedcolumns

\subsection{Per-Model Joint Eight-Metric Results}

Joint statistics use decision points as the unit, combining all assessed applicable checks at each point. A point fails if any check has a confirmed failure; if no check has a confirmed failure but an unable-to-determine item exists, the point is undetermined; a point passes only when all assessed applicable checks pass. Unannotated items are excluded from semantic checks. At non-trading points, only O's four checks are assessed. Table B4 covers all decision points and summarizes completed assessments at each point; Table B5 matches the main text and includes only annotated trading points. Gap rates combine failures and undetermined cases.

\end{multicols}

\par\noindent\begin{minipage}{\textwidth}
\centering
\small
\begin{tabularx}{\textwidth}{@{}>{\raggedright\arraybackslash}p{0.2550\textwidth}>{\hsize=1.0000\hsize \raggedleft\arraybackslash}X>{\hsize=1.0000\hsize \raggedleft\arraybackslash}X>{\hsize=1.0000\hsize \raggedleft\arraybackslash}X>{\hsize=1.0000\hsize \raggedleft\arraybackslash}X@{}}
\toprule
\textbf{Model} & \textbf{Points} & \textbf{Pass/\allowbreak{}fail/\allowbreak{}undetermined} & \textbf{Joint gap rate \%} & \textbf{Undetermined \%} \\
\midrule
Claude Opus 4.5 & 246 & 162/82/2 & 34.15 & 0.81 \\
GLM 4.7 Flash & 246 & 189/56/1 & 23.17 & 0.41 \\
DeepSeek V3.2 & 246 & 151/90/5 & 38.62 & 2.03 \\
Gemini 3.1 Pro & 246 & 152/91/3 & 38.21 & 1.22 \\
Claude Haiku 4.5 & 246 & 97/146/3 & 60.57 & 1.22 \\
Qwen3 Max & 246 & 195/48/3 & 20.73 & 1.22 \\
DeepSeek V3.1 & 246 & 214/31/1 & 13.01 & 0.41 \\
GLM 4.7 & 246 & 202/44/0 & 17.89 & 0.00 \\
MiniMax M2.1 & 246 & 196/48/2 & 20.33 & 0.81 \\
GPT-5.1 & 246 & 246/0/0 & 0.00 & 0.00 \\
GPT-5.2 & 246 & 161/77/8 & 34.55 & 3.25 \\
Qwen3 235B & 246 & 157/88/1 & 36.18 & 0.41 \\
Gemini 2.5 Flash & 246 & 164/81/1 & 33.33 & 0.41 \\
All & 3198 & 2286/882/30 & 28.52 & 0.94 \\
\bottomrule
\end{tabularx}
\renewcommand{\thetable}{B4}
\captionof{table}{Joint results over all points.}
\label{tab:B4}
\end{minipage}\par\addvspace{\floatsep}

\par\noindent\begin{minipage}{\textwidth}
\centering
\small
\begin{tabularx}{\textwidth}{@{}>{\raggedright\arraybackslash}p{0.2550\textwidth}>{\hsize=1.0000\hsize \raggedleft\arraybackslash}X>{\hsize=1.0000\hsize \raggedleft\arraybackslash}X>{\hsize=1.0000\hsize \raggedleft\arraybackslash}X>{\hsize=1.0000\hsize \raggedleft\arraybackslash}X@{}}
\toprule
\textbf{Model} & \textbf{Points} & \textbf{Pass/\allowbreak{}fail/\allowbreak{}undetermined} & \textbf{Joint gap rate \%} & \textbf{Undetermined \%} \\
\midrule
Claude Opus 4.5 & 123 & 39/82/2 & 68.29 & 1.63 \\
GLM 4.7 Flash & 53 & 8/44/1 & 84.91 & 1.89 \\
DeepSeek V3.2 & 120 & 28/87/5 & 76.67 & 4.17 \\
Gemini 3.1 Pro & 112 & 19/90/3 & 83.04 & 2.68 \\
Claude Haiku 4.5 & 147 & 13/131/3 & 91.16 & 2.04 \\
Qwen3 Max & 86 & 35/48/3 & 59.30 & 3.49 \\
DeepSeek V3.1 & 46 & 14/31/1 & 69.57 & 2.17 \\
GLM 4.7 & 30 & 7/23/0 & 76.67 & 0.00 \\
MiniMax M2.1 & 68 & 23/43/2 & 66.18 & 2.94 \\
GPT-5.1 & 0 & 0/0/0 & N/A & N/A \\
GPT-5.2 & 161 & 77/76/8 & 52.17 & 4.97 \\
Qwen3 235B & 40 & 6/33/1 & 85.00 & 2.50 \\
Gemini 2.5 Flash & 84 & 20/63/1 & 76.19 & 1.19 \\
All & 1070 & 289/751/30 & 72.99 & 2.80 \\
\bottomrule
\end{tabularx}
\renewcommand{\thetable}{B5}
\captionof{table}{Joint results at annotated trading points. The sample scope matches main-text Tables~\ref{tab:2} and~\ref{tab:3}.}
\label{tab:B5}
\end{minipage}\par\addvspace{\floatsep}

\begin{multicols}{2}[][180pt]\raggedcolumns

\subsection{Complete Original Investment Performance}

\end{multicols}

\par\noindent\begin{minipage}{\textwidth}
\centering
\small
\begin{tabularx}{\textwidth}{@{}>{\raggedright\arraybackslash}p{0.2050\textwidth}>{\hsize=1.0000\hsize \raggedleft\arraybackslash}X>{\hsize=1.0000\hsize \raggedleft\arraybackslash}X>{\hsize=1.0000\hsize \raggedleft\arraybackslash}X>{\hsize=1.0000\hsize \raggedleft\arraybackslash}X>{\hsize=1.0000\hsize \raggedleft\arraybackslash}X>{\hsize=1.0000\hsize \raggedleft\arraybackslash}X>{\hsize=1.0000\hsize \raggedleft\arraybackslash}X@{}}
\toprule
\textbf{Model} & \textbf{CR\%} & \textbf{AR\%} & \textbf{AV\%} & \textbf{Sharpe} & \textbf{Sortino} & \textbf{MDD\%} & \textbf{Calmar} \\
\midrule
Claude Opus 4.5 & 10.84 & 37.21 & 18.64 & 1.79 & 3.31 & -7.98 & 4.66 \\
GLM 4.7 Flash & 8.48 & 28.41 & 14.24 & 1.83 & 3.48 & -5.83 & 4.87 \\
DeepSeek V3.2 & 7.50 & 24.88 & 12.75 & 1.81 & 3.97 & -3.92 & 6.34 \\
Gemini 3.1 Pro & 7.38 & 24.46 & 23.19 & 1.06 & 1.96 & -9.56 & 2.56 \\
Claude Haiku 4.5 & 6.00 & 19.60 & 18.77 & 1.05 & 1.48 & -9.08 & 2.16 \\
Qwen3 Max & 5.46 & 17.76 & 14.48 & 1.20 & 1.91 & -6.41 & 2.77 \\
DeepSeek V3.1 & 4.53 & 14.57 & 6.18 & 2.23 & 4.12 & -2.42 & 6.03 \\
GLM 4.7 & 2.64 & 8.33 & 10.96 & 0.78 & 1.28 & -7.66 & 1.09 \\
MiniMax M2.1 & 1.41 & 4.39 & 9.25 & 0.51 & 0.85 & -4.33 & 1.01 \\
GPT-5.1 & 0.00 & 0.00 & 0.00 & 0.00 & 0.00 & 0.00 & 0.00 \\
GPT-5.2 & -1.43 & -4.32 & 20.95 & -0.11 & -0.17 & -11.63 & -0.37 \\
Qwen3 235B & -1.85 & -5.57 & 7.59 & -0.72 & -0.94 & -4.58 & -1.22 \\
Gemini 2.5 Flash & -3.59 & -10.62 & 5.39 & -2.06 & -2.65 & -4.06 & -2.62 \\
\bottomrule
\end{tabularx}
\renewcommand{\thetable}{B6}
\captionof{table}{Complete investment performance. GPT-5.1 makes no trades, with zero returns and volatility; its risk ratios are recorded as zero by computational convention.}
\label{tab:B6}
\end{minipage}\par\addvspace{\floatsep}

\par\noindent\begin{minipage}{\textwidth}
\centering
\small
\begin{tabularx}{\textwidth}{@{}>{\raggedright\arraybackslash}p{0.1050\textwidth}>{\hsize=1.9070\hsize \raggedright\arraybackslash}X>{\hsize=0.6977\hsize \raggedleft\arraybackslash}X>{\hsize=0.6977\hsize \raggedleft\arraybackslash}X>{\hsize=0.6977\hsize \raggedleft\arraybackslash}X@{}}
\toprule
\textbf{Type} & \textbf{Model /\allowbreak{} Benchmark} & \textbf{CR\%} & \textbf{MDD\%} & \textbf{Sharpe} \\
\midrule
LLM & Claude Opus 4.5 & 10.84 & -7.98 & 1.79 \\
 & GLM 4.7 Flash & 8.48 & -5.83 & 1.83 \\
 & DeepSeek V3.2 & 7.50 & -3.92 & 1.81 \\
 & Gemini 3.1 Pro & 7.38 & -9.56 & 1.06 \\
 & Claude Haiku 4.5 & 6.00 & -9.08 & 1.05 \\
 & Qwen3 Max & 5.46 & -6.41 & 1.20 \\
 & DeepSeek V3.1 & 4.53 & -2.42 & 2.23 \\
 & GLM 4.7 & 2.64 & -7.66 & 0.78 \\
 & MiniMax M2.1 & 1.41 & -4.33 & 0.51 \\
 & GPT-5.1 & 0.00 & 0.00 & 0.00 \\
 & GPT-5.2 & -1.43 & -11.63 & -0.11 \\
 & Qwen3 235B & -1.85 & -4.58 & -0.72 \\
 & Gemini 2.5 Flash & -3.59 & -4.06 & -2.06 \\
\midrule
Market & Equal-weight buy-and-hold & 1.66 & -11.92 & 0.33 \\
 & SSE 50 & -1.87 & -8.34 & -0.24 \\
\midrule
Strategy & RSI & 3.25 & -2.25 & 1.45 \\
 & MACD & -1.85 & -6.02 & -0.42 \\
 & SMA & 3.90 & -10.77 & 0.86 \\
 & KDJ (closing price) & -5.50 & -7.96 & -1.58 \\
 & Mean reversion & -0.92 & -3.77 & -0.25 \\
\bottomrule
\end{tabularx}
\renewcommand{\thetable}{B6a}
\captionof{table}{Investment performance of models and reference benchmarks. CR and MDD are percentages; an MDD closer to zero indicates a smaller drawdown.}
\label{tab:B6a}
\end{minipage}\par\addvspace{\floatsep}

\par\noindent\begin{minipage}{\textwidth}
\centering
\small
\renewcommand{\arraystretch}{1.12}
\setlength{\tabcolsep}{4pt}
\begin{tabularx}{\textwidth}{@{}>{\raggedright\arraybackslash}p{0.2050\textwidth}>{\hsize=1.0000\hsize \raggedleft\arraybackslash}X>{\hsize=1.0000\hsize \raggedleft\arraybackslash}X>{\hsize=1.0000\hsize \raggedleft\arraybackslash}X@{}}
\toprule
\morecmidrules\cmidrule{1-4}
\textbf{Model /\allowbreak{} baseline} & \textbf{CR (\%)} & \textbf{Sharpe} & \textbf{Calmar} \\
\midrule
\rowcolor{TableBand}
Claude Opus 4.5 & 10.84 & 1.79 & 4.66 \\
GLM 4.7 Flash & 8.48 & 1.83 & 4.87 \\
\rowcolor{TableBand}
DeepSeek V3.2 & 7.50 & 1.81 & 6.34 \\
Gemini 3.1 Pro & 7.38 & 1.06 & 2.56 \\
\rowcolor{TableBand}
Claude Haiku 4.5 & 6.00 & 1.05 & 2.16 \\
Qwen3 Max & 5.46 & 1.20 & 2.77 \\
\rowcolor{TableBand}
DeepSeek V3.1 & 4.53 & 2.23 & 6.03 \\
GLM 4.7 & 2.64 & 0.78 & 1.09 \\
\rowcolor{TableBand}
MiniMax M2.1 & 1.41 & 0.51 & 1.01 \\
GPT-5.1 & 0.00 & 0.00 & 0.00 \\
\rowcolor{TableBand}
GPT-5.2 & -1.43 & -0.11 & -0.37 \\
Qwen3 235B & -1.85 & -0.72 & -1.22 \\
\rowcolor{TableBand}
Gemini 2.5 Flash & -3.59 & -2.06 & -2.62 \\
\rowcolor{TableAccent}
\textbf{Mean (13 models)} & \textbf{3.64} & \textbf{0.72} & \textbf{2.10} \\
\midrule
\rowcolor{TableBand}
Equal-weight B\&H & 1.66 & 0.33 & 0.43 \\
SSE 50 & -1.87 & -0.24 & -0.68 \\
\midrule
\rowcolor{TableBand}
RSI & 3.25 & 1.45 & 4.59 \\
MACD & -1.85 & -0.42 & -0.92 \\
\rowcolor{TableBand}
SMA & 3.90 & 0.86 & 1.16 \\
KDJ & -5.50 & -1.58 & -2.00 \\
\rowcolor{TableBand}
Mean reversion & -0.92 & -0.25 & -0.75 \\
\cmidrule{1-4}
\bottomrule
\end{tabularx}
\renewcommand{\thetable}{B6b}
\captionof{table}{Original account and benchmark performance. CR is reported as a percentage; Mean equally averages each metric across all 13 models. Full metrics appear in Appendix B.4.}
\label{tab:B6b}
\end{minipage}\par\addvspace{\floatsep}

\begin{multicols}{2}[][180pt]\raggedcolumns

\subsection{Temp Label Distribution and Boundary Cases}

\end{multicols}

\par\noindent\begin{minipage}{\textwidth}
\centering
\small
\begin{tabularx}{\textwidth}{@{}>{\raggedright\arraybackslash}p{0.2050\textwidth}>{\hsize=1.0000\hsize \raggedleft\arraybackslash}X>{\hsize=1.0000\hsize \raggedleft\arraybackslash}X>{\hsize=1.0000\hsize \raggedleft\arraybackslash}X@{}}
\toprule
\textbf{Label} & \textbf{Annotator A} & \textbf{Annotator B} & \textbf{Final consensus} \\
\midrule
Coherent & 1251 & 1189 & 1196 \\
Partially coherent & 15 & 77 & 70 \\
Incoherent & 0 & 0 & 0 \\
Insufficient materials; indeterminate & 27 & 27 & 27 \\
Semantically unresolved & 0 & 0 & 0 \\
\bottomrule
\end{tabularx}
\renewcommand{\thetable}{B7}
\captionof{table}{Temp label distribution for adjacent trade pairs. Each column contains 1,293 formal pairs; only coherent labels count as passing, and the other applicable labels count as gaps.}
\label{tab:B7}
\end{minipage}\par\addvspace{\floatsep}

\begin{multicols}{2}[][180pt]\raggedcolumns

The final consensus comprises 1,196 coherent, 70 partially coherent, and 27 unable-to-determine pairs. Both partially coherent and unable-to-determine cases count as gaps, giving a Temp gap rate of 7.50\% (97/1,293); unable-to-determine cases account for 2.09\%.

\end{multicols}

\par\noindent\begin{minipage}{\textwidth}
\centering
\small
\begin{tabularx}{\textwidth}{@{}>{\raggedright\arraybackslash}p{0.2050\textwidth}>{\hsize=1.0000\hsize \raggedleft\arraybackslash}X>{\hsize=1.0000\hsize \raggedleft\arraybackslash}X>{\hsize=1.0000\hsize \raggedleft\arraybackslash}X>{\hsize=1.0000\hsize \raggedleft\arraybackslash}X@{}}
\toprule
\textbf{A /\allowbreak{} B} & \textbf{Coherent} & \textbf{Partially coherent} & \textbf{Incoherent} & \textbf{Unable to determine} \\
\midrule
Coherent & 1184 & 67 & 0 & 0 \\
Partially coherent & 5 & 10 & 0 & 0 \\
Incoherent & 0 & 0 & 0 & 0 \\
Unable to determine & 0 & 0 & 0 & 27 \\
\bottomrule
\end{tabularx}
\renewcommand{\thetable}{B7a}
\captionof{table}{Expert A/B labels for Temp. All 1,293 pairs; rows denote A and columns B.}
\label{tab:B7a}
\end{minipage}\par\addvspace{\floatsep}

\begin{multicols}{2}[][180pt]\raggedcolumns

All 72 disagreements concern the boundary between coherent and partially coherent, with 60 occurring when the trading direction reverses. Joint review classifies 60 as partially coherent and 12 as coherent. In the original annotations, symmetric A/B F1 for the partially coherent category is only 0.217, reflecting substantial differences between the annotators' judgments at this boundary. Disagreements mainly concern how to interpret local problems in trading rationales: missing or unsupported core grounds for a transition affect the explanation linking adjacent trades; problems limited to incidental facts do not lower the coherence judgment when the main transition grounds remain supported.

\subsection{All-Point and Detailed Audit Results}

Of all 3,198 decision points, 912 have a gap in at least one assessed check, yielding an all-point joint gap rate of 28.52\%. Tables B8a and B8b summarize the complete decision trajectories without counting unannotated items as gaps; Table~\ref{tab:B9} retains statistics by applicable record, formally annotated trade event, and labeled adjacent trade pair. Main-text trading-point counts and joint gap counts appear in Table B5.

\end{multicols}

\par\noindent\begin{minipage}{\textwidth}
\centering
\small
\begin{tabularx}{\textwidth}{@{}>{\raggedright\arraybackslash}p{0.2550\textwidth}>{\hsize=1.0000\hsize \raggedleft\arraybackslash}X>{\hsize=1.0000\hsize \raggedleft\arraybackslash}X>{\hsize=1.0000\hsize \raggedleft\arraybackslash}X>{\hsize=1.0000\hsize \raggedleft\arraybackslash}X@{}}
\toprule
\textbf{Model} & \textbf{O} & \textbf{RK} & \textbf{B} & \textbf{Joint} \\
\midrule
Claude Opus 4.5 & 1.2 & 32.5 & 4.5 & 34.1 \\
GLM 4.7 Flash & 9.3 & 16.3 & 8.9 & 23.2 \\
DeepSeek V3.2 & 6.1 & 35.4 & 8.1 & 38.6 \\
Gemini 3.1 Pro & 6.5 & 35.0 & 8.1 & 38.2 \\
Claude Haiku 4.5 & 13.0 & 45.9 & 31.7 & 60.6 \\
Qwen3 Max & 0.8 & 19.5 & 5.7 & 20.7 \\
DeepSeek V3.1 & 0.0 & 13.0 & 0.8 & 13.0 \\
GLM 4.7 & 10.2 & 7.3 & 3.7 & 17.9 \\
MiniMax M2.1 & 3.3 & 16.7 & 5.3 & 20.3 \\
GPT-5.1 & 0.0 & 0.0 & 0.0 & 0.0 \\
GPT-5.2 & 4.1 & 32.1 & 0.0 & 34.6 \\
Qwen3 235B & 26.4 & 13.0 & 5.3 & 36.2 \\
Gemini 2.5 Flash & 11.4 & 23.6 & 6.1 & 33.3 \\
\bottomrule
\end{tabularx}
\renewcommand{\thetable}{B8a}
\captionof{table}{Dimensional and joint audit gap rates across all decision points (\%, \ensuremath{\downarrow}). Each model's 246 decision points form the denominator; each point is counted once per check.}
\label{tab:B8a}
\end{minipage}\par\addvspace{\floatsep}

\par\noindent\begin{minipage}{\textwidth}
\centering
\small
\begin{tabularx}{\textwidth}{@{}>{\raggedright\arraybackslash}p{0.2050\textwidth}>{\hsize=1.0000\hsize \raggedleft\arraybackslash}X>{\hsize=1.0000\hsize \raggedleft\arraybackslash}X>{\hsize=1.0000\hsize \raggedleft\arraybackslash}X>{\hsize=1.0000\hsize \raggedleft\arraybackslash}X>{\hsize=1.0000\hsize \raggedleft\arraybackslash}X>{\hsize=1.0000\hsize \raggedleft\arraybackslash}X>{\hsize=1.0000\hsize \raggedleft\arraybackslash}X>{\hsize=1.0000\hsize \raggedleft\arraybackslash}X@{}}
\toprule
\textbf{Model} & \textbf{Report} & \textbf{Rpt} & \textbf{Call} & \textbf{Resp} & \textbf{R} & \textbf{K} & \textbf{D} & \textbf{Temp} \\
\midrule
Claude Opus 4.5 & 0.0 & 1.2 & 1.2 & 1.2 & 22.0 & 22.0 & 3.7 & 1.2 \\
GLM 4.7 Flash & 0.0 & 6.5 & 4.5 & 0.4 & 9.3 & 12.6 & 6.5 & 4.5 \\
DeepSeek V3.2 & 1.2 & 5.7 & 0.8 & 0.4 & 14.2 & 30.9 & 4.1 & 5.7 \\
Gemini 3.1 Pro & 1.6 & 5.3 & 4.9 & 3.7 & 20.7 & 26.0 & 3.7 & 6.5 \\
Claude Haiku 4.5 & 0.0 & 13.0 & 7.7 & 7.7 & 33.7 & 26.0 & 29.3 & 4.5 \\
Qwen3 Max & 0.0 & 0.8 & 0.8 & 0.8 & 8.1 & 14.6 & 4.1 & 2.0 \\
DeepSeek V3.1 & 0.0 & 0.0 & 0.0 & 0.0 & 6.1 & 11.0 & 0.0 & 0.8 \\
GLM 4.7 & 9.8 & 9.8 & 0.4 & 0.0 & 3.3 & 5.3 & 3.3 & 1.2 \\
MiniMax M2.1 & 0.0 & 2.8 & 0.4 & 0.0 & 8.9 & 13.0 & 3.7 & 3.3 \\
GPT-5.1 & 0.0 & 0.0 & 0.0 & 0.0 & 0.0 & 0.0 & 0.0 & 0.0 \\
GPT-5.2 & 0.0 & 1.2 & 4.1 & 4.1 & 8.9 & 25.2 & 0.0 & 0.0 \\
Qwen3 235B & 0.0 & 25.6 & 5.3 & 4.1 & 5.7 & 9.8 & 2.4 & 3.7 \\
Gemini 2.5 Flash & 1.2 & 11.0 & 3.7 & 3.3 & 8.1 & 19.9 & 3.3 & 5.7 \\
\bottomrule
\end{tabularx}
\renewcommand{\thetable}{B8b}
\captionof{table}{Gap rates for the eight audit checks across all decision points (\%, \ensuremath{\downarrow}). Each model's 246 decision points form the denominator; each point is counted once per check.}
\label{tab:B8b}
\end{minipage}\par\addvspace{\floatsep}

\par\noindent\begin{minipage}{\textwidth}
\centering
\small
\begin{tabularx}{\textwidth}{@{}>{\raggedright\arraybackslash}p{0.2050\textwidth}>{\hsize=1.0000\hsize \raggedleft\arraybackslash}X>{\hsize=1.0000\hsize \raggedleft\arraybackslash}X>{\hsize=1.0000\hsize \raggedleft\arraybackslash}X>{\hsize=1.0000\hsize \raggedleft\arraybackslash}X>{\hsize=1.0000\hsize \raggedleft\arraybackslash}X>{\hsize=1.0000\hsize \raggedleft\arraybackslash}X>{\hsize=1.0000\hsize \raggedleft\arraybackslash}X>{\hsize=1.0000\hsize \raggedleft\arraybackslash}X>{\hsize=1.0000\hsize \raggedleft\arraybackslash}X@{}}
\toprule
\textbf{Model} & \textbf{Report} & \textbf{Rpt} & \textbf{Call} & \textbf{Resp} & \textbf{R} & \textbf{K} & \textbf{RK} & \textbf{D} & \textbf{Temp} \\
\midrule
Claude Opus 4.5 & 0.0 & 1.2 & 1.2 & 1.2 & 35.5 & 39.4 & 60.0 & 7.8 & 2.0 \\
GLM 4.7 Flash & 0.0 & 6.5 & 4.5 & 0.4 & 29.9 & 46.0 & 63.2 & 56.2 & 13.9 \\
DeepSeek V3.2 & 1.2 & 5.7 & 0.8 & 0.4 & 20.7 & 53.6 & 63.7 & 12.5 & 8.1 \\
Gemini 3.1 Pro & 1.6 & 5.3 & 4.9 & 3.7 & 33.0 & 43.7 & 63.5 & 6.7 & 8.3 \\
Claude Haiku 4.5 & 0.0 & 13.0 & 7.7 & 7.7 & 49.7 & 39.3 & 73.2 & 42.7 & 6.9 \\
Qwen3 Max & 0.0 & 0.8 & 0.8 & 0.8 & 21.3 & 43.6 & 56.4 & 20.4 & 6.0 \\
DeepSeek V3.1 & 0.0 & 0.0 & 0.0 & 0.0 & 29.4 & 52.9 & 62.7 & 0.0 & 4.7 \\
GLM 4.7 & 9.8 & 9.8 & 0.4 & 0.0 & 24.2 & 39.4 & 54.5 & 30.8 & 11.5 \\
MiniMax M2.1 & 0.0 & 2.8 & 0.4 & 0.0 & 29.8 & 39.3 & 53.6 & 18.0 & 10.7 \\
GPT-5.1 & 0.0 & 0.0 & 0.0 & 0.0 & N/A & N/A & N/A & N/A & N/A \\
GPT-5.2 & 0.0 & 1.2 & 4.1 & 4.1 & 12.6 & 36.6 & 46.3 & 0.0 & 0.0 \\
Qwen3 235B & 0.0 & 25.6 & 5.3 & 4.1 & 29.2 & 56.2 & 72.9 & 37.5 & 22.5 \\
Gemini 2.5 Flash & 1.2 & 11.0 & 3.7 & 3.3 & 21.4 & 54.1 & 65.3 & 17.0 & 15.2 \\
\bottomrule
\end{tabularx}
\renewcommand{\thetable}{B9}
\captionof{table}{Detailed audit gap rates by applicable record, trade event, and trade pair (\%, \ensuremath{\downarrow}). Report uses scheduled windows; Rpt, Call, and Resp use their applicable records; R, K, and RK use trade events; D uses events involving account assessments; Temp uses adjacent trade pairs. N/A denotes no applicable samples.}
\label{tab:B9}
\end{minipage}\par\addvspace{\floatsep}

\begin{multicols}{2}[][180pt]\raggedcolumns

\FloatBarrier

\section{Historical Replay and Ablation Analysis}

\subsection{Execution and Valuation}

The main-experiment simulator uses current snapshot prices. Orders that pass the checks execute for the full requested quantity; rejected orders do not change cash or positions. Execution is immediate fill or rejection, without pending orders, partial fills, or cancellation of resting orders. Execution auditing aggregates the share quantities in fill records and compares them with position changes in the same window. Slippage costs are analyzed separately in C.6.

The simulated ledger updates cash and positions through fill records and updates sellable quantities across trading days; dividends and stock distributions are not entered as separate corporate-action ledger events.

All main-experiment replays start with CNY 1 million and use shared historical prices. Sales precede purchases within each window, with same-direction trades ordered by security. Retained actions keep their original net quantities as targets. Purchases are sized to available cash, 100-share lots, and fees; sales are sized to existing positions and T+1 sellable quantities. Accounts are continuously marked to market, and positions still open at the end are valued at historical prices. Each trade incurs a 0.03\% commission with a minimum of CNY 5, plus a 0.05\% stamp duty on sales.

All filters remove trades with failed or unable-to-determine checks. O removes all net trades at a decision point when any of its four record checks has a gap; RK takes the union of R/K removal sets; B removes 164 D gap actions or 93 Temp gap actions, with 32 overlapping actions and a union of 225. A Temp gap is assigned to the later trade in the labeled pair. No-account-assessment and first-trade cases do not independently trigger removal.

Of the 1,293 Temp pairs, 1,257 correspond to formal R/K/D actions: 1,164 are coherent, 66 partially coherent, and 27 unable to determine. The latter two categories give 93 removals. Another 36 pairs correspond to calibration actions, including four partially coherent pairs counted only in point-level audits. Increases and decreases require \ensuremath{\Delta}CR above +0.01 or below -0.01 percentage points, respectively; all other changes are classified as unchanged. Main results cover 12 actively trading models.

\subsection{Complete Three-Dimensional Ablations for Each Model}

\end{multicols}

\par\noindent\begin{minipage}{\textwidth}
\centering
\small
\begin{tabularx}{\textwidth}{@{}>{\raggedright\arraybackslash}p{0.2550\textwidth}>{\hsize=1.0000\hsize \raggedleft\arraybackslash}X>{\hsize=1.0000\hsize \raggedleft\arraybackslash}X>{\hsize=1.0000\hsize \raggedleft\arraybackslash}X>{\hsize=1.0000\hsize \raggedleft\arraybackslash}X@{}}
\toprule
\textbf{Model} & \textbf{Original} & \textbf{O} & \textbf{RK} & \textbf{B} \\
\midrule
Claude Opus 4.5 & 10.84 & 13.68 & 2.04 & 14.54 \\
GLM 4.7 Flash & 8.48 & 3.04 & 27.53 & 7.70 \\
DeepSeek V3.2 & 7.50 & 11.52 & 36.19 & 20.19 \\
Gemini 3.1 Pro & 7.38 & 8.26 & 32.00 & 5.84 \\
Claude Haiku 4.5 & 6.00 & 7.68 & 10.67 & 27.02 \\
Qwen3 Max & 5.46 & 6.71 & 4.39 & 9.08 \\
DeepSeek V3.1 & 4.53 & 4.53 & 11.59 & 4.39 \\
GLM 4.7 & 2.64 & 2.66 & 8.59 & 5.14 \\
MiniMax M2.1 & 1.41 & 12.42 & 8.61 & 8.26 \\
GPT-5.1 & 0.00 & 0.00 & 0.00 & 0.00 \\
GPT-5.2 & -1.43 & 0.96 & -3.32 & -1.43 \\
Qwen3 235B & -1.85 & 23.19 & 13.56 & 13.79 \\
Gemini 2.5 Flash & -3.59 & -3.50 & 1.89 & 6.17 \\
\bottomrule
\end{tabularx}
\renewcommand{\thetable}{C1}
\captionof{table}{Original CR and CR after single-dimension filtering (\%). O, RK, and B represent execution traceability, evidence sufficiency, and financial meta-awareness, respectively.}
\label{tab:C1}
\end{minipage}\par\addvspace{\floatsep}

\par\noindent\begin{minipage}{\textwidth}
\centering
\small
\begin{tabularx}{\textwidth}{@{}>{\raggedright\arraybackslash}p{0.2550\textwidth}>{\hsize=1.0000\hsize \raggedleft\arraybackslash}X>{\hsize=1.0000\hsize \raggedleft\arraybackslash}X>{\hsize=1.0000\hsize \raggedleft\arraybackslash}X>{\hsize=1.0000\hsize \raggedleft\arraybackslash}X@{}}
\toprule
\textbf{Model} & \textbf{O+RK} & \textbf{O+B} & \textbf{RK+B} & \textbf{O+RK+B} \\
\midrule
Claude Opus 4.5 & 2.04 & 19.03 & 8.79 & 8.79 \\
GLM 4.7 Flash & 22.48 & -2.45 & 24.25 & 17.42 \\
DeepSeek V3.2 & 36.23 & 18.02 & 37.75 & 37.80 \\
Gemini 3.1 Pro & 30.48 & 9.34 & 19.48 & 16.12 \\
Claude Haiku 4.5 & 17.51 & 32.59 & 11.38 & 12.77 \\
Qwen3 Max & 5.13 & 8.09 & 4.76 & 4.76 \\
DeepSeek V3.1 & 11.59 & 4.39 & 11.59 & 11.59 \\
GLM 4.7 & 9.16 & 4.67 & 7.85 & 8.42 \\
MiniMax M2.1 & 8.79 & 8.68 & 8.84 & 8.69 \\
GPT-5.1 & 0.00 & 0.00 & 0.00 & 0.00 \\
GPT-5.2 & -3.50 & 0.96 & -3.32 & -3.50 \\
Qwen3 235B & 14.25 & 24.37 & 12.03 & 12.73 \\
Gemini 2.5 Flash & 1.30 & 7.00 & 3.02 & 3.55 \\
\bottomrule
\end{tabularx}
\renewcommand{\thetable}{C2}
\captionof{table}{CR after pairwise and three-dimensional joint filtering (\%). Each cell is an independent replay result.}
\label{tab:C2}
\end{minipage}\par\addvspace{\floatsep}

\par\noindent\begin{minipage}{\textwidth}
\centering
\small
\begin{tabularx}{\textwidth}{@{}>{\raggedright\arraybackslash}p{0.1400\textwidth}>{\hsize=0.9000\hsize \raggedleft\arraybackslash}X>{\hsize=1.4000\hsize \raggedleft\arraybackslash}X>{\hsize=0.9000\hsize \raggedleft\arraybackslash}X>{\hsize=0.9000\hsize \raggedleft\arraybackslash}X>{\hsize=0.9000\hsize \raggedleft\arraybackslash}X@{}}
\toprule
\textbf{Filter} & \textbf{Mean CR \%} & \textbf{Buys/\allowbreak{}sells removed} & \textbf{Sharpe} & \textbf{Sortino} & \textbf{Calmar} \\
\midrule
Original & 3.95 & 0/0 & 0.78 & 1.55 & 2.27 \\
R & 10.60 & 126/273 & 1.01 & 1.53 & 2.81 \\
K & 13.23 & 201/412 & 1.42 & 2.35 & 4.56 \\
D & 8.43 & 41/123 & 1.12 & 1.91 & 2.84 \\
Temp & 7.40 & 36/57 & 1.25 & 2.11 & 3.31 \\
O & 7.60 & 58/74 & 0.90 & 1.49 & 2.38 \\
RK & 12.81 & 286/563 & 1.39 & 2.37 & 4.95 \\
B & 10.06 & 68/157 & 1.22 & 1.98 & 3.12 \\
O+RK & 12.96 & 308/585 & 1.35 & 2.25 & 4.39 \\
O+B & 11.22 & 115/209 & 1.16 & 1.88 & 3.03 \\
RK+B & 12.20 & 315/593 & 1.37 & 2.25 & 4.33 \\
O+RK+B & 11.59 & 337/611 & 1.30 & 2.07 & 3.62 \\
\bottomrule
\end{tabularx}
\renewcommand{\thetable}{C3}
\captionof{table}{Cumulative returns, trade removals, and risk-adjusted returns for submetrics and combinations. CR and risk-adjusted return metrics are equally weighted means over 12 active models. Higher Sharpe, Sortino, and Calmar values are better.}
\label{tab:C3}
\end{minipage}\par\addvspace{\floatsep}

\begin{multicols}{2}[][180pt]\raggedcolumns

R/K construction, filtering analysis, and direction-matched random controls use the same historical trades, including exploratory analysis on that dataset and no separate held-out set. Results describe portfolio changes along fixed trading trajectories.

\subsection{Replay Checks for the New Model}

GPT-6 Sol's 96 executed trades yield an original CR of 4.85\%. RK flags 41 trades: 13 purchases and 28 sales, including 36 confirmed gaps and five unable-to-determine cases. The 28 flagged sales make exits a prominent part of the intervention. O, D, and Temp have no gaps, so their separate filters leave CR unchanged.

Decision-level RK filtering replaces flagged trades with No Trade and replays the others at their original times and target quantities under updated account constraints. Removing 41 trades prevents 22 additional orders from executing; 33 trades execute and CR reaches 38.28\%. This result reflects the new cash and holding path, not the isolated payoffs of removed trades. Filtering the 36 confirmed gaps alone raises CR by 23.91 percentage points; SH688008 contributes a 28.50-point increase, exceeding the net portfolio change.

\subsection{Risk-Adjusted Return Calculations and Complete Results}

All main-experiment filtering conditions use daily 14:00 account assets and initial capital of CNY 1 million to calculate 82 daily returns, with an annualization factor of 252 and zero risk-free and downside target returns. Sharpe is the mean daily return divided by the sample standard deviation of daily returns, multiplied by \(\sqrt{252}\); Sortino replaces the denominator with the square root of the mean squared negative returns over all trading days; Calmar is the evaluation period's annualized return divided by the absolute daily maximum drawdown. Metrics are calculated for each model first, then equally averaged over 12 models.

Ratio increases and decreases use unrounded values, with absolute differences no greater than \(10^{-8}\) treated as unchanged; return changes in percentage points retain the 0.01 threshold.

\end{multicols}

\par\noindent\begin{minipage}{\textwidth}
\centering
\small
\begin{tabularx}{\textwidth}{@{}>{\raggedright\arraybackslash}p{0.1050\textwidth}>{\hsize=1.0000\hsize \raggedleft\arraybackslash}X>{\hsize=1.0000\hsize \raggedleft\arraybackslash}X>{\hsize=1.0000\hsize \raggedleft\arraybackslash}X>{\hsize=1.0000\hsize \raggedleft\arraybackslash}X>{\hsize=1.0000\hsize \raggedleft\arraybackslash}X@{}}
\toprule
\textbf{Filter} & \textbf{CR\%} & \textbf{AR\%} & \textbf{Sharpe} & \textbf{Sortino} & \textbf{Calmar} \\
\midrule
original & 3.95 & 13.26 & 0.78 & 1.55 & 2.27 \\
R & 10.60 & 39.88 & 1.01 & 1.53 & 2.81 \\
K & 13.23 & 49.70 & 1.42 & 2.35 & 4.56 \\
D & 8.43 & 30.76 & 1.12 & 1.91 & 2.84 \\
Temp & 7.40 & 25.12 & 1.25 & 2.11 & 3.31 \\
O & 7.60 & 26.81 & 0.90 & 1.49 & 2.38 \\
RK & 12.81 & 50.25 & 1.39 & 2.37 & 4.95 \\
B & 10.06 & 36.21 & 1.22 & 1.98 & 3.12 \\
O+RK & 12.96 & 50.31 & 1.35 & 2.25 & 4.39 \\
O+B & 11.22 & 42.19 & 1.16 & 1.88 & 3.03 \\
RK+B & 12.20 & 46.41 & 1.37 & 2.25 & 4.33 \\
O+RK+B & 11.59 & 43.54 & 1.30 & 2.07 & 3.62 \\
\bottomrule
\end{tabularx}
\renewcommand{\thetable}{C4}
\captionof{table}{Complete mean financial metrics for each filtering condition. Higher values are better for all metrics.}
\label{tab:C4}
\end{minipage}\par\addvspace{\floatsep}

\par\noindent\begin{minipage}{\textwidth}
\centering
\small
\begin{tabularx}{\textwidth}{@{}>{\raggedright\arraybackslash}p{0.2050\textwidth}>{\hsize=1.0000\hsize \raggedleft\arraybackslash}X>{\hsize=1.0000\hsize \raggedleft\arraybackslash}X>{\hsize=1.0000\hsize \raggedleft\arraybackslash}X@{}}
\toprule
\textbf{Filter} & \textbf{Sharpe} & \textbf{Sortino} & \textbf{Calmar} \\
\midrule
original & 0/0/12 & 0/0/12 & 0/0/12 \\
R & 8/4/0 & 8/4/0 & 8/4/0 \\
K & 9/3/0 & 8/4/0 & 9/3/0 \\
D & 7/3/2 & 7/3/2 & 7/3/2 \\
Temp & 6/5/1 & 7/4/1 & 7/4/1 \\
O & 6/5/1 & 6/5/1 & 6/5/1 \\
RK & 8/4/0 & 7/5/0 & 8/4/0 \\
B & 7/4/1 & 6/5/1 & 6/5/1 \\
O+RK & 9/3/0 & 7/5/0 & 9/3/0 \\
O+B & 8/4/0 & 7/5/0 & 7/5/0 \\
RK+B & 8/4/0 & 7/5/0 & 8/4/0 \\
O+RK+B & 8/4/0 & 6/6/0 & 9/3/0 \\
\bottomrule
\end{tabularx}
\renewcommand{\thetable}{C5}
\captionof{table}{Numbers of models with improved risk-adjusted returns. Each cell shows increased/decreased/unchanged counts, summing to 12.}
\label{tab:C5}
\end{minipage}\par\addvspace{\floatsep}

\par\noindent\begin{minipage}{\textwidth}
\centering\small
\setlength{\tabcolsep}{3pt}
\begin{tabularx}{\textwidth}{@{}>{\raggedright\arraybackslash}p{.22\textwidth}>{\raggedleft\arraybackslash}X>{\raggedleft\arraybackslash}X>{\raggedleft\arraybackslash}X>{\raggedleft\arraybackslash}X>{\raggedleft\arraybackslash}X>{\raggedleft\arraybackslash}X>{\raggedleft\arraybackslash}X>{\raggedleft\arraybackslash}X>{\raggedleft\arraybackslash}X@{}}
\toprule
\textbf{Model} & \multicolumn{3}{c}{\textbf{Original}} & \multicolumn{3}{c}{\textbf{R}} & \multicolumn{3}{c}{\textbf{K}} \\
\cmidrule(lr){2-4}\cmidrule(lr){5-7}\cmidrule(lr){8-10}
 & \textbf{Sharpe} & \textbf{Sortino} & \textbf{Calmar} & \textbf{Sharpe} & \textbf{Sortino} & \textbf{Calmar} & \textbf{Sharpe} & \textbf{Sortino} & \textbf{Calmar} \\
\midrule
\rowcolor{TableBand}
Claude Haiku 4.5 & 1.05 & 1.48 & 2.16 & 1.49 & 2.07 & 3.69 & 1.23 & 1.58 & 2.71 \\
Claude Opus 4.5 & 1.79 & 3.31 & 4.66 & 0.60 & 0.92 & 0.89 & 1.51 & 2.55 & 3.17 \\
\rowcolor{TableBand}
DeepSeek V3.1 & 2.23 & 4.12 & 6.03 & 0.21 & 0.31 & 0.25 & 2.40 & 3.54 & 9.07 \\
DeepSeek V3.2 & 1.81 & 3.97 & 6.34 & 1.38 & 1.78 & 3.54 & 2.39 & 4.46 & 7.93 \\
\rowcolor{TableBand}
Gemini 2.5 Flash & -2.06 & -2.65 & -2.62 & -0.59 & -0.79 & -0.95 & 1.17 & 1.95 & 1.95 \\
Gemini 3.1 Pro & 1.06 & 1.96 & 2.56 & 1.17 & 2.11 & 2.93 & 2.12 & 3.64 & 9.53 \\
\rowcolor{TableBand}
GPT-5.2 & -0.11 & -0.17 & -0.37 & 1.29 & 1.85 & 2.77 & -0.54 & -0.77 & -0.81 \\
MiniMax M2.1 & 0.51 & 0.85 & 1.01 & -0.02 & -0.02 & -0.55 & 0.96 & 1.25 & 1.75 \\
\rowcolor{TableBand}
Qwen3 235B & -0.72 & -0.94 & -1.22 & 1.69 & 2.31 & 5.74 & 1.24 & 1.65 & 3.16 \\
Qwen3 Max & 1.20 & 1.91 & 2.77 & 1.33 & 1.97 & 3.84 & 0.38 & 0.56 & 0.38 \\
\rowcolor{TableBand}
GLM 4.7 & 0.78 & 1.28 & 1.09 & 1.65 & 2.26 & 5.39 & 1.77 & 3.64 & 4.60 \\
GLM 4.7 Flash & 1.83 & 3.48 & 4.87 & 1.88 & 3.52 & 6.21 & 2.44 & 4.22 & 11.27 \\
\bottomrule
\end{tabularx}
\renewcommand{\thetable}{C6}
\captionof{table}{Per-model risk-adjusted returns. Each filtering condition is replayed independently, retaining original results for comparison.}
\label{tab:C6}
\end{minipage}\par\addvspace{\floatsep}
\par\noindent\begin{minipage}{\textwidth}
\centering\small
\setlength{\tabcolsep}{3pt}
\begin{tabularx}{\textwidth}{@{}>{\raggedright\arraybackslash}p{.22\textwidth}>{\raggedleft\arraybackslash}X>{\raggedleft\arraybackslash}X>{\raggedleft\arraybackslash}X>{\raggedleft\arraybackslash}X>{\raggedleft\arraybackslash}X>{\raggedleft\arraybackslash}X>{\raggedleft\arraybackslash}X>{\raggedleft\arraybackslash}X>{\raggedleft\arraybackslash}X@{}}
\toprule
\textbf{Model} & \multicolumn{3}{c}{\textbf{D}} & \multicolumn{3}{c}{\textbf{Temp}} & \multicolumn{3}{c}{\textbf{O}} \\
\cmidrule(lr){2-4}\cmidrule(lr){5-7}\cmidrule(lr){8-10}
 & \textbf{Sharpe} & \textbf{Sortino} & \textbf{Calmar} & \textbf{Sharpe} & \textbf{Sortino} & \textbf{Calmar} & \textbf{Sharpe} & \textbf{Sortino} & \textbf{Calmar} \\
\midrule
\rowcolor{TableBand}
Claude Haiku 4.5 & 1.25 & 1.62 & 2.99 & 2.47 & 4.93 & 8.74 & 0.91 & 1.18 & 1.72 \\
Claude Opus 4.5 & 2.14 & 4.02 & 6.36 & 1.79 & 3.49 & 5.23 & 1.98 & 3.78 & 5.75 \\
\rowcolor{TableBand}
DeepSeek V3.1 & 2.23 & 4.12 & 6.03 & 2.17 & 3.95 & 5.83 & 2.23 & 4.12 & 6.03 \\
DeepSeek V3.2 & 2.55 & 5.07 & 7.32 & 1.54 & 2.45 & 5.83 & 1.54 & 2.46 & 4.91 \\
\rowcolor{TableBand}
Gemini 2.5 Flash & 0.79 & 1.02 & 1.53 & 0.79 & 1.03 & 1.48 & -1.97 & -2.56 & -2.58 \\
Gemini 3.1 Pro & 1.62 & 2.41 & 4.28 & 0.90 & 1.45 & 2.15 & 0.95 & 1.54 & 2.28 \\
\rowcolor{TableBand}
GPT-5.2 & -0.11 & -0.17 & -0.37 & -0.11 & -0.17 & -0.37 & 0.24 & 0.37 & 0.16 \\
MiniMax M2.1 & -0.59 & -0.86 & -0.80 & 1.01 & 1.32 & 2.01 & 1.06 & 1.40 & 2.23 \\
\rowcolor{TableBand}
Qwen3 235B & 0.04 & 0.06 & -0.03 & 1.00 & 1.30 & 2.00 & 1.48 & 1.97 & 4.33 \\
Qwen3 Max & 1.03 & 1.45 & 1.63 & 1.32 & 2.09 & 3.24 & 0.91 & 1.43 & 1.48 \\
\rowcolor{TableBand}
GLM 4.7 & 1.16 & 1.98 & 2.21 & 0.94 & 1.53 & 1.30 & 0.82 & 1.32 & 1.10 \\
GLM 4.7 Flash & 1.29 & 2.26 & 2.87 & 1.11 & 1.93 & 2.34 & 0.65 & 0.83 & 1.20 \\
\bottomrule
\end{tabularx}
\renewcommand{\thetable}{C6}
\captionof{table}{Per-model risk-adjusted returns (continued).}
\end{minipage}\par\addvspace{\floatsep}
\par\noindent\begin{minipage}{\textwidth}
\centering\small
\setlength{\tabcolsep}{3pt}
\begin{tabularx}{\textwidth}{@{}>{\raggedright\arraybackslash}p{.22\textwidth}>{\raggedleft\arraybackslash}X>{\raggedleft\arraybackslash}X>{\raggedleft\arraybackslash}X>{\raggedleft\arraybackslash}X>{\raggedleft\arraybackslash}X>{\raggedleft\arraybackslash}X>{\raggedleft\arraybackslash}X>{\raggedleft\arraybackslash}X>{\raggedleft\arraybackslash}X@{}}
\toprule
\textbf{Model} & \multicolumn{3}{c}{\textbf{RK}} & \multicolumn{3}{c}{\textbf{B}} & \multicolumn{3}{c}{\textbf{O+RK}} \\
\cmidrule(lr){2-4}\cmidrule(lr){5-7}\cmidrule(lr){8-10}
 & \textbf{Sharpe} & \textbf{Sortino} & \textbf{Calmar} & \textbf{Sharpe} & \textbf{Sortino} & \textbf{Calmar} & \textbf{Sharpe} & \textbf{Sortino} & \textbf{Calmar} \\
\midrule
\rowcolor{TableBand}
Claude Haiku 4.5 & 0.91 & 1.15 & 1.66 & 1.60 & 2.22 & 5.19 & 1.17 & 1.51 & 2.62 \\
Claude Opus 4.5 & 0.36 & 0.51 & 0.31 & 1.86 & 3.38 & 5.31 & 0.36 & 0.51 & 0.31 \\
\rowcolor{TableBand}
DeepSeek V3.1 & 3.39 & 6.61 & 11.39 & 2.17 & 3.95 & 5.83 & 3.39 & 6.61 & 11.39 \\
DeepSeek V3.2 & 1.97 & 2.74 & 7.94 & 2.27 & 3.94 & 7.44 & 1.96 & 2.73 & 7.96 \\
\rowcolor{TableBand}
Gemini 2.5 Flash & 0.39 & 0.59 & 0.45 & 0.77 & 1.00 & 1.42 & 0.30 & 0.45 & 0.31 \\
Gemini 3.1 Pro & 2.39 & 4.60 & 12.08 & 0.79 & 1.24 & 1.73 & 2.46 & 4.85 & 10.81 \\
\rowcolor{TableBand}
GPT-5.2 & -0.25 & -0.34 & -0.52 & -0.11 & -0.17 & -0.37 & -0.30 & -0.40 & -0.55 \\
MiniMax M2.1 & 0.75 & 0.96 & 1.13 & 0.73 & 0.96 & 1.01 & 0.76 & 0.98 & 1.16 \\
\rowcolor{TableBand}
Qwen3 235B & 1.39 & 1.86 & 3.88 & 1.13 & 1.50 & 2.47 & 1.44 & 1.94 & 4.13 \\
Qwen3 Max & 0.54 & 0.74 & 0.63 & 1.08 & 1.54 & 1.82 & 0.59 & 0.82 & 0.73 \\
\rowcolor{TableBand}
GLM 4.7 & 2.01 & 3.84 & 5.85 & 1.16 & 1.98 & 2.21 & 2.13 & 4.18 & 6.28 \\
GLM 4.7 Flash & 2.86 & 5.19 & 14.63 & 1.21 & 2.17 & 3.36 & 1.95 & 2.79 & 7.55 \\
\bottomrule
\end{tabularx}
\renewcommand{\thetable}{C6}
\captionof{table}{Per-model risk-adjusted returns (continued).}
\end{minipage}\par\addvspace{\floatsep}
\par\noindent\begin{minipage}{\textwidth}
\centering\small
\setlength{\tabcolsep}{3pt}
\begin{tabularx}{\textwidth}{@{}>{\raggedright\arraybackslash}p{.22\textwidth}>{\raggedleft\arraybackslash}X>{\raggedleft\arraybackslash}X>{\raggedleft\arraybackslash}X>{\raggedleft\arraybackslash}X>{\raggedleft\arraybackslash}X>{\raggedleft\arraybackslash}X>{\raggedleft\arraybackslash}X>{\raggedleft\arraybackslash}X>{\raggedleft\arraybackslash}X@{}}
\toprule
\textbf{Model} & \multicolumn{3}{c}{\textbf{O+B}} & \multicolumn{3}{c}{\textbf{RK+B}} & \multicolumn{3}{c}{\textbf{O+RK+B}} \\
\cmidrule(lr){2-4}\cmidrule(lr){5-7}\cmidrule(lr){8-10}
 & \textbf{Sharpe} & \textbf{Sortino} & \textbf{Calmar} & \textbf{Sharpe} & \textbf{Sortino} & \textbf{Calmar} & \textbf{Sharpe} & \textbf{Sortino} & \textbf{Calmar} \\
\midrule
\rowcolor{TableBand}
Claude Haiku 4.5 & 1.67 & 2.31 & 5.63 & 0.98 & 1.27 & 2.04 & 1.08 & 1.40 & 2.46 \\
Claude Opus 4.5 & 2.09 & 4.03 & 6.37 & 0.97 & 1.45 & 1.57 & 0.97 & 1.45 & 1.57 \\
\rowcolor{TableBand}
DeepSeek V3.1 & 2.17 & 3.95 & 5.83 & 3.39 & 6.61 & 11.39 & 3.39 & 6.61 & 11.39 \\
DeepSeek V3.2 & 1.99 & 3.33 & 5.88 & 2.03 & 2.83 & 8.43 & 2.02 & 2.82 & 8.44 \\
\rowcolor{TableBand}
Gemini 2.5 Flash & 0.85 & 1.11 & 1.64 & 0.56 & 0.86 & 0.74 & 0.64 & 0.99 & 0.89 \\
Gemini 3.1 Pro & 1.04 & 1.78 & 2.41 & 1.79 & 3.16 & 5.74 & 1.65 & 2.87 & 4.00 \\
\rowcolor{TableBand}
GPT-5.2 & 0.24 & 0.37 & 0.16 & -0.25 & -0.34 & -0.52 & -0.30 & -0.40 & -0.55 \\
MiniMax M2.1 & 0.75 & 0.99 & 1.07 & 0.75 & 0.99 & 1.11 & 0.74 & 0.98 & 1.08 \\
\rowcolor{TableBand}
Qwen3 235B & 1.52 & 2.03 & 4.59 & 1.29 & 1.70 & 3.47 & 1.35 & 1.78 & 3.70 \\
Qwen3 Max & 0.87 & 1.29 & 1.40 & 0.56 & 0.77 & 0.67 & 0.56 & 0.77 & 0.67 \\
\rowcolor{TableBand}
GLM 4.7 & 1.07 & 1.81 & 2.00 & 1.70 & 3.01 & 3.99 & 1.81 & 3.28 & 4.30 \\
GLM 4.7 Flash & -0.36 & -0.44 & -0.68 & 2.67 & 4.71 & 13.28 & 1.66 & 2.29 & 5.47 \\
\bottomrule
\end{tabularx}
\renewcommand{\thetable}{C6}
\captionof{table}{Per-model risk-adjusted returns (continued).}
\end{minipage}\par\addvspace{\floatsep}

\par\noindent\begin{minipage}{\textwidth}
\centering
\small
\begin{tabularx}{\textwidth}{@{}>{\raggedright\arraybackslash}p{0.205\textwidth}*{4}{>{\centering\arraybackslash}X}@{}}
\toprule
\textbf{Condition} & \textbf{CR\%} & \textbf{Sharpe} & \textbf{Sortino} & \textbf{Calmar} \\
\midrule
Original & 4.85 & 1.02 & 1.88 & 1.94 \\
RK & 38.28 & 2.86 & 5.20 & 12.84 \\
\bottomrule
\end{tabularx}
\renewcommand{\thetable}{C7}
\captionof{table}{Daily return and risk-adjusted metrics for GPT-6 Sol.}
\label{tab:C7}
\end{minipage}\par\addvspace{\floatsep}

\begin{multicols}{2}[][180pt]\raggedcolumns

\subsection{Confirmed Errors and Random Filtering}

Confirmed-error-only replay retains trades labeled unable to determine. O removes 121 trades with confirmed record-verification discrepancies; RK removes 803 trades with confirmed R or K gaps; B removes trades with confirmed D errors or Temp non-passes, including partially coherent cases, for a union of 190 trades. Combinations take the union of removal sets.

Random filtering samples each model's formal trades over the full March-to-June period, separately matching the numbers of purchases and sales removed. Sampling is without replacement and repeated 1,000 times; remaining trades are replayed under the same cash, T+1, lot-size, and fee rules. Table C12 reports mean return changes across the 12 actively trading models and the random distributions.

\end{multicols}

\par\noindent\begin{minipage}{\textwidth}
\centering
\small
\begin{tabularx}{\textwidth}{@{}>{\raggedright\arraybackslash}p{0.16\textwidth}>{\raggedright\arraybackslash}p{0.12\textwidth}*{5}{>{\centering\arraybackslash}X}@{}}
\toprule
\textbf{Setting} & \textbf{Filter} & \textbf{CR\%} & \textbf{AR\%} & \textbf{Sharpe} & \textbf{Sortino} & \textbf{Calmar} \\
\midrule
Confirmed errors only & O & 7.45 & 26.28 & 0.87 & 1.45 & 2.33 \\
Confirmed errors only & RK & 13.08 & 48.78 & 1.45 & 2.39 & 4.47 \\
Confirmed errors only & B & 8.57 & 31.87 & 1.08 & 1.81 & 2.76 \\
Confirmed errors only & O + RK + B & 13.33 & 49.75 & 1.42 & 2.24 & 3.84 \\
\bottomrule
\end{tabularx}
\renewcommand{\thetable}{C8}
\captionof{table}{Selected mean financial metrics for confirmed-error-only filtering. All values are equally weighted means over the 12 actively trading models.}
\label{tab:C8}
\end{minipage}\par\addvspace{\floatsep}

\begin{multicols}{2}[][180pt]\raggedcolumns

\subsection{Slippage Sensitivity}

Holding labels, trading times, and target quantities fixed, we set slippage at 0, 5, 10, 20, and 50 basis points per side. One basis point equals 0.01\%; the purchase execution price is the snapshot price multiplied by \(1+b/10000\), and the sale execution price is the snapshot price multiplied by \(1-b/10000\), where \(b\) denotes slippage in basis points per side. Account valuation continues to use snapshot prices, while cash, T+1, 100-share lot, and fee rules remain unchanged. Each filtered account is compared with an unfiltered account under the same slippage setting.

\end{multicols}

\par\noindent\begin{minipage}{\textwidth}
\begingroup
\centering
\small
\begin{tabularx}{\textwidth}{@{}>{\raggedright\arraybackslash}p{0.1050\textwidth}>{\hsize=1.0000\hsize \raggedleft\arraybackslash}X>{\hsize=1.0000\hsize \raggedleft\arraybackslash}X>{\hsize=1.0000\hsize \raggedleft\arraybackslash}X>{\hsize=1.0000\hsize \raggedleft\arraybackslash}X>{\hsize=1.0000\hsize \raggedleft\arraybackslash}X>{\hsize=1.0000\hsize \raggedleft\arraybackslash}X>{\hsize=1.0000\hsize \raggedleft\arraybackslash}X@{}}
\toprule
\textbf{Slippage per side (bps)} & \textbf{Unfiltered CR} & \textbf{O CR} & \textbf{RK CR} & \textbf{B CR} & \textbf{Joint CR} & \textbf{RK \ensuremath{\Delta}CR} & \textbf{Joint \ensuremath{\Delta}CR} \\
\midrule
0 & 3.95 & 7.60 & 12.81 & 10.06 & 11.59 & +8.86 & +7.65 \\
5 & 3.43 & 6.95 & 12.53 & 9.56 & 11.48 & +9.10 & +8.05 \\
10 & 2.91 & 6.62 & 12.41 & 9.21 & 11.40 & +9.50 & +8.50 \\
20 & 1.86 & 5.61 & 12.03 & 8.51 & 11.07 & +10.17 & +9.21 \\
50 & -1.41 & 3.24 & 10.98 & 5.98 & 10.38 & +12.39 & +11.78 \\
\bottomrule
\end{tabularx}
\renewcommand{\thetable}{C9}
\captionof{table}{Slippage sensitivity under the main replay setting. CR is the equally weighted mean across 12 actively trading models, expressed as a percentage; \ensuremath{\Delta}CR is the difference from the unfiltered account at the same slippage level, in percentage points.}
\label{tab:C9}
\endgroup
\par\addvspace{\floatsep}
\begingroup
\centering
\includegraphics[width=\textwidth]{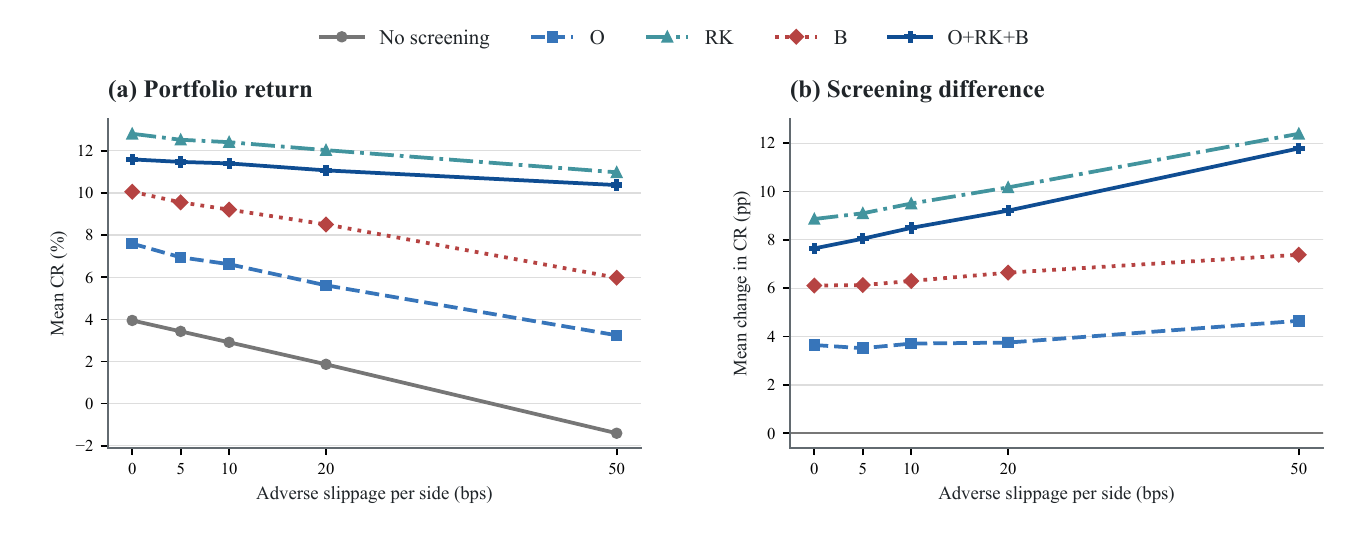}
\renewcommand{\thefigure}{C1}
\captionof{figure}{Slippage sensitivity. (a) Mean CR across 12 actively trading models; (b) mean CR difference between filtered and unfiltered accounts at the same slippage level. The horizontal axis shows slippage in basis points per side.}
\label{fig:C1}
\endgroup
\end{minipage}\par\addvspace{\floatsep}

\begin{multicols}{2}[][180pt]\raggedcolumns

\subsection{Complete Financial Metrics for Random Controls}

Random controls use the fixed seed 20261003, matching removal counts by model and buy/sell direction, and sample without replacement from full-period trades for 1,000 repetitions. Audit and random filtering both use the replay rules in Appendix C.1. Each repetition computes per-model changes from the original accounts and then takes the equally weighted mean across the 12 actively trading models.

Let \(k\) count random changes at least as large as the audit change. Monte Carlo reference values use \((k+1)/1001\), with Holm adjustment for multiple comparisons. For RK, Sharpe, Sortino, and Calmar adjusted reference values are 0.198, 0.076, and 0.040. These comparisons describe random deletions on the same historical trajectory, not confidence intervals across runs or market periods.

\end{multicols}

\par\noindent\begin{minipage}{\textwidth}
\centering
\small
\renewcommand{\arraystretch}{1.12}
\setlength{\tabcolsep}{4pt}
\begin{tabularx}{\textwidth}{@{}l
>{\raggedright\arraybackslash}X
*{2}{>{\centering\arraybackslash}X}
>{\hsize=1.5\hsize\centering\arraybackslash}X
>{\hsize=0.5\hsize\centering\arraybackslash}X@{}}
\toprule
\textbf{Filter} & \textbf{Metric} & \textbf{Audit change}
& \textbf{Random mean change} & \textbf{Random 95\% range}
& \textbf{Holm ref.} \\
\midrule
O & \ensuremath{\Delta}CR & +3.65 & +1.78 & [-1.37, +5.39] & 1.000 \\
O & \ensuremath{\Delta}Sharpe & +0.12 & +0.14 & [-0.30, +0.57] & 1.000 \\
O & \ensuremath{\Delta}Sortino & -0.06 & +0.12 & [-0.59, +0.95] & 1.000 \\
O & \ensuremath{\Delta}Calmar & +0.11 & +0.58 & [-0.64, +2.19] & 1.000 \\
RK & \ensuremath{\Delta}CR & +8.86 & +5.32 & [-0.08, +11.26] & 1.000 \\
RK & \ensuremath{\Delta}Sharpe & +0.61 & +0.03 & [-0.52, +0.52] & 0.198 \\
RK & \ensuremath{\Delta}Sortino & +0.82 & -0.31 & [-1.12, +0.45] & 0.076 \\
RK & \ensuremath{\Delta}Calmar & +2.68 & +0.08 & [-1.23, +1.70] & 0.040 \\
B & \ensuremath{\Delta}CR & +6.11 & +4.09 & [+0.12, +8.32] & 1.000 \\
B & \ensuremath{\Delta}Sharpe & +0.44 & +0.18 & [-0.25, +0.56] & 1.000 \\
B & \ensuremath{\Delta}Sortino & +0.43 & +0.06 & [-0.61, +0.92] & 1.000 \\
B & \ensuremath{\Delta}Calmar & +0.84 & +0.63 & [-0.60, +2.72] & 1.000 \\
O+RK+B & \ensuremath{\Delta}CR & +7.65 & +4.89 & [-0.89, +11.11] & 1.000 \\
O+RK+B & \ensuremath{\Delta}Sharpe & +0.52 & +0.02 & [-0.61, +0.57] & 0.831 \\
O+RK+B & \ensuremath{\Delta}Sortino & +0.52 & -0.28 & [-1.21, +0.61] & 0.713 \\
O+RK+B & \ensuremath{\Delta}Calmar & +1.35 & +0.14 & [-1.35, +1.89] & 1.000 \\
\bottomrule
\end{tabularx}
\renewcommand{\thetable}{C10}
\captionof{table}{Return and risk-adjusted metrics for audit filtering and random controls. The random 95\% range comprises the 2.5th to 97.5th percentiles of the 1,000 replays. \ensuremath{\Delta}CR is in percentage points; other metrics are ratio differences. Higher values indicate more favorable changes relative to the original accounts.}
\label{tab:C10}
\end{minipage}\par\addvspace{\floatsep}

\begin{multicols}{2}[][180pt]\raggedcolumns

\subsection{Summary of Filtering Definitions and Random Controls}

\end{multicols}

\par\noindent\begin{minipage}{\textwidth}
\centering
\small
\renewcommand{\arraystretch}{1.12}
\setlength{\tabcolsep}{4pt}
\begin{tabularx}{\textwidth}{@{}>{\raggedright\arraybackslash}p{0.16\textwidth}*{5}{>{\centering\arraybackslash}X}@{}}
\toprule
\textbf{Filter} & \multicolumn{1}{c}{\textbf{Trades}} & \multicolumn{1}{c}{\textbf{Return (pp)}} & \multicolumn{3}{c}{\textbf{Risk-adjusted ratios}} \\
\cmidrule(lr){2-2}\cmidrule(lr){3-3}\cmidrule(lr){4-6}
 & \textbf{Trades removed} & \textbf{\ensuremath{\Delta}CR} & \textbf{\ensuremath{\Delta}Sharpe} & \textbf{\ensuremath{\Delta}Sortino} & \textbf{\ensuremath{\Delta}Calmar} \\
\midrule
\rowcolor{TableBand}
O & 121 & \textbf{+3.50 (9)} & +0.09 (6) & -0.10 (6) & +0.05 (6) \\
RK & 803 & \textbf{+9.13 (11)} & +0.67 (9) & +0.84 (8) & +2.19 (8) \\
\rowcolor{TableBand}
B & 190 & \textbf{+4.63 (8)} & +0.30 (7) & +0.26 (7) & +0.49 (6) \\
\rowcolor{TableAccent}
O + RK + B & 898 & \textbf{+9.39 (11)} & +0.64 (10) & +0.69 (8) & +1.56 (9) \\
\bottomrule
\end{tabularx}
\renewcommand{\thetable}{C11}
\captionof{table}{Replay results with only confirmed errors filtered out and unable-to-determine trades retained. Complete financial metrics appear in Table C8.}
\label{tab:C11}
\end{minipage}\par\addvspace{\floatsep}

\par\noindent\begin{minipage}{\textwidth}
\centering
\small
\renewcommand{\arraystretch}{1.12}
\setlength{\tabcolsep}{4pt}
\begin{tabularx}{\textwidth}{@{}>{\raggedright\arraybackslash}p{0.1050\textwidth}>{\hsize=0.6977\hsize \raggedleft\arraybackslash}X>{\hsize=0.8140\hsize \raggedleft\arraybackslash}X>{\hsize=1.1047\hsize \raggedleft\arraybackslash}X>{\hsize=0.8140\hsize \raggedleft\arraybackslash}X>{\hsize=1.5698\hsize \raggedleft\arraybackslash}X@{}}
\toprule
\morecmidrules\cmidrule{1-6}
\textbf{Filter} & \textbf{Trades removed} & \textbf{Audit \ensuremath{\Delta}CR} & \textbf{Random mean \ensuremath{\Delta}CR} & \textbf{Above random} & \textbf{Random 95\% range} \\
\midrule
\rowcolor{TableBand}
O & 132 & +3.65 & +1.78 & +1.87 & [-1.37, +5.39] \\
RK & 849 & +8.86 & +5.32 & +3.55 & [-0.08, +11.26] \\
\rowcolor{TableBand}
B & 225 & +6.11 & +4.09 & +2.02 & [+0.12, +8.32] \\
\rowcolor{TableAccent}
O + RK + B & 948 & +7.65 & +4.89 & +2.75 & [-0.89, +11.11] \\
\cmidrule{1-6}
\bottomrule
\end{tabularx}
\renewcommand{\thetable}{C12}
\captionof{table}{Random-filtering controls matched by model and buy/sell direction under the baseline replay rules. Audits and random controls use the removal counts and execution settings in Table~\ref{tab:4}; intervals are the 2.5th to 97.5th percentiles across 1,000 random controls. Values are in percentage points.}
\label{tab:C12}
\end{minipage}\par\addvspace{\floatsep}

\begin{multicols}{2}[][180pt]\raggedcolumns

\FloatBarrier

\section{Reruns and Joint Gap Statistics}

\subsection{Complete Counts for the Three Dimensions}

Counts below retain pass, fail, unresolved, not-applicable, and unannotated states by decision point. Gap rates combine failures and expert unable-to-determine labels; unannotated items do not enter the check. Unresolved proportions among assessed applicable points for O, RK, and B are 0.00\%, 3.36\%, and 3.12\%, respectively; proportions for all-point and annotated-trading-point joint audits are 0.94\% and 2.80\%. D is inapplicable when no account statement exists.

\end{multicols}

\par\noindent\begin{minipage}{\textwidth}
\centering
\small
\begin{tabularx}{\textwidth}{@{}>{\raggedright\arraybackslash}p{0.2550\textwidth}>{\hsize=1.0000\hsize \raggedleft\arraybackslash}X>{\hsize=1.0000\hsize \raggedleft\arraybackslash}X>{\hsize=1.0000\hsize \raggedleft\arraybackslash}X>{\hsize=1.0000\hsize \raggedleft\arraybackslash}X@{}}
\toprule
\textbf{Model} & \textbf{O} & \textbf{RK} & \textbf{B} & \textbf{Joint} \\
\midrule
Claude Opus 4.5 & 243/3/0/0/0 & 43/78/2/122/1 & 107/11/0/128/0 & 162/82/2/0/0 \\
GLM 4.7 Flash & 223/23/0/0/0 & 13/40/0/193/0 & 28/15/7/196/0 & 189/56/1/0/0 \\
DeepSeek V3.2 & 231/15/0/0/0 & 33/81/6/123/3 & 99/16/4/127/0 & 151/90/5/0/0 \\
Gemini 3.1 Pro & 230/16/0/0/0 & 26/80/6/132/2 & 91/12/8/135/0 & 152/91/3/0/0 \\
Claude Haiku 4.5 & 214/32/0/0/0 & 34/110/3/95/4 & 69/75/3/99/0 & 97/146/3/0/0 \\
Qwen3 Max & 244/2/0/0/0 & 38/45/3/160/0 & 64/14/0/168/0 & 195/48/3/0/0 \\
DeepSeek V3.1 & 246/0/0/0/0 & 14/31/1/200/0 & 38/2/0/206/0 & 214/31/1/0/0 \\
GLM 4.7 & 221/25/0/0/0 & 12/16/2/216/0 & 17/7/2/220/0 & 202/44/0/0/0 \\
MiniMax M2.1 & 238/8/0/0/0 & 27/39/2/177/1 & 51/11/2/182/0 & 196/48/2/0/0 \\
GPT-5.1 & 246/0/0/0/0 & 0/0/0/246/0 & 0/0/0/246/0 & 246/0/0/0/0 \\
GPT-5.2 & 236/10/0/0/0 & 82/71/8/80/5 & 160/0/0/86/0 & 161/77/8/0/0 \\
Qwen3 235B & 181/65/0/0/0 & 8/32/0/206/0 & 21/10/3/212/0 & 157/88/1/0/0 \\
Gemini 2.5 Flash & 218/28/0/0/0 & 26/55/3/158/4 & 65/12/3/166/0 & 164/81/1/0/0 \\
\bottomrule
\end{tabularx}
\renewcommand{\thetable}{D1}
\captionof{table}{Pass/fail/unresolved/not-applicable/unannotated counts. Each cell sums to 246; applicable-check gap rates exclude not-applicable and unannotated items. Tables B4 and B5 report all-point and annotated-trading-point joint results.}
\label{tab:D1}
\end{minipage}\par\addvspace{\floatsep}

\begin{multicols}{2}[][180pt]\raggedcolumns

\subsection{Individual Results for Ten-Case Reruns}

The four models are each tested on ten independent historical snapshots, producing 41 net trade events in total. Each test restores market information and the pre-trade account state at the corresponding decision time and uses the same system prompt, interaction-step limit, and number of overall attempts to maintain consistent experimental settings. All outputs are assessed under the same O, RK, and D rules. Figure 4 presents the audit results for each model under the same experimental conditions.

Two rounds of independent blind annotation are performed for R/K/D across the 41 net trade events in the fixed ten-case sets. Raw agreement between the two rounds is as follows:

\end{multicols}

\par\noindent\begin{minipage}{\textwidth}
\centering
\small
\begin{tabularx}{\textwidth}{@{}>{\raggedright\arraybackslash}p{0.2050\textwidth}>{\hsize=1.0000\hsize \raggedleft\arraybackslash}X>{\hsize=1.0000\hsize \raggedleft\arraybackslash}X>{\hsize=1.0000\hsize \raggedleft\arraybackslash}X>{\hsize=1.0000\hsize \raggedleft\arraybackslash}X@{}}
\toprule
\textbf{Item} & \textbf{Agreements/\allowbreak{}events} & \textbf{Raw agreement} & \textbf{Cohen's \ensuremath{\kappa}} & \textbf{Gwet's AC1} \\
\midrule
R & 41/41 & 100.0\% & 1.000 & 1.000 \\
K & 40/41 & 97.6\% & 0.959 & 0.965 \\
D & 41/41 & 100.0\% & 1.000 & 1.000 \\
\bottomrule
\end{tabularx}
\renewcommand{\thetable}{D1b}
\captionof{table}{Annotation agreement for historical snapshot tests.}
\label{tab:D1b}
\end{minipage}\par\addvspace{\floatsep}

\begin{multicols}{2}[][180pt]\raggedcolumns

Raw agreement includes records both rounds label unable to determine; AC1 is calculated using each item's complete label categories.

\end{multicols}

\par\noindent\begin{minipage}{\textwidth}
\centering
\small
\begin{tabularx}{\textwidth}{@{}>{\raggedright\arraybackslash}p{0.2550\textwidth}>{\hsize=1.0000\hsize \raggedleft\arraybackslash}X>{\hsize=1.0000\hsize \raggedleft\arraybackslash}X>{\hsize=1.0000\hsize \raggedleft\arraybackslash}X>{\hsize=1.0000\hsize \raggedleft\arraybackslash}X@{}}
\toprule
\textbf{Model} & \textbf{Report} & \textbf{Rpt} & \textbf{Call} & \textbf{Resp} \\
\midrule
DeepSeek V3.2 & 60.0 & 60.0 & 20.0 & 20.0 \\
GLM 4.7 Flash & 0.0 & 10.0 & 0.0 & 0.0 \\
GPT-6 Astra(max) & 0.0 & 0.0 & 0.0 & 0.0 \\
GPT-5.6 Sol & 0.0 & 0.0 & 0.0 & 0.0 \\
\bottomrule
\end{tabularx}
\renewcommand{\thetable}{D2}
\captionof{table}{Gap rates for O's four checks (\%, \ensuremath{\downarrow}). Report uses 10 windows as its denominator; the other checks use applicable windows. Complete item states appear in Table D2a.}
\label{tab:D2}
\end{minipage}\par\addvspace{\floatsep}

\par\noindent\begin{minipage}{\textwidth}
\centering
\small
\begin{tabularx}{\textwidth}{@{}>{\raggedright\arraybackslash}p{0.2550\textwidth}>{\hsize=1.0000\hsize \raggedleft\arraybackslash}X>{\hsize=1.0000\hsize \raggedleft\arraybackslash}X>{\hsize=1.0000\hsize \raggedleft\arraybackslash}X>{\hsize=1.0000\hsize \raggedleft\arraybackslash}X@{}}
\toprule
\textbf{Model} & \textbf{Report} & \textbf{Rpt} & \textbf{Call} & \textbf{Resp} \\
\midrule
DeepSeek V3.2 & 4/6/0/0 & 4/0/6/0 & 8/0/2/0 & 8/0/2/0 \\
GLM 4.7 Flash & 10/0/0/0 & 9/1/0/0 & 10/0/0/0 & 10/0/0/0 \\
GPT-6 Astra(max) & 10/0/0/0 & 10/0/0/0 & 10/0/0/0 & 10/0/0/0 \\
GPT-5.6 Sol & 10/0/0/0 & 10/0/0/0 & 10/0/0/0 & 10/0/0/0 \\
\bottomrule
\end{tabularx}
\renewcommand{\thetable}{D2a}
\captionof{table}{Complete counts for O's four checks. Counts are ordered as pass/confirmed failure/undetermined/not applicable.}
\label{tab:D2a}
\end{minipage}\par\addvspace{\floatsep}

\begin{multicols}{2}[][180pt]\raggedcolumns

The joint seven-check audit comprises Report, Rpt, Call, Resp, R, K, and D. O and joint rates use each group's 10 decision points; RK and D use applicable points, and gaps combine confirmed failures and undetermined cases. D has 5, 2, 10, and 10 applicable points, including 4 unable-to-determine DeepSeek points. GLM's D gap rate is 1/2=50.0\% by decision point and 33.3\% by trade event (Table D3c).

\end{multicols}

\par\noindent\begin{minipage}{\textwidth}
\centering
\small
\begin{tabularx}{\textwidth}{@{}>{\raggedright\arraybackslash}p{0.2550\textwidth}>{\hsize=1.0000\hsize \raggedleft\arraybackslash}X>{\hsize=1.0000\hsize \raggedleft\arraybackslash}X>{\hsize=1.0000\hsize \raggedleft\arraybackslash}X>{\hsize=1.0000\hsize \raggedleft\arraybackslash}X@{}}
\toprule
\textbf{Model} & \textbf{Trading windows} & \textbf{Joint passes} & \textbf{Confirmed failures} & \textbf{Undetermined} \\
\midrule
DeepSeek V3.2 & 6 & 2 & 8 & 0 \\
GLM 4.7 Flash & 3 & 7 & 3 & 0 \\
GPT-6 Astra(max) & 10 & 8 & 2 & 0 \\
GPT-5.6 Sol & 10 & 5 & 5 & 0 \\
\bottomrule
\end{tabularx}
\renewcommand{\thetable}{D3}
\captionof{table}{Joint seven-check states. A point passes only when all applicable checks pass with no undetermined items; any confirmed failure takes precedence and determines joint failure; points without failures but with undetermined items remain undetermined.}
\label{tab:D3}
\end{minipage}\par\addvspace{\floatsep}

\par\noindent\begin{minipage}{\textwidth}
\centering
\small
\begin{tabularx}{\textwidth}{@{}>{\raggedright\arraybackslash}p{0.2550\textwidth}>{\hsize=1.0000\hsize \raggedleft\arraybackslash}X>{\hsize=1.0000\hsize \raggedleft\arraybackslash}X>{\hsize=1.0000\hsize \raggedleft\arraybackslash}X>{\hsize=1.0000\hsize \raggedleft\arraybackslash}X@{}}
\toprule
\textbf{Model} & \textbf{O pass/\allowbreak{}fail/\allowbreak{}undetermined/\allowbreak{}not applicable} & \textbf{RK pass/\allowbreak{}fail/\allowbreak{}undetermined/\allowbreak{}not applicable} & \textbf{D pass/\allowbreak{}fail/\allowbreak{}undetermined/\allowbreak{}not applicable} & \textbf{Joint pass/\allowbreak{}fail/\allowbreak{}undetermined/\allowbreak{}not applicable} \\
\midrule
DeepSeek V3.2 & 4/6/0/0 & 0/2/4/4 & 1/0/4/5 & 2/8/0/0 \\
GLM 4.7 Flash & 9/1/0/0 & 1/2/0/7 & 1/1/0/8 & 7/3/0/0 \\
GPT-6 Astra(max) & 10/0/0/0 & 8/2/0/0 & 10/0/0/0 & 8/2/0/0 \\
GPT-5.6 Sol & 10/0/0/0 & 5/5/0/0 & 10/0/0/0 & 5/5/0/0 \\
\bottomrule
\end{tabularx}
\renewcommand{\thetable}{D3a}
\captionof{table}{Complete counts for O, RK, D, and the joint seven-check audit. Counts are ordered as pass/confirmed failure/undetermined/not applicable; applicable denominators equal totals minus not-applicable cases, including undetermined cases.}
\label{tab:D3a}
\end{minipage}\par\addvspace{\floatsep}

\par\noindent\begin{minipage}{\textwidth}
\centering
\small
\begin{tabularx}{\textwidth}{
@{}>{\raggedright\arraybackslash}p{0.2550\textwidth}
*{3}{>{\centering\arraybackslash}X}@{}}
\toprule
\textbf{Model} & \textbf{R} & \textbf{K} & \textbf{D} \\
\midrule
DeepSeek V3.2 & 1/1/4/4 & 0/2/4/4 & 1/0/4/5 \\
GLM 4.7 Flash & 2/0/1/7 & 1/2/0/7 & 1/1/0/8 \\
GPT-6 Astra(max) & 10/0/0/0 & 8/2/0/0 & 10/0/0/0 \\
GPT-5.6 Sol & 10/0/0/0 & 5/5/0/0 & 10/0/0/0 \\
\bottomrule
\end{tabularx}
\renewcommand{\thetable}{D3b}
\captionof{table}{Decision-point states for R, K, and D.
Multi-security events use the joint aggregation rules;
no verifiable account assessment maps to D not applicable.}
\label{tab:D3b}
\end{minipage}\par\addvspace{\floatsep}

\par\noindent\begin{minipage}{\textwidth}
\centering
\small
\begin{tabularx}{\textwidth}{
@{}>{\raggedright\arraybackslash}p{0.2550\textwidth}
>{\centering\arraybackslash}X
>{\hsize=1.5\hsize\centering\arraybackslash}X
>{\hsize=0.5\hsize\centering\arraybackslash}X@{}}
\toprule
\textbf{Model} & \textbf{Net trade events}
& \textbf{D pass/\allowbreak{}error/\allowbreak{}unable to determine/\allowbreak{}no account assessment}
& \textbf{D gap rate \%} \\
\midrule
DeepSeek V3.2 & 9 & 1/0/7/1 & 87.5 \\
GLM 4.7 Flash & 4 & 2/1/0/1 & 33.3 \\
GPT-6 Astra(max) & 18 & 18/0/0/0 & 0.0 \\
GPT-5.6 Sol & 10 & 10/0/0/0 & 0.0 \\
\bottomrule
\end{tabularx}
\renewcommand{\thetable}{D3c}
\captionof{table}{Net trade event counts for account facts.
D gap rates combine confirmed errors and unable-to-determine
cases and exclude events with no account assessment.}
\label{tab:D3c}
\end{minipage}\par\addvspace{\floatsep}

\begin{multicols}{2}[][180pt]\raggedcolumns

DeepSeek V3.2 lacks formal reports in four trading windows covering seven net trade events, whose R, K, and D records are labeled unable to determine. Statistics cover each model's fixed ten historical windows; additional GLM attempts are archived separately.

\subsection{Supplementary Results for Continuous Trading by the New Model}

The continuous trading experiment uses the original 246 historical snapshots, carrying account cash and positions forward across windows.

RK filtering raises CR from 4.85\% to 38.28\%, Sharpe from 1.02 to 2.86, and Sortino from 1.88 to 5.20. All-window maximum drawdown magnitude increases from 8.23\% to 15.02\%; daily financial metrics appear in Table C7.

\end{multicols}

\par\noindent\begin{minipage}{\textwidth}
\centering
\small
\begin{tabularx}{\textwidth}{@{}>{\raggedright\arraybackslash}p{0.2550\textwidth}>{\hsize=1.0000\hsize \raggedleft\arraybackslash}X>{\hsize=1.0000\hsize \raggedleft\arraybackslash}X>{\hsize=1.0000\hsize \raggedleft\arraybackslash}X>{\hsize=1.0000\hsize \raggedleft\arraybackslash}X@{}}
\toprule
\textbf{Model} & \textbf{Report} & \textbf{Rpt} & \textbf{Call} & \textbf{Resp} \\
\midrule
GPT-6 Sol / medium & 0.0 & 0.0 & 0.0 & 0.0 \\
\bottomrule
\end{tabularx}
\renewcommand{\thetable}{D4}
\captionof{table}{Gap rates for O's four checks on GPT-6 Sol's continuous trajectory (\%, \ensuremath{\downarrow}). All denominators are 246.}
\label{tab:D4}
\end{minipage}\par\addvspace{\floatsep}

\begin{multicols}{2}[][180pt]\raggedcolumns

Across all 246 points, 205 pass the joint audit, 36 have confirmed failures, and 5 are unable to determine, giving a joint gap rate of 16.67\%. The corresponding trading-point counts are 55, 36, and 5, giving a trading joint gap rate of 42.71\%. Temp is not applicable to the first trade, while other applicable checks must still pass. Among 87 trades for which all eight metrics apply, 48 pass, 34 have confirmed failures, and 5 are unable to determine, giving a combined gap rate of 44.83\%.

\FloatBarrier

\section{Securities, Data Sources, and Runtime Configurations}

\subsection{Stock Pool and Data Sources}

All ten stocks in Table E1 are listed on the Shanghai Stock Exchange's STAR Market, with accounts and market prices denominated in CNY. Prices and technical indicators use TinySoft market data, while news and announcements are obtained through Tushare, AkShare, Sina Finance, the Shanghai Stock Exchange, and CNINFO, truncated at each decision time.

\end{multicols}

\par\noindent\begin{minipage}{\textwidth}
\centering
\small
\begin{tabularx}{\textwidth}{@{}>{\raggedright\arraybackslash}p{0.1250\textwidth}>{\hsize=1.2143\hsize \raggedright\arraybackslash}X>{\hsize=0.8929\hsize \raggedright\arraybackslash}X>{\hsize=0.8929\hsize \raggedright\arraybackslash}X@{}}
\toprule
\textbf{Security code} & \textbf{Company} & \textbf{Industry} & \textbf{Market} \\
\midrule
SH688008 & Montage Technology & Semiconductors & SSE STAR Market \\
SH688111 & Kingsoft Office & Software & SSE STAR Market \\
SH688009 & China Railway Signal \& Communication & Rail transport & SSE STAR Market \\
SH688981 & SMIC & Semiconductors & SSE STAR Market \\
SH688256 & Cambricon & Semiconductors & SSE STAR Market \\
SH688271 & United Imaging Healthcare & Medical devices & SSE STAR Market \\
SH688047 & Loongson Technology & Semiconductors & SSE STAR Market \\
SH688617 & APT Medical & Medical devices & SSE STAR Market \\
SH688303 & Daqo New Energy & Photovoltaic materials & SSE STAR Market \\
SH688180 & Junshi Biosciences & Biopharmaceuticals & SSE STAR Market \\
\bottomrule
\end{tabularx}
\renewcommand{\thetable}{E1}
\captionof{table}{Stock pool.}
\label{tab:E1}
\end{minipage}\par\addvspace{\floatsep}

\par\noindent\begin{minipage}{\textwidth}
\centering
\small
\begin{tabularx}{\textwidth}{@{}>{\raggedright\arraybackslash}p{0.2983\textwidth}>{\hsize=1.0000\hsize \raggedright\arraybackslash}X>{\hsize=1.0000\hsize \raggedright\arraybackslash}X@{}}
\toprule
\textbf{Data} & \textbf{Source} & \textbf{Contents} \\
\midrule
Individual-stock market data & TinySoft & Prices and hourly market data for the ten securities, from which technical indicators are calculated \\
Financial news & Tushare, Sina Finance & Financial information from Sina Finance, Tonghuashun, East Money, Wallstreetcn, Cailian Press, Yicai, JRJ, and other outlets \\
Stock-specific information and events & AkShare & Individual-stock news, company developments, and financial events \\
Company announcements & Shanghai Stock Exchange, CNINFO & Company announcements and available announcement texts \\
\bottomrule
\end{tabularx}
\renewcommand{\thetable}{E2}
\captionof{table}{Data sources.}
\label{tab:E2}
\end{minipage}\par\addvspace{\floatsep}

\begin{multicols}{2}[][180pt]\raggedcolumns

News retrieval looks back 30 days, returning at most 20 items per security, deduplicated by security, publication time, and normalized title; different models receive the shared snapshot for the same window.

\subsection{Model and Interaction Configurations}

The main experiment calls models through an OpenAI-compatible interface, using a uniform ReAct workflow with at most 30 interaction steps per window and 3 attempts per agent task. Table E3 lists model identifiers from the actual call records. Models use the interface's default reasoning configuration.

\end{multicols}

\par\noindent\begin{minipage}{\textwidth}
\centering
\small
\begin{tabularx}{\textwidth}{@{}>{\raggedright\arraybackslash}p{0.2350\textwidth}>{\hsize=1.3699\hsize \raggedright\arraybackslash}X>{\hsize=0.6301\hsize \raggedright\arraybackslash}X@{}}
\toprule
\textbf{Model} & \textbf{Model identifier in call records} & \textbf{Additional reasoning request} \\
\midrule
Claude Opus 4.5 & anthropic/claude-4.5-opus-20251124 & None specified \\
GLM 4.7 Flash & z-ai/glm-4.7-flash-20260119 & None specified \\
DeepSeek V3.2 & deepseek/deepseek-v3.2 & None specified \\
Gemini 3.1 Pro & google/gemini-3.1-pro-preview-20260219 & None specified \\
Claude Haiku 4.5 & anthropic/claude-4.5-haiku-20251001 & None specified \\
Qwen3 Max & qwen/qwen3-max & None specified \\
DeepSeek V3.1 & deepseek/deepseek-v3.1-terminus & None specified \\
GLM 4.7 & z-ai/glm-4.7; z-ai/glm-4.7-20251222 & None specified \\
MiniMax M2.1 & minimax/minimax-m2.1 & None specified \\
GPT-5.1 & openai/gpt-5.1-20251113 & None specified \\
GPT-5.2 & openai/gpt-5.2-20251211 & None specified \\
Qwen3 235B & qwen/qwen3-235b-a22b-04-28 & None specified \\
Gemini 2.5 Flash & google/gemini-2.5-flash & None specified \\
\bottomrule
\end{tabularx}
\renewcommand{\thetable}{E3}
\captionof{table}{Main-experiment model identifiers.}
\label{tab:E3}
\end{minipage}\par\addvspace{\floatsep}

\begin{multicols}{2}[][180pt]\raggedcolumns

Parameters such as temperature and top\_p use the interface defaults.

The uniform prompt asks models to inspect candidate stocks, account positions, and trading constraints, autonomously choose to buy, sell, or not trade, and submit structured reports containing material citations, trade descriptions, and decision rationales. Tools cover news retrieval, technical indicators, hourly market data, current prices, account-state queries, purchases, sales, No Trade recording, and report submission. Specific budgets for historical cases and continuous trading by the new model appear in Appendices D.2 and D.3.

\subsection{Reference Benchmark Settings}

The equally weighted buy-and-hold portfolio uses the same ten stocks; SSE 50 uses a normalized index series for the same period as a market reference. Five rule-based strategies start from empty accounts with CNY 1 million and calculate signals on shared snapshots; when a signal is active and the security is not already held, the target purchase value is 10\% of pre-trade total assets; when the signal disappears, sellable holdings are sold. Strategies apply 100-share purchase lots, cash constraints, a short-selling prohibition, T+1, a 50\% single-stock limit, and the same commissions and stamp duty as the main experiment.

\end{multicols}

\par\noindent\begin{minipage}{\textwidth}
\centering
\small
\begin{tabularx}{\textwidth}{@{}>{\raggedright\arraybackslash}p{0.2983\textwidth}>{\hsize=1.0000\hsize \raggedright\arraybackslash}X>{\hsize=1.0000\hsize \raggedright\arraybackslash}X@{}}
\toprule
\textbf{Strategy} & \textbf{Parameters} & \textbf{Buy signal} \\
\midrule
RSI & 14 periods & RSI below 30 \\
MACD & 12/26/9 & Histogram value above 0 \\
SMA & 5 and 20 periods & 5-period moving average above the 20-period moving average \\
KDJ (closing price) & 9 periods & K above D and below 80 \\
Mean reversion & 60 periods & Closing price more than 3\% below the rolling mean \\
\bottomrule
\end{tabularx}
\renewcommand{\thetable}{E4}
\captionof{table}{Strategy benchmark parameters. Periods are calculated over the shared snapshot sequence.}
\label{tab:E4}
\end{minipage}\par\addvspace{\floatsep}

\begin{multicols}{2}[][180pt]\raggedcolumns

The KDJ benchmark constructs signals using only snapshot closing prices. All market and strategy benchmarks cover the same 82 trading days and use daily closing account values to calculate the financial metrics in Table B6a.

\subsection{Material Language and Saved Content}

News, company announcements, and annotation instructions are primarily in Chinese, while model reports and tool interfaces contain English structured fields. Annotators view the titles, summaries, or full texts saved in the contemporaneous archives; citation checks and key-fact assessments retain the corresponding original text for reference.

The key-fact material packets used in the independent second round of annotation primarily contain news titles and summaries within titles, and key facts are extracted only from content explicitly stated in these materials. Materials corresponding to the current formal trades comprise 872 packets and 68,741 records. Publication times are no later than the corresponding decision times, and material identifiers cited by key facts can be located in the corresponding packets.

\subsection{Verification Records and Audit Dimensions}

\end{multicols}

\par\noindent\begin{minipage}{\textwidth}
\centering
\small
\begin{tabularx}{\textwidth}{@{}>{\raggedright\arraybackslash}p{0.1350\textwidth}>{\hsize=0.7229\hsize \raggedright\arraybackslash}X>{\hsize=1.2771\hsize \raggedright\arraybackslash}X@{}}
\toprule
\textbf{Record type} & \textbf{Record contents} & \textbf{Core audit use} \\
\midrule
\textbf{Model report} & Structured trade descriptions and rationales & Check whether the report exists and has a valid structure; extract descriptions and rationales for subsequent verification \\
\textbf{Trade request} & Security code, buy/sell direction, and requested quantity submitted by the model & Aggregate net trade quantities in requests and compare them with net account position changes \\
\textbf{Execution response} & Processing status and execution records returned by the trading interface & Aggregate net trade quantities recorded in successful responses and compare them with net account position changes \\
\textbf{Account snapshot} & Position quantities, cash, sellable quantities, and other states before and after the window & Verify actual trade effects through net position changes and provide a reference for account-fact assessments \\
\bottomrule
\end{tabularx}
\renewcommand{\thetable}{E5}
\captionof{table}{Underlying verification records and their audit uses.}
\label{tab:E5}
\end{minipage}\par\addvspace{\floatsep}

\par\noindent\begin{minipage}{\textwidth}
\centering
\small
\begin{tabularx}{\textwidth}{@{}>{\raggedright\arraybackslash}p{0.3750\textwidth}>{\hsize=0.7119\hsize \raggedright\arraybackslash}X>{\hsize=1.2881\hsize \raggedright\arraybackslash}X@{}}
\toprule
\textbf{Audit question} & \textbf{Audit dimension} & \textbf{Associated records} \\
\midrule
Was a valid report retained, and do reports, requests, and responses match outcomes posted to the ledger? & \textbf{O: Execution traceability} & Model reports, trade requests, execution responses, and actual booked trade effects \\
Are the materials supporting the current decision sufficient to support the stated trading claim? & \textbf{RK: Evidence sufficiency} & Cited materials, trading claims, and verified actual trade directions \\
Do decision rationales accord with account facts, and are successive trades reasonably explained? & \textbf{B: Financial meta-awareness} & Model reports, account facts, and sequential trading records \\
\bottomrule
\end{tabularx}
\renewcommand{\thetable}{E6}
\captionof{table}{Core questions and associated records in the three-dimensional decision-auditing framework.}
\label{tab:E6}
\end{minipage}\par\addvspace{\floatsep}

\begin{multicols}{2}[][180pt]\raggedcolumns

\subsection{Open-Source Resources}

We will release four resource groups: \textbf{evaluation code}, including agents, tool interfaces, and account execution logic; \textbf{expert annotations}, including anonymized A/B labels, final consensus, rationales, and source references; \textbf{analysis and replay scripts} to reproduce audits, filtered replay, and figures; and \textbf{prompts}, including trading and semantic annotation instructions, fixed few-shot examples, and exclusion lists. Market data, news, and announcements are accessed through the system's data interfaces; the README will document data sources, environment configuration, and execution commands.

\end{multicols}

\FloatBarrier

\end{document}